\documentclass[twoside,11pt]{article}

\usepackage{jmlr2e} 
\usepackage{bm, amsmath, amsfonts, amssymb}
\usepackage{multirow}
\usepackage[figuresright]{rotating}
\usepackage{amsmath}
\usepackage{cases}
\usepackage{color}
\usepackage{rotating}
\usepackage{setspace}
\usepackage{graphics, graphicx}
\usepackage{subfigure}
\usepackage{float}
\usepackage{epsfig}
\usepackage{lscape}
\usepackage{multirow}
\usepackage{mathrsfs,amssymb}
\usepackage{lscape}
\usepackage{threeparttable}
\usepackage{bm, amsmath, amsfonts, amssymb}
\usepackage{natbib}
\usepackage{hyperref}
\usepackage{csquotes}
\usepackage{array,booktabs}
\newcommand{\beq}{\begin{equation}}
\newcommand{\eeq}{\end{equation}}
\newcommand{\beas}{\begin{align*}}
\newcommand{\eeas}{\end{align*}}
\newcommand{\bea}{\begin{align}}
\newcommand{\eea}{\end{align}}
\newcommand{\bei}{\begin{itemize}}
	\newcommand{\eei}{\end{itemize}}
\newcommand{\ben}{\begin{enumerate}}
	\newcommand{\een}{\end{enumerate}}
\newcommand{\bet}{\begin{theorem}}
	\newcommand{\eet}{\end{theorem}}
\newcommand{\bel}{\begin{lemma}}
	\newcommand{\eel}{\end{lemma}}
\newcommand{\bep}{\begin{proposition}}
	\newcommand{\eep}{\end{proposition}}
\newcommand{\bec}{\begin{corollary}}
	\newcommand{\eec}{\end{corollary}}

\newcommand{\bu}{\bold{u}}

\newcommand{\bg}{\bold{g}}

\newcommand{\bt}{\bold{t}}

\newcommand{\bz}{\bold{z}}

\newcommand{\bU}{\bold{U}}
\newcommand{\bE}{\bold{E}}

\newcommand{\bV}{\bold{V}}
\newcommand{\bG}{\bold{G}}

\newcommand{\bQ}{\bold{Q}}
\newcommand{\bH}{\bold{H}}

\newcommand{\bA}{\bold{A}}
\newcommand{\bL}{\bold{L}}
\newcommand{\bK}{\bold{K}}
\newcommand{\bP}{\bold{P}}
\newcommand{\bY}{\bold{Y}}

\newcommand{\bD}{\bold{D}}

\newcommand{\bS}{\bold{S}}
\newcommand{\bGam}{\bold{\Gamma}}

\newcommand{\bepsilon}{\boldsymbol{\epsilon}}
\newcommand{\bbeta}{\boldsymbol{\beta}}
 
\newcommand{\btheta}{\boldsymbol{\theta}}

\newcommand{\yy}{\boldsymbol{y}}

\newcommand{\R}{\mathbb{R}}
\newcommand{\E}{\mathbb{E}}

\newcommand{\argmin}{\mathop{\rm arg\min}}
\newcommand{\argmax}{\mathop{\rm arg\max}}

\usepackage{lastpage}
	\jmlrheading{23}{2022}{1-\pageref{LastPage}}{5/21; Revised
		10/22}{10/22}{21-0524}{Tony Cai and Rong Ma}
	\ShortHeadings{Theoretical Foundation of  t-SNE}{Cai and Ma}
\firstpageno{1}

\begin{document}

\title{Theoretical Foundations of  t-SNE for Visualizing High-Dimensional Clustered Data}

\author{\name T. Tony Cai \email tcai@wharton.upenn.edu \\
       \addr Department of Statistics and Data Science\\
       University of Pennsylvania\\
      Philadelphia, PA 19104, USA  
       \AND       
   \name Rong Ma \email rongm@stanford.edu \\
   \addr Department of Statistics\\
   Stanford University\\
   Stanford, CA 94305, USA}

 \editor{Ji Zhu}

\maketitle

\begin{abstract}%   <- trailing '%' for backward compatibility of .sty file
This paper investigates the theoretical foundations of the t-distributed stochastic neighbor embedding (t-SNE) algorithm, a popular nonlinear dimension reduction and data visualization method. A novel theoretical framework for the analysis of t-SNE based on the gradient descent approach is presented. For the early exaggeration stage of t-SNE, we show its asymptotic equivalence to power iterations based on the underlying graph Laplacian, characterize its limiting behavior, and uncover its deep connection to  Laplacian spectral clustering, and fundamental principles including early stopping as implicit regularization. The results explain the intrinsic mechanism and the empirical benefits of such a computational strategy. For the embedding stage of t-SNE, we characterize the kinematics of the low-dimensional map throughout the iterations, and identify an amplification phase, featuring the intercluster repulsion and the expansive behavior of the low-dimensional map, and a stabilization phase. The general theory explains the fast convergence rate and the exceptional empirical performance of t-SNE for visualizing clustered data, brings forth the interpretations of the t-SNE visualizations, and provides theoretical guidance for applying t-SNE and selecting its tuning parameters in various applications.
\end{abstract}


\section{Introduction} \label{intro.sec}

Data visualization is  critically important for understanding and interpreting the structure of large datasets, 
and has been recognized as one of the fundamental topics in data science \citep{donoho201750}.  A collection of machine learning algorithms for data visualization and dimension reduction have been developed. Among them, the t-distributed stochastic neighbor embedding (t-SNE) algorithm, proposed by \cite{maaten2008visualizing}, is arguably one of the most popular methods and a state-of-art technique for a wide range of applications \citep{wang2021understanding}.  

Specifically, t-SNE is an  iterative algorithm for visualizing high-dimensional data by mapping the data points to a two- or finite-dimensional space. It creates a single map that reveals the intrinsic structures in a high-dimensional dataset, including trends, patterns, and outliers, 
through a nonlinear dimension reduction technique. In the past decade, the original t-SNE algorithm, along with its many variants (for example, \cite{yang2009heavy,carreira2010elastic,xie2011m,van2014accelerating,gisbrecht2015parametric,pezzotti2016approximated,im2018stochastic,linderman2019fast,chatzimparmpas2020t}), has made profound impact to the practice of scientific research, including genetics \citep{platzer2013visualization}, molecular biology \citep{olivon2018metgem}, single-cell transcriptomics \citep{kobak2019art}, computer vision \citep{cheng2015silhouette} and astrophysics \citep{traven2017galah}. In particular, the extraordinary performance of t-SNE for visualizing high-dimensional data with intrinsic clusters has  been widely acknowledged \citep{van2014accelerating,kobak2019art}. 

Compared to the extensive literature on the computational and numerical aspects of t-SNE, there is a paucity of  fundamental results about its theoretical foundations (see Section \ref{related.work} for a brief overview). The lack of  theoretical understanding and justifications profoundly  limits the users' interpretation of the results as well as the potentials for further improvement of the method.

This paper aims to investigate the theoretical foundations of t-SNE. Specifically, we present a  novel framework for the analysis of  t-SNE, provide  theoretical justifications for its competence in dimension reduction and visualizing clustered data, and uncover the fundamental principles underlying its exceptional empirical performance.

\subsection{Basic t-SNE Algorithm} \label{tsne.intro.sec}

Let $\{X_i\}_{1\le i\le n}$ be a set of $p$-dimensional data points.
t-SNE  starts by computing a joint probability distribution  over all pairs of data points $\{(X_i,X_j)\}_{1\le i\ne j\le n}$, represented by a symmetric matrix $\bP=(p_{ij})_{1\le i, j\le n}\in \R^{n\times n}$, where $p_{ii}=0$ for all $1\le i\le n$, and for $i\ne j$, 
\beq \label{P.def}
p_{ij}=\frac{p_{i|j}+p_{j|i}}{2n} \quad\text{with}\quad  p_{j|i}=\frac{\exp(-\|X_i-X_j\|_2^2/2\tau_i^2)}{\sum_{\ell\in\{1,2,...,n\}\setminus\{i\}} \exp(-\|X_i-X_\ell\|_2^2/2\tau_i^2)}.
\eeq
Here $\tau_i$ are tuning parameters, which are usually determined based on a certain perplexity measure and a binary search strategy \citep{hinton2002stochastic,maaten2008visualizing}. Similarly, for a two-dimensional\footnote{Throughout, we focus on the two-dimensional embedding for ease of presentation. However, all the theoretical results obtained in this work holds for any  finite constant embedding dimension.} map $\{y_i\}_{1\le i\le n}\subset \R^2$, we define the joint probability distribution over  all pairs  $\{(y_i,y_j)\}_{1\le i\ne j\le n}$ through a symmetric matrix  $\bQ=(q_{ij})_{1\le i, j\le n}$ where $q_{ii}=0$ for all $1\le i\le n$ and for $i\ne j$,
\beq \label{Q.def}
q_{ij}=\frac{(1+\|y_i-y_j\|_2^2)^{-1}}{\sum_{\ell,s\in\{1,2,...,n\},\ell\ne s} (1+\|y_\ell-y_s\|_2^2)^{-1}}.
\eeq
Intuitively, $\bP$ and $\bQ$ are similarity matrices summarizing the pairwise distances of the high-dimensional data points $\{X_i\}_{1\le i\le n}$, and the two-dimensional map $\{y_i\}_{1\le i\le n}$, respectively.
Then  t-SNE aims to  find $\{y_i\}_{1\le i\le n}$ that minimizes the KL-divergence between $\bP$ and $\bQ$, that is,
\beq
(y_1,...,y_n)=\argmin_{y_1,...,y_n}D_{KL}(\bP,\bQ)=\argmin_{y_1,...,y_n}\sum_{\substack{i,j\in\{1,2,...,n\}\\
		i\ne j}} p_{ij}\log\frac{p_{ij}}{q_{ij}}.
\eeq
Many algorithms have been proposed to solve this optimization problem. The most widely used algorithm was proposed in
\cite{maaten2008visualizing}, which draws on a variant of gradient descent algorithm, with an updating equation
\beq \label{tsne.ori}
y_i^{(k+1)}=y_i^{(k)}+hD_i^{(k)}+m^{(k+1)}(y_i^{(k)}-y_i^{(k-1)}),\quad \text{for $i=1,...,n,$}
\eeq
where $h\in \R_+$ is a prespecified step size parameter, $D_i^{(k)}= 4\sum_{1\le j\le n,j\ne i}(y^{(k)}_j-y^{(k)}_i)S^{(k)}_{ij}\in \R^2$ is the gradient term corresponding to $y_i$, with $S^{(k)}_{ij}=(p_{ij}-q^{(k)}_{ij})/(1+\|y_i^{(k)}-y_j^{(k)}\|_2^2)\in \R$, and $m^{(k)}\in \R_+$ is a prespecified momentum parameter. The algorithm starts with an initialization $y^{(0)}_{i}=y^{(-1)}_{i}$ for $i\in\{1,2,...,n\}$,  drawn independently from a uniform distribution on $[-0.01,0.01]^2$, or from $N(0,\delta^2 I)$ for some small $\delta> 0$. 
 
As indicated by  \cite{maaten2008visualizing}, the inclusion of the momentum term $m^{(k+1)}(y_i^{(k)}-y_i^{(k-1)})$ in (\ref{tsne.ori}) is mainly to speed up the convergence and to reduce the risk of getting stuck in a local minimum. 
In this paper, for simplicity and generality we focus on the basic version of the t-SNE algorithm based on the simple gradient descent, with the updating equation
\beq \label{ue1}
y_i^{(k+1)}=y_i^{(k)}+hD_i^{(k)},\quad \text{for $i=1,...,n.$}
\eeq
In \cite{maaten2008visualizing} and \cite{van2014accelerating}, the recommended total number of iterations is 1000, while the step size $h$ is initially set as 400 or 800, and is updated at each iteration by an adaptive learning rate scheme of \cite{jacobs1988increased}. 

The standard gradient descent algorithm as in (\ref{ue1}) suffers from a slow convergence rate and even non-convergence in some applications. As an amelioration, \cite{maaten2008visualizing} proposed an \emph{early exaggeration} technique, applied to the initial stages of the optimization, that helps create patterns in the visualization and speed up the convergence. Such a computational strategy has been standard in practical use. In fact, most of the current software implementations of t-SNE are based on an early exaggeration stage followed by an embedding stage that iterates a certain gradient descent algorithm. In our setting, these two stages can be summarized as follows.

\paragraph{Early exaggeration stage.}
For the first $K_0>0$ iterations, the $p_{ij}$'s in the gradient term $D_i^{(k)}$ are multiplied by some exaggeration parameter $\alpha>0$, so the updating equation for this early exaggeration stage becomes
\beq \label{ue2}
y_i^{(k+1)}=y_i^{(k)}+h\sum_{1\le j\le n,j\ne i}(y^{(k)}_j-y^{(k)}_i)S^{(k)}_{ij}(\alpha),\quad i=1,...,n,
\eeq
where $S^{(k)}_{ij}(\alpha) = (\alpha p_{ij}-q^{(k)}_{ij})/(1+\|y_i^{(k)}-y_j^{(k)}\|_2^2)\in \R$, and the factor $4$  in $D_i^{(k)}$ is absorbed into the step size parameter $h$. We refer to this first stage of the t-SNE algorithm as the {\it early exaggeration stage}.

In \cite{maaten2008visualizing}, the authors choose $\alpha=4$ and $K_0=50$ for the early exaggeration stage, whereas later in \cite{van2014accelerating}, it is recommended that $\alpha=12$ and $K_0=250$. In particular, it is empirically observed that, the early exaggeration technique enables t-SNE to find a better global structure in the early stages of the optimization by creating very tight clusters of points that easily move around in the embedding space \citep{van2014accelerating}; this observation is later supported by some pioneering theoretical investigations (see Section \ref{related.work}). Nevertheless, there are interesting questions to be answered concerning (i) the underlying principles and  mechanism behind such a computational strategy, (ii) the limit behavior of the low-dimensional map,
(iii) how sensitive is the performance of t-SNE with respect to the choice of tuning parameters $(\alpha,h,K_0)$, and (iv) how to efficiently determine these  parameters to achieve the best empirical performance.

\begin{figure}[h!]
	\centering
	\includegraphics[angle=0,width=15cm]{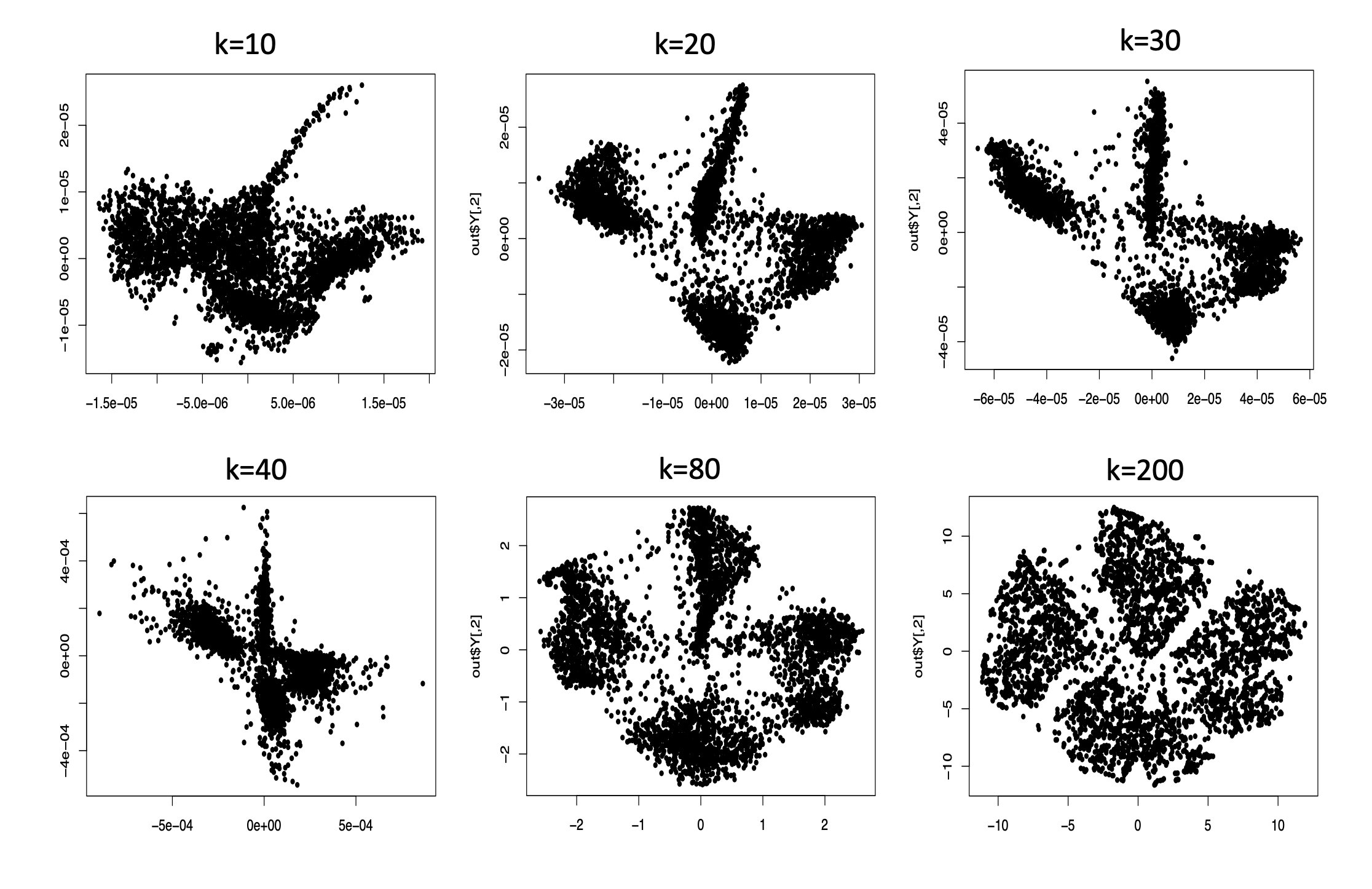}
	\caption{An illustration of the t-SNE iterations that visualize samples from the MNIST dataset (Section \ref{real.sec}). Each sample corresponds an image of handwritten digit ``2,"  ``4,"  ``6," or ``8." The visulaizations are obtained using the \texttt{Rtsne} function in the R package \texttt{Rtsne}, by selecting the exact t-SNE mode (\texttt{theta=0, pca=F}), dropping the momentum terms (\texttt{momentum=0, final\_momentum = 0}), and setting \texttt{perplexity=30} (default), $\alpha=12$ (default), $h=200$ (default) in (\ref{ue2}), and $K_0=40$. The first three plots (top row) correspond to the early exaggeration stage, while the last three plots (bottom row) correspond to the embedding stage.} 
	\label{ees.fig}
\end{figure} 

\paragraph{Embedding stage.}
After the early exaggeration stage, the exaggeration parameter $\alpha$ is dropped and the original iterative algorithm (\ref{ue1}) is carried out till attaining a prespecified number of steps. We refer to this second stage as the {\it embedding stage}. The final output  is a two-dimensional map $\{y_i^{(K_1)}\}_{1\le i\le n}$, commonly treated as a low-dimensional embedding of the original data $\{X_i\}_{1\le i\le n}$, expected to preserve its intrinsic geometric structures. 

In addition to data visualization, t-SNE is sometimes also used as an intermediate step for clustering, signal detection, among many other purposes.  In particular, it has been observed that, when applied to high-dimensional clustered data, t-SNE tends to produce a visualization with more separated clusters, which are often in good agreement with the clusters found by a dedicated clustering algorithm \citep{kobak2019art}. See Figure \ref{ees.fig} for an example of data visualization using such a basic t-SNE algorithm.

\subsection{Main Results and Our Contribution} \label{contr.sec}

A formal theoretical framework is introduced for the analysis of t-SNE that relies on a joint statistical and computational analysis. % based on a double-index asymptotic scheme, that is, the large sample limit where the sample size goes to infinity, and the computational time limit where the number of algorithmic iterations goes to infinity. {\blue In general,  
The key contribution of the present work can be summarized as follows:
\begin{itemize}
	\item We rigorously establish the asymptotic equivalence between the early exaggeration stage and power iterations. Our theory unveils novel properties such as the implicit regularization effect and the necessity of early stopping in the early exaggeration stage for weakly clustered data.
	\item We characterize the behavior of t-SNE iterations at the embedding stage by identifying an amplification phase along with its intercluster repulsion and expansion phenomena, and a stabilization phase of this stage.
	\item We give the theoretical guidance for  initialization and selecting the tuning parameters at both stages in a flexible and data-adaptive manner.
	\item We provide practical advice on applying t-SNE and interpreting the t-SNE visualizations of high-dimensional clustered data.
\end{itemize}
The main results can be explained in more detail from three perspectives.

\paragraph{Early exaggeration stage.} Through a discrete-time analysis (Sections \ref{graph.sec} and \ref{power.sec}), we establish the asymptotic equivalence between the early exaggeration stage and power iterations based on the underlying graph Laplacian associated with the high-dimensional data, providing a spectral-graphical interpretation of the algorithm. We show the implicit spectral clustering mechanism underlying this stage, which explains the adaptivity and flexibility of t-SNE for visualizing clustered data without specifying the number of clusters. Specifically, for the cases where $\{X_i\}_{1\le i\le n}$ are approximately clustered into $R$ groups, we make the key observation that the coordinates of $\{y^{(k)}_i\}_{1\le i\le n}$ converge to the $R$-dimensional Laplacian null space, leading to a limiting embedding where the elements of $\{y^{(k)}_i\}_{1\le i\le n}$ are  well-clustered according to their true cluster membership. On the other hand, through a continuous-time analysis (Section \ref{cont.sec}), we study the underlying gradient flow and uncover an implicit regularization effect depending on the number of iterations.  In particular, our analysis implies that when dealing with noisy and approximately clustered data, one should stop early in the early exaggeration stage to avoid ``overshooting."  These  results justify the empirical observations about the benefits of the early exaggeration technique in creating cluster structures and speeding up the algorithm. 
For more details about comparison with the existing results, see Section \ref{related.work} and the discussions after Corollaries \ref{tsne.cor2} in Section \ref{ee.sec}. 

\begin{figure}[h!]
	\centering
	\includegraphics[angle=0,width=8cm]{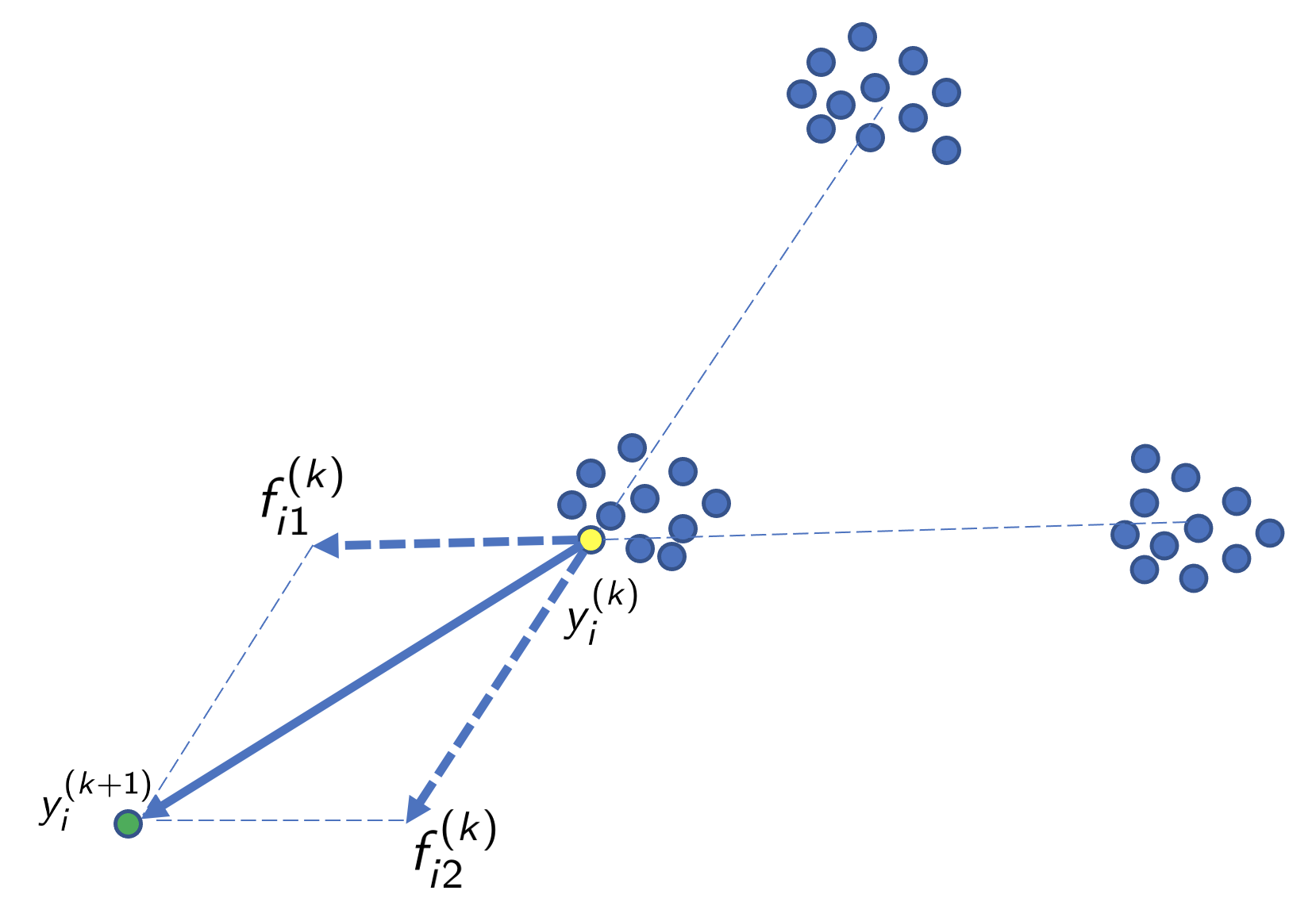}
	\caption{Illustration of the intercluster repulsion where the original data $\{X_i\}_{1\le i\le n}$ have three clusters. The position of $y_i^{(k+1)}$ is jointly determined by $y_i^{(k)}$ and two repulsive forces $f_{i1}^{(k)}$ and $f_{i2}^{(k)}$ pushing $y_i^{(k)}$ away from the other two clusters.} 
	\label{repulsion.fig}
\end{figure}

\paragraph{Embedding stage.} We provide a mechanical interpretation of the algorithm by characterizing the kinematics of the low-dimensional map  at each iteration. Specifically,  in Section \ref{eb.sec} we identify an amplification phase within the embedding stage, featuring the local intercluster repulsion  (Theorem \ref{net.force.thm}) and the global expansive behavior (Theorem \ref{expand.thm}) of $\{y_i^{(k)}\}_{1\le i\le n}$. In the former case, it is shown that the movement of each $y_i^{(k)}$ to $y_i^{(k+1)}$ is jointly determined by the repulsive forces pointing toward $y_i^{(k)}$ from each of the other clusters (Figure \ref{repulsion.fig}), that amounts to increasing spaces between the existing clusters;  in the latter case, it is shown the diameter of $\{y_i^{(k)}\}_{1\le i\le n}$ may strictly increase  after each iteration. We   observe that, following the amplification phase, there is a stabilization phase where $\{y_i^{(k)}\}_{1\le i\le n}$ is locally adjusted to achieve at a finer embedding of  $\{X_i\}_{1\le i\le n}$. 
These results together explain the fast convergence rate and the exceptional empirical performance  of t-SNE for visualizing clustered data. 
The articulation of these phenomena also leads to useful practical guidances. See below and Remark \ref{em.remark} in Section \ref{eb.sec} for more details.

\paragraph{Practical implications.} The general theory brings forth the interpretations of the t-SNE output, and provides theoretical guidance for selecting tuning parameters and for initialization. In Section \ref{examples.sec} we illustrate the general theory on two examples of high-dimensional clustered data, one generated from a Gaussian mixture model, and another from a  noisy nested sphere model. We also analyze in Section \ref{real.sec} a real-world dataset to further demonstrate the practical implications of our theory. In particular, our analysis allows for a wider spectrum of tuning parameters (Figure \ref{delta.fig} and Equation (\ref{tuning})) and initialization procedures than those considered in previous theoretical works \citep{arora2018analysis,linderman2019clustering}. Moreover, our theoretical results support the state-of-art practice \citep{kobak2019art,kobak2021initialization}, but also lead to novel insights (e.g., the first item below) that has been unknown to our knowledge. In the following, we summarize our general advice on applying t-SNE to potentially clustered data:
\begin{itemize}
	\item For weakly clustered data, one may adopt the early exaggeration technique, but needs to stop early (for example, set $K_0= \lfloor(\log n)^2\rfloor$) to avoid overshooting -- failure of stopping early may lead to false clustering; see Figure \ref{early_stop.fig} below for an illustration.
	\item t-SNE visualization based on random initialization and early exaggeration is  reliable in terms of cluster membership but not relative position of clusters. For example, the neighboring clusters in visualization may not be interpreted as neighboring clusters in the original data;  see Figure \ref{initialize.fig} below for an illustration.
	\item Occasionally, false clustering may appear as an artifact of random initialization and intercluster repulsion. Therefore, it is helpful to run t-SNE multiple times to fully assess the effect of random initialization; see Figure \ref{initialize.fig} below for an illustration.
	\item For strongly clustered data, one can speed up the algorithm by replacing the early exaggeration stage by a simple spectral initialization\footnote{An illustration is provided in Figure 4.1 of \cite{linderman2019clustering}.} where $(\yy_1^{(0)}, \yy_2^{(0)})$ are the eigenvectors associated with the smallest two eigenvalues of $\bL(\bP)$. %{\blue see  for an illustration.} {\red If not crucial, we don't have to refer to the figure in \cite{linderman2019clustering}. We can either remove this sentence or create a similar figure in the present paper.}
\end{itemize}

\subsection{Related Work} \label{related.work}	

The impressive empirical performance of t-SNE has recently attracted much theoretical interests. \cite{lee2011shift} investigated the benefits of the so-called shift-invariant similarities used in the stochastic neighbor embedding and its variants. Later on, they further identified two key properties of these visualization methods \citep{lee2014two}. In \cite{shaham2017stochastic}, a large family of methods including t-SNE as a special case were studied and shown to successfully map well-separated disjoint clusters from high dimensions to the real line so as to approximately preserve the clustering.  In \cite{arora2018analysis}, a theoretical framework was developed to formalize the notion of visualizing clustered data, which is used to analyze the early exaggeration stage of t-SNE, and to justify its high visualization quality. \cite{linderman2019clustering} showed that, in the early exaggeration stage of t-SNE, with properly chosen parameters $\alpha$ and $h$, a subset of the two-dimensional map belonging to the same cluster will shrink in diameter, suggesting well-clustered visualization following iterations. We note that connections between the early exaggeration stage and power iterations have been pointed out in \cite{arora2018analysis} and \cite{linderman2019clustering}, but the discussions therein are mostly informal and heuristic. In contrast, we provide rigorous theoretical justification for such a connection, identify its condition and explicate its consequences.
%{\red It might be better to add here a sentence like ``In contrast, we provide rigorous theoretical justificatioons for ..."}
By extending the idea of t-SNE, \cite{im2018stochastic} considered a class of  methods with various loss functions based on the $f$-divergence, and theoretically assessed the performances of these methods based on a neighborhood-level precision-recall analysis. More recently, \cite{zhang2021t} proposed to view t-SNE as a force-based method which generates embeddings by balancing attractive and repulsive forces between data points. In particular,  the limiting behavior of t-SNE was analyzed under a mean-field model where a single homogeneous cluster is present. At the empirical side, the recent works of \cite{kobak2019art} and \cite{kobak2021initialization} summarize the state-of-art practice of using t-SNE to biological data. A comprehensive survey of existing data visualization methods and their properties can be found in \cite{nonato2018multidimensional}.

Despite these pioneering endeavors, the  theoretical understanding of t-SNE is still limited. Many intriguing phenomena and important features that arise commonly in practice have not been well understood or properly explained. Moreover, it remains unclear how to properly interpret the t-SNE visualization and its potential artifacts. These important questions are carefully addressed in the current work for the case of clustered data. Compared to the existing works, the theoretical framework developed in our work leads to identification and explication of novel properties, phenomena, and important practical implications on t-SNE, as summarized at the beginning of Section \ref{contr.sec}.

\subsection{Notation and Organization}

For a vector $\bold{a} = (a_1,...,a_n)^\top \in \mathbb{R}^{n}$, we denote $\text{diag}(a_1,...,a_n)\in\R^{n\times n}$ as the diagonal matrix whose $i$-th diagonal entry is $a_i$, and define the $\ell_p$ norm $\| \bold{a} \|_p = \big(\sum_{i=1}^n a_i^p\big)^{1/p}$.  %We write $a\land b=\min\{a,b\}$ and $a\lor b=\max\{a,b\}$. 
For a matrix $ \bold{A}=(a_{ij})\in \R^{n\times n}$,  we define its Frobenius norm as $\| \bold{A}\|_F = \sqrt{ \sum_{i=1}^{n}\sum_{j=1}^{n} a^2_{ij}}$,  its $\ell_\infty$-norm as $\|\bA\|_{\infty}=\max_{1\le i,j\le n}|a_{ij}|$, 
and its spectral norm as $\| \bold{A} \| =\sup_{\|\bold{x}\|_2\le 1}\|\bold{A}\bold{x}\|_2 $; we also denote
$ \bold{A}_{.i}\in \R^{n}$ as its $i$-th column and $ \bold{A}_{i.}\in \R^{n}$ as its $i$-th row. Let $O(n,k)=\{ \bV\in \R^{n\times k}: \bV^\top \bV={\bf I}_k \}$ be the set of all $n\times k$ orthonormal matrices and $O_n=O(n,n)$, the set of $n$-dimensional orthonormal matrices. For a rank $r$ matrix $\bold{A}\in \R^{n\times n}$ with $1\le r\le n$,  its eigendecomposition is denoted as $\bold{A}=\bU\bGam\bU^\top$ where $\bU\in O(n,r)$ with its columns being the eigenvectors, and $\bGam=\text{diag}(\lambda_1(\bold{A}),\lambda_2(\bold{A}),...,\lambda_r(\bold{A}))$ with $\lambda_{\min}(\bold{A})=\lambda_1(\bold{A})\le...\le\lambda_{n}(\bold{A})=\lambda_{\max}(\bold{A})$ being the ordered eigenvalues of $\bold{A}$. For a smooth function $f(x)$, we denote $\dot{f}(x)=df(x)/dx$ and $\ddot{f}(x)=d^2f(x)/dx^2$.
For any integer $n>0$, we denote the set $[n]=\{1,2,...,n\}$. 
For a finite set $S$, we denote its cardinality as $|S|$. For a subset $S\subseteq \R^n$, we define its diameter $\text{diam}(S)=\sup_{x,y\in S}\|x-y\|_2$. %{\red $\leftarrow$ Double check it is the maximum squared distance, not the maximum distance.}  
For sequences $\{a_n\}$ and $\{b_n\}$, we write $a_n = o(b_n)$ or $a_n\ll b_n$ if $\lim_{n} a_n/b_n =0$, and write $a_n = O(b_n)$, $a_n\lesssim b_n$ or $b_n \gtrsim a_n$ if there exists a constant $C$ such that $a_n \le Cb_n$ for all $n$. We write $a_n\asymp b_n$ if $a_n \lesssim b_n$ and $a_n\gtrsim b_n$. Throughout, $C,C_1,C_2,...$ are universal constants, that can vary from line to line.

The rest of the paper is organized as follows. 
Section \ref{ee.sec} presents the theoretical analysis for the early exaggeration stage of t-SNE. 
Section \ref{eb.sec} analyzes the embedding stage.  The general theory is then applied in Section \ref{examples.sec} to two specific settings of model-based clustered data, one under a Gaussian mixture model and another under a noisy nested sphere model. Analysis of a real-world dataset is presented in Section \ref{real.sec}.  Section \ref{dis.sec} discusses potential applications, extensions and other related problems. Proofs of our main results and supplementary figures are collected in Appendix \ref{proof.sec} to \ref{supp.fig.sec}.

%%%%%%%%%%%%%%%%%%%%%%%%%%%%%%%%%%%%%%%%%%%%%%%%%%%%%%%%	
\section{Analysis of the Early Exaggeration Stage} \label{ee.sec}
%%%%%%%%%%%%%%%%%%%%%%%%%%%%%%%%%%%%%%%%%%%%%%%%%%%%%%%%	

\subsection{Asymptotic Graphical Interpretation and Localization} \label{graph.sec}

We start with a key observation that connects the updating equation (\ref{ue2}) to some graph-related concepts. To this end, we introduce the following definition. 

\begin{definition}[Degree \& Laplacian Operators]
	For a symmetric matrix ${\bf A}=(a_{ij})_{1\le i,j\le n}\in \R^{n\times n}$, define the degree operator $\bD: \R^{n\times n}\to \R^{n\times n}$ by $\bD({\bf A})=\textup{diag}(\sum_{i=1}^n a_{i1}, ..., \sum_{i=1}^n a_{in})$, and the Laplacian operator $\bL: \R^{n\times n}\to \R^{n\times n}$  by $\bL({\bf A})=\bD({\bf A} )-{\bf A}$.
\end{definition}

We define $\bS^{(k)}_\alpha=(S^{(k)}_{ij}(\alpha))_{1\le i,j\le n}\in \R^{n\times n}$ with $S^{(k)}_{ii}(\alpha)\equiv0$ for all $i\in[n]$.  Then we can rewrite the updating equation (\ref{ue2}) using the matrix form as
\beq \label{ue3}
\yy_\ell^{(k+1)}=%\yy_\ell^{(k)}+h\bS^{(k)}_\alpha \yy_\ell^{(k)}-h\bD(\bS^{(k)}_\alpha )\yy^{(k)}_{\ell} =
[{\bf I}_n-h\bL(\bS^{(k)}_\alpha )]\yy_\ell^{(k)},\quad \ell=1,2,
\eeq
where ${\bf I}_n\in \R^{n\times n}$ is the identity matrix, and $\yy^{(k)}_{\ell}\in\R^n$ consists of the $\ell$-th coordinates of $\{y_i^{(k)}\}_{1\le i\le n}$.
As a consequence, for each iteration $k$, if we treat the symmetric matrix $\bS_\alpha^{(k)}$ as the adjacency matrix of a weighted graph $G^{(k)}$ with $n$ nodes that summarizes the pairwise relationships between $n$ data points $\{X_i\}_{1\le i\le n}$, Equation (\ref{ue3}) has an interpretation that links to the Laplacian matrix of such a weighted graph.

To better understand the meaning and the properties of the underlying graph $G^{(k)}$ that evolve over iterations, we take a closer look at its adjacency matrix $\bS_\alpha^{(k)}$. In particular, one should keep in mind that in common applications  of t-SNE, the early exaggeration stage has the following empirical features: (i) moderate or relatively large values of the exaggeration parameter $\alpha$ (default 12 in the R package \texttt{Rtsne}), (ii) local initializations $\{y_i^{(0)}\}_{1\le i\le n}$ around the origin (see Section \ref{tsne.intro.sec}), and (iii) relative 
small diameters $\text{diam}(\{y_i^{(k)}\}_{1\le i\le n})$ over the iterations (Figure \ref{ees.fig}).

Our next result shows that, these empirical features of t-SNE have deep connections to the asymptotic behavior of the evolving underlying graphs and their adjacency matrices $\{\bS_\alpha^{(k)}\}_{k\ge 1}$ in the large sample limit (as $n\to\infty$).

\bet[Asymptotic Graphical Interpretation] \label{graph.prop}
Recall that $\bP=(p_{ij})_{1\le i,j\le n}$ is defined in (\ref{P.def}) and denote $\eta^{(k)}=[\textup{diam}(\{y_i^{(k)}\}_{1\le i\le n})]^2$. Then for any $i,j\in[n]$ with $i\ne j$, and  each $k\ge 1$ such that $\eta^{(k)}<1$, we have
\beq\label{g.inter}
{\bigg|S_{ij}^{(k)}(\alpha)-{\alpha p_{ij}}+\frac{1}{n(n-1)}\bigg|}\le \alpha p_{ij}\eta^{(k)}+\frac{2\eta^{(k)}}{n(n-1)(1-\eta^{(k)})}.
\eeq
Consequently, if we denote ${\bf 1}_n=(1,...,1)^\top\in\R^n$, %{\red it is better to define this as ${\bf 1}_n$ in the notation section and use it here and elsewhere to make the statements more concise. (We also used it in Propositions 2 and 4, equation (2.21), etc.)}  
and $\bH_n=\frac{1}{n(n-1)}({\bf 1}_n{\bf1}_n^\top-{\bf I}_n)$, 
then 
for each $k\ge 1$, as long as  $(\eta^{(k)},\alpha)$ satisfies
\beq \label{cond.graph}
\eta^{(k)}\ll {\frac{\|\bP\|}{n\|\bP\|_\infty}},\quad \alpha\gg \frac{1}{n\|\bP\|},\quad \text{as $n\to\infty$,}
\eeq
we have
\beq \label{graph}
\lim_{n\to\infty}\frac{\|\bS_\alpha^{(k)}-(\alpha \bP-\bH_n)\|}{\|\alpha \bP-\bH_n\|}=0.
\eeq
\eet

The above theorem implies that, for large $n$, as long as the diameter of $\{y_i^{(k)}\}_{1\le i\le n}$ remains sufficiently small and the exaggeration parameter $\alpha$ sufficiently large, the adjacency matrix $\bS_\alpha^{(k)}$ behaves almost like a fixed matrix $\alpha\bP-\bH_n$ across the iterations. In other words, we may treat the updating equation (\ref{ue3}) as an approximately linear equation 
\beq
\yy_\ell^{(k+1)}\approx[{\bf I}_n-h\bL(\alpha\bP-\bH_n)]\yy_\ell^{(k)},\quad \ell=1,2,
\eeq
where the linear operator ${\bf I}_n-h \bL(\alpha\bP-\bH_n)$ only relies on the Laplacian of a fixed weighted graph whose adjacency matrix is given by the scaled and shifted similarity matrix $\alpha\bP-\bH_n$. This essentially opens the door to our key result on the asymptotic equivalence between the early exaggeration stage and power iterations.

Before we formally present such a  result, we need to first point out an important phenomenon concerning the global behavior of the low-dimensional map at the early exaggeration stage. Specifically, we make the following assumptions on the initialization and the tuning parameters $(\alpha,h,k)$:
\begin{displayquote}
 {\bf (I1)} $\{y_i^{(0)}\}_{1\le i\le n}$ satisfies $\min_{\ell\in[2]}\|\yy_\ell^{(0)}\|_2>0$, and $\max_{\ell\in[2]} \|\yy_{\ell}^{(0)}\|_\infty=O(1)$ as $n\to\infty$; and
 
{\bf (T1)} the parameters $(\alpha,h,k)$ satisfy $k(nh\alpha\|\bP\|_\infty +h/n)=O(1)$ as $n\to\infty$.
\end{displayquote}
Intuitively, Condition (I1) says that the initialization $\{y_i^{(0)}\}_{1\le i\le n}$  should not be simply all zeros or unbounded, whereas the condition (T1) -- as a consequence of (\ref{g.inter}) -- essentially requires the cumulative deviations of $h\bL(\bS_\alpha^{(k)})$ from $h\bL(\alpha\bP-\bH_n)$ to be bounded.
Our next result shows that, under these assumptions, the diameter of $\{y_i^{(k)}\}_{1\le i\le n}$ may not increase throughout the iterations, so the embedding remains localized within the initial range.

\bep[Localization] \label{cont.prop}
Suppose ({I1}) and ({T1}) hold. % and $\max_{i\in[n], \ell\in[2]} |y_{i\ell}^{(0)}|=O(1)$. 
We have
\beq \label{cont.ineq}
\textup{diam}(\{y_i^{(k+1)}\}_{1\le i\le n})\le C\max_{\ell\in[2]}\|\yy_\ell^{(0)}\|_\infty, %\max_{i\in[n], \ell\in[2]} |y_{i\ell}^{(0)}|,
\eeq
for some universal constant $C>0$.
%{\red Last sentence not correct. $k$ grows or fixed?}
\eep

The above proposition confirmed the globally localized and non-expansive behavior of $\{y_i^{(k)}\}_{1\le i\le n}$ over the early exaggeration stage observed in practice (Figure \ref{ees.fig}). Concerning Theorem \ref{graph.prop}, it tells us the step-specific condition $\eta^{(k)}\ll {\|\bP\|}/(n\|\bP\|_\infty)$ therein can be generalized to all finite $k$'s as long as the initialization $\{y_i^{(0)}\}_{1\le i\le n}$ is concentrated around 0, that is, $\max_{\ell\in[2]}\|\yy_\ell^{(0)}\|_\infty^2\ll \|\bP\|/(n\|\bP\|_\infty)$.   Furthermore, when $(\alpha,h)$ are chosen such that the step-wise deviation diminishes (i.e., $r_n=nh\alpha\|\bP\|_\infty+h/n\to 0$), Proposition \ref{cont.prop} indicates that (\ref{graph}) may remain true for even larger numbers of iterations as long as $k=O(r_n^{-1})$.

\subsection{Asymptotic Power Iterations, Implicit Spectral Clustering and Early Stopping} \label{power.sec}

With the above graphical interpretation of the updating equation (\ref{ue3}) in mind,  we now present our key result concerning the asymptotic equivalence between the early exaggeration stage and a power method based on  the Laplacian matrix $\bL(\alpha \bP-\bH_n)$. In particular, we make the following assumptions on the initialization and the tuning parameters:
\begin{displayquote}
{\bf (I2)} $\{y_i^{(0)}\}_{1\le i\le n}$ satisfies $\max_{\ell\in[2]} \|\yy_{\ell}^{(0)}\|_\infty^2=o(\|\bP\|/(n\|\bP\|_\infty))$ as $n\to\infty$; and

{\bf (T1.D)} The parameters $(\alpha,h,k)$ satisfy $\alpha\gg (n\|\bP\|)^{-1}$ and $k(nh\alpha\|\bP\|_\infty +h/n)=o(1)$ as $n\to\infty$.
\end{displayquote}
Condition (I2) follows from the discussion subsequent to Proposition \ref{cont.prop}, which along with Condition (T1.D), which is analogous to but stronger than (T1), ensures  the conditions for  Theorem \ref{graph.prop} and Proposition \ref{cont.prop} to hold simultaneously.

\bet[Asymptotic power iterations] \label{power.thm}
Under Conditions (I1) (I2) and (T1.D), we have  (\ref{graph}) and (\ref{cont.ineq})  hold, and so does the asymptotic equivalence
\beq \label{pm.eq}
\lim_{n\to\infty}\frac{\|\yy_\ell^{(k)} -[{\bf I}_n-h\bL(\alpha\bP-\bH_n)]^{k}\yy_\ell^{(0)} \|_2}{\|\yy_\ell^{(0)} \|_2 }=0.
\eeq 
\eet

The above theorem suggests that each  step of the early exaggeration stage may be treated as a power method in the sense that
\beq \label{pm.approx}
\yy_\ell^{(k+1)} \approx [{\bf I}_n-h\bL(\alpha\bP-\bH_n)]^k\yy_\ell^{(0)}.
\eeq
The normalization by $\|\yy_\ell^{(0)} \|_2$ in (\ref{pm.eq}) makes sure the result to be scale-invariant to the  initialization.
It is well-known that, for a fixed matrix $\bG\in \R^{n\times n}$ with 1 as its unique largest eigenvalue in magnitude, the power iteration $\yy^{(k)}=\bG^{k}\yy^{(0)}$ converges to the associated eigenvector  as $k\to \infty$. As a result, when treated as an approximate power method, the early exaggeration stage of t-SNE essentially aims to find the direction of the leading eigenvector(s) of the matrix ${\bf I}_n-h\bL(\alpha\bP-\bH_n)$, which, as will be shown shortly, is actually equivalent to finding the eigenvector(s) associated with the smallest eigenvalue  of the graph Laplacian $\bL(\alpha\bP-\bH_n)$, or the null space of $\bL(\bP)$.

Led by these observations,
our next results concern the limiting behavior of the low-dimensional map $\{y_i^{(k)}\}_{1\le i\le n}$ as the number of iterations $k\to\infty$. Note that any Laplacian matrix has an eigenvalue $0$  associated with a trivial eigenvector $n^{-1/2}{\bf 1}$. Given the affinity (\ref{pm.approx}) between t-SNE and the power method, we start by showing that, the linear operator $[{\bf I}_n-h\bL(\alpha\bP-\bH_n)]^k$ would converge eventually to a projection operator associated with the null space of the Laplacian $\bL(\bP)$.
In particular, we let $R\ge 1$ be the dimension of the null space of the Laplacian $\bL(\bP)\in \R^{n\times n}$; and assume
\begin{displayquote}
{\bf(T2)} the parameters $(\alpha,h)$ satisfies  $\kappa<h\lambda_{R+1}(\bL(\alpha\bP))\le h\lambda_{n}(\bL(\alpha\bP))<1$ for some constant $\kappa\in(0,1)$.
\end{displayquote}
This assumption corresponds to the so-called ``eigengap" condition in the random matrix literature, which gives the signal strength requirements for the recovery of the eigenvalues/eigenvectors, and, in the meantime, the conditions for the tuning parameters.

\bet[Convergence of power iterations] \label{power.thm2}
Let $\bU\in O(n,R)$ such that its columns consist of an orthogonal basis for the null space of $\bL(\bP)$. 
%Let $\bU={\bf 0}\in\R^{n\times 1}$ if $R=1$, and let $\bU\in O(n,R-1)$ if $2\le R\le n$ such that its columns lie in the $R$-dimensional null space and are orthogonal to the  eigenvector $n^{-1/2}{\bf 1}\in\R^n$.
Suppose $kh=o(n)$ and (T2) hold. % ({\red this is eigengap for discrete analysis}). 
Then, we have
\beq
\lim_{k\to\infty}\frac{\|[{\bf I}_n-h\bL(\alpha \bP-\bH_n)]^k\yy-\bU\bU^\top \yy \|_2}{\|\yy\|_2}= 0.%\quad %\lim_{k\to\infty}\frac{\|({\bf I}_n-h\bL(\bP-\bH_n))^k\yy-\bU\bU^\top \yy \|_\infty}{\|\yy\|_2}= 0.
\eeq
\eet

Combining Theorems \ref{power.thm} and \ref{power.thm2}, we know that for sufficiently large $n$ and $k$, the t-SNE iterations $\yy_\ell^{(k)}$ may converge to the projection of the initial vectors $\yy_\ell^{(0)}$ into the null space of the Laplacian $\bL(\bP)$,  that is
\beq \label{y.subspace}
\yy_\ell^{(k)} \approx \bU\bU^\top\yy_\ell^{(0)},\quad \ell\in[2].
\eeq
Now, to better understand the above theorem and its implications on the limiting behavior of t-SNE applied to clustered data, we study the null space of a special class of Laplacian matrices, corresponding to the family of weighted graphs consisting of $R\ge 2$ connected components. In fact,  when the original data  $\{X_i\}_{1\le i\le n}$ are well-clustered and $\tau_i$'s are appropriately chosen, the family of disconnected weighted graphs arise naturally since their adjacency matrices are good approximations of $\bP$ based on these data \citep{balakrishnan2011noise}. We illustrate this point further in Section \ref{examples.sec}. In the following, we say a symmetric adjacency matrix $\bP$ is ``well-conditioned" if its associated weighted graph has $R\ge 2$ connected components.
Our next result characterizes the Laplacian null space corresponding to these disconnected weighted graphs.

\bep[Laplacian null space] \label{eigen}
Suppose $\bA\in \R^{n\times n}$ is symmetric and well conditioned. Then the smallest eigenvalue of the Laplacian $\bL(\bA)$ is $0$ and has multiplicity $R$, and the associated eigen subspace is spanned by $\{\btheta_1,...,\btheta_R\}$ where for each $r\in\{1,...,R\}$,
\[
[\btheta_r]_j = \left\{ \begin{array}{ll}
1/\sqrt{n_r} & \textrm{if  the $j$-th node belongs to the $r$-th component}\\
0 & \textrm{otherwise}
\end{array} \right. ,
\]
and $n_r$ is the number of nodes in the $r$-th connected component. In particular, up to possible permutation of coordinates, any vector $\bu$ in the null space of $\bL(\bA)$ can be expressed as
\beq \label{bu}
\bu=\frac{a_1}{\sqrt{n_1}}\begin{bmatrix}
	{\bf 1}_{n_1}\\ {\bf 0}\\ \vdots \\{\bf 0}
\end{bmatrix}+ \frac{a_2}{\sqrt{n_2}}\begin{bmatrix}
	{\bf 0}\\ {\bf 1}_{n_2}\\ \vdots \\{\bf 0}
\end{bmatrix}+...+ \frac{a_R}{\sqrt{n_R}}\begin{bmatrix}
	{\bf 0}\\ {\bf 0}\\ \vdots \\{\bf 1}_{n_R}
\end{bmatrix},
\eeq
for some $a_1,...,a_R\in \R$. 
\eep

From the above proposition, for a well-conditioned matrix, the components of any $\bu$ in the Laplacian null space has at most $R$ distinct values, and whenever $|\{a_1,...,a_R\}|=R$, the coordinates share the same value if and only if the corresponding nodes fall in the same connected component, i.e., the same cluster.   Combining (\ref{y.subspace}) and (\ref{bu}), one can see that, for strongly clustered data, the output from the early exaggeration stage essentially converges to the eigenvectors associated with the Laplacian null space. This leads to our fourth practical advice at the end of Section \ref{contr.sec}. 

We now generalize the analysis to the setting where the data $\{X_i\}_{1\le i\le n}$ is only weakly clustered in the sense that there exists  a well-conditioned symmetric  matrix $\bP^*$ close to $\bP$ under properly chosen $\{\tau_i\}$, and the underlying graph associated with $\bP$  may not be necessarily disconnected. More specifically, we assume
\begin{displayquote}
 {\bf (T2.D)} there exists a symmetric and well-conditioned matrix $\bP^*\in\R^{n\times n}$ satisfying (T2) and is sufficiently close to $\bP$ in the sense that $kh\alpha \|\bL(\bP^*-\bP)\|=o(1)$.
 \end{displayquote}
%Since no additional condition on the rate of convergence is required for the approximation error $\|\bL(\bP^*-\bP)\|$, 
%Condition (T2.D) essentially requires that $\{X_i\}_{1\le i\le n}$ is approximately clustered with properly chosen $\{\tau_i\}$. Again,
For a given $\bP$ satisfying (T2.D), let $n_r$ with $r\in[R]$ be the size of the $r$-th connected component in the graph associated with $\bP^*$. Our next theorem obtains the implicit spectral clustering and early stopping properties of the early exaggeration stage.

\bet[Implicit clustering and early stopping] \label{tsne.cor2}
Suppose the similarity $\bP$ and the tuning parameters $(\alpha,h,k)$ satisfy (T1.D) and  (T2.D), and the initialization satisfies (I1) and (I2). Then there exists some permutation matrix $O\in\R^{n\times n}$ such that, for $\ell\in[2]$,
\beq \label{y.conv}
\lim_{(k,n)\to\infty}\frac{\|\yy_{\ell}^{(k)}-O\bz_\ell\|_2}{\|\yy_\ell^{(0)}\|_2}=0,
\eeq
where 
\beq
\bz_\ell=(\underbrace{z_{\ell 1},...,z_{\ell 1}}_{n_1}, \underbrace{z_{\ell 2},...,z_{\ell 2}}_{n_2},...,\underbrace{z_{\ell R},...,z_{\ell R}}_{n_R})^\top\in\R^n,
\eeq
and $z_{\ell r}=\btheta_r^\top\yy_\ell^{(0)}/\sqrt{n_r}$ for $r\in[R]$.
%In addition, if the $i$-th node and the $j$-th node in $G$ are connected, then $\lim_{(n,k)\to \infty}\|y_i^{(k)}-y_j^{(k)}\|_2/\|\yy_\ell=0$.
\eet

Theorem \ref{tsne.cor2} describes the limiting behavior of the low-dimensional map $\{y_i^{(k)}\}_{1\le i\le n}$ as $(n,k)\to\infty$, when the original data is approximately clustered. Specifically, elements from $\{y_i^{(k)}\}_{1\le i\le n}$  associated with a connected component of the underlying graph would converge cluster-wise towards a few  points on $\R^2$. 
In particular, Theorem \ref{tsne.cor2} suggests that, at the end of the early exaggeration stage, although the samples belonging to the same underlying cluster tend to be clustered together in the t-SNE embeddings, the cluster centers of the t-SNE embeddings only rely on the initialization, rather than the actual positions of the underlying clusters. Therefore, if \emph{initialized randomly and noninformatively}, the t-SNE embeddings at the end of the early exaggeration stage tend to preserve only the local structures (i.e., the closeness of the samples from the same cluster) but not the global structures (i.e., the relative positions of different clusters) of the original data \citep{kobak2019art,kobak2021initialization}. This observation, as illustrated in  Figure \ref{initialize.fig}, leads to our second practical advice at the end of Section \ref{contr.sec}.
%Using language of information retrieval theory \citep{venna2010information,im2018stochastic,lui2018dimensionality}, this is to say achieves asymptotically perfect recall at the end of the early exaggeration stage, but may not have guaranteed precision if initialized randomly and noninformatively. This lines up with the empirical observation that randomly initialized t-SNE tends to preserve  local structures (i.e., the closeness of samples from the same cluster) rather than global structures (i.e., the relative position of different clusters) of the original data . 

Our theory refines and improves the existing works such as  \cite{linderman2019clustering} and \cite{arora2018analysis} in various aspects. 
Firstly, our theoretical framework formalizes  and explains the asymptotic equivalence between the early exaggeration stage and the power iterations. The theory provides a precise description of the limiting behavior of the low-dimensional map and the theoretical conditions. Secondly, unlike the previous works where only one particular initialization and relatively limited range of tuning parameters were considered, our analysis yields general conditions and allows for more flexible choices of the initialization procedures and tuning parameters. Finally, our analysis unveils the need of stopping early in the early exaggeration stage for weakly clustered data, which is a novel feature.
%(cf. the first practical advice in Section \ref{contr.sec}). 
Specifically, both Conditions (T1.D) and (T2.D) allow $k\to\infty$ but  in a controlled manner -- whenever $\|\bL(\bP^*-\bP)\|\ne 0$, there is a data dependent upper bound on the iteration number
	$$k\ll \frac{1}{h\alpha\|\bL(\bP^*-\bP)\|},$$ 
%whereas when $\|\bL(\bP^*-\bP)\|\approx 0$, such an upper bound will not be effective, so that 
which becomes more stringent for weakly clustered data (i.e., $\|\bL(\bP^*-\bP)\|$ not too small).
Such a phenomenon is also observed empirically (Figure \ref{early_stop.fig}), where  failing to stop early would lead to false clustering. %A deeper reason for early stopping is shown in the next section; see, for example, the discussion after Corollary \ref{es.cor}.  

%\begin{figure}[h!]
%	\centering
%	\includegraphics[angle=0,width=4.9cm]{40b.pdf}
%	\includegraphics[angle=0,width=4.9cm]{70b.pdf}
%	\includegraphics[angle=0,width=4.9cm]{100b.pdf}
%	\caption{An illustration of the t-SNE iterations at the early exaggeration stage, using the same dataset and the same parameters as in Figure \ref{ees.fig}, except for without a limit on $K_0$. The weakening of the cluster patterns suggests the need of an early stopping.} 
%	\label{ees.fig2}
%\end{figure} 

\subsection{Gradient Flow and Implicit Regularization} \label{cont.sec}

For $\ell\in\{1,2\}$, let $\{\tilde{\yy}_\ell^{(k)}\}_{k\ge0}$ be the sequence defined by the  power iterations $\tilde{\yy}_\ell^{(k)}=[{\bf I}_n-h\bL(\alpha \bP-\bH_n)]^k\yy_{\ell}^{(0)}$. Theorem \ref{power.thm} shows that $\{\tilde{\yy}_\ell^{(k)}\}_{k\ge0}$ well approximates the t-SNE iterations $\{\yy_\ell^{(k)}\}_{k\ge 0}$ in the large sample limit. The sequence $\{\tilde{\yy}_\ell^{(k)}\}_{k\ge0}$ admits the updating equation
\beq \label{y.tilde.alg}
\tilde{\yy}_\ell^{(k+1)}=\tilde{\yy}_\ell^{(k)}-h \bL(\alpha\bP-\bH_n)\tilde{\yy}_\ell^{(k)},\quad k\ge 0,
\eeq
with an initial value $\tilde{\yy}_\ell^{(0)}={\yy}_\ell^{(0)}$. Treating Equation (\ref{y.tilde.alg})  as an auxiliary gradient descent algorithm to the original algorithm (\ref{ue3}), a continuous-time analysis can be developed accordingly, which yields interesting insights about the t-SNE iterations $\{\yy_\ell^{(k)}\}_{k\ge0}$. 

We begin by modeling $\{\tilde{\yy}_\ell^{(k)}\}_{k\ge0}$ by a smooth curve $\bY_\ell(t)$ with the Ansatz $\tilde{\yy}_\ell^{(k)}\approx \bY_\ell(kh)$. Define a step function $\yy_{\ell,h}(t)=\tilde{\yy}_\ell^{(k)}$ for $kh\le t<(k+1)h$, and as $h\to 0$, $\yy_{\ell,h}(t)$ approaches $\bY_\ell(t)$ satisfying
\beq\label{GF}
\dot{\bY}_\ell(t)=\bL(\alpha\bP-\bH_n)\bY_\ell(t),
\eeq 
with the initial value $\bY_\ell(0)=\tilde{\yy}_\ell^{(0)}={\yy}_\ell^{(0)}$.
The above first-order differential equation (\ref{GF}) is usually referred as the gradient flow associated with the power iteration sequence $\{\tilde{\yy}_\ell^{(k)}\}_{k\ge0}$, whose limiting behavior can be studied through the step function $\yy_{\ell,h}(t)$. The following theorem provides a non-asymptotic uniform  upper bound on the deviation of $\yy_{\ell,h}(t)$ from $\bY(t)$ over $t\in [0,T]$ and that of $\tilde{\yy}_{\ell}^{(k)}$ from $\bY(kh)$ over $k\le T/h$.

\bep[Gradient flow] \label{cont.thm}
For $\ell=1,2$, and any given $T>0$, we have
\beq
\sup_{t\in[0,T]}\frac{\|{\yy}_{\ell,h}(t)-\bY_{\ell}(t)\|_2}{\|\bY_\ell(t)\|_2}\le Th\|\bL(\alpha\bP-\bH_n)\|^2,
%\quad \sup_{k\le T/h}\frac{\|\tilde{\yy}_{\ell}^{(k)}-\bY_{\ell}(hk)\|_2}{\|\bY_\ell(hk)\|_2}\le Th\|\bL(\alpha\bP-\bH_n)\|^2,
\eeq
where $\yy_{\ell,h}(t)$ is the continuous-time step process of $\{\tilde{\yy}_\ell^{(k)}\}$ generated by (\ref{y.tilde.alg}), and $\bY_\ell(t)$ is the solution to the ordinary differential equation (\ref{GF}). As a consequence, for $t=hk$, if $kh^2\|\bL(\alpha\bP-\bH_n)\|^2\to 0$ as $n\to \infty$, then 	for $\ell\in\{1,2\}$,
\beq \label{affi}
\lim_{(n,k)\to\infty}\frac{\|\tilde{\yy}_{\ell}^{(k)}-\bY_{\ell}(hk)\|_2}{\|\yy_\ell^{(0)}\|_2}=0.
\eeq
\eep

%	{\red The additional constraint $h\|\bL(\alpha\bP)\|^2\to 0$ on the tuning parameter selection ensures the effect of implicit regularization, which may lead to better visualization.}

Combining Theorems \ref{power.thm} and \ref{cont.thm}, we obtain the approximation $\yy_\ell^{(k)}\approx \bY_\ell(kh)$ over a range of $k\ge 0$, for properly chosen parameters $(\alpha,h,k)$ and initialization.
Consequently, the properties of the solution path $\bY_\ell(t)$ may provide important insights on the behavior of the t-SNE iterations at the early exaggeration stage. We start by stating the following proposition concerning the explicit expression of $\bY_{\ell}(t)$.

\bep[Solution path] \label{Yt.thm}
For $\ell\in\{1,2\}$, the first-order linear differential equation (\ref{GF}) with initial value $\bY_\ell(0)=\yy_\ell^{(0)}$ has the unique solution $\bY_\ell(t)=\exp(-t\bL(\alpha\bP-\bH_n))\yy_\ell^{(0)}$, where $\exp(\cdot)$ is the matrix exponential defined as $\exp(\bA)=\sum_{k=0}^\infty \frac{1}{k!}\bA^k$. In particular, suppose $\bL(\bP)$ have the eigendecomposition $\bL(\bP)=\sum_{i=1}^n\lambda_i\bu_i\bu_i^\top$ where $0=\lambda_1\le ...\le \lambda_n$ and $\bu_1=n^{-1/2}{\bf 1}_n$. Then we also have
\beq \label{Yt}
\bY_\ell(t)=(\bu_1^\top\yy_\ell^{(0)})\bu_1+\sum_{i=2}^ne^{-t(\alpha\lambda_i-\frac{1}{n-1})}(\bu_i^\top\yy_\ell^{(0)})\bu_i.
\eeq
\eep

Several important observations about the solution path $\bY_\ell(t)$ can be made. Firstly, by Proposition \ref{Yt.thm}, for $\{\bu_1,...,\bu_m\}$ where $m\in[n]$ is the largest integer such that $\alpha\lambda_m\le \frac{1}{n-1}$, we have
\beq \label{Yt.limit}
\lim_{t\to \infty}\bY_\ell(t) \in \text{span}(\{\bu_1,...,\bu_m\}).
\eeq
This can be treated as a continuous version of the limiting behavior of the power iterations obtained in Theorem \ref{power.thm2}: under the conditions of Theorem \ref{power.thm2}, we have $\alpha\lambda_m\le \frac{1}{n-1}$ for all $m\in[R]$ but $\alpha\lambda_{R+1}> \frac{1}{n-1}$, so that (\ref{Yt.limit}) implies that $\bY_\ell(t)$ converges to the null space of $\bL(\bP)$. Secondly, as long as $t=O(n)$, by orthogonality of $\{\bu_i\}$, we have
\beq
\|\bY_\ell(t)\|_2^2\lesssim  {\sum_{i=1}^ne^{-2t\alpha\lambda_i}(\bu_i^\top\yy_\ell^{(0)})^2}.
\eeq
The right-hand side is monotonically nonincreasing in $t$. Hence, the average distance of the rows in $(\bY_1(hk), \bY_2(hk))$ to the origin remains non-expansive over the iterations and is bounded up to a constant by that of $(\bY_1(0), \bY_2(0))$. This result echos Proposition \ref{cont.prop} based on the discrete-time analysis.

The third and more insightful observation from (\ref{Yt}) is its implications on the finite-time behavior of the original t-SNE sequence $\{\yy_\ell^{(k)}\}_{k\ge1}$, which complements our discrete-time analysis.  Specifically, for finite $t>0$, the coefficient of the $i$-th basis $\bu_i$ in $\bY_\ell(t)$ is proportional to $e^{-t\alpha\lambda_i}$, which is nonincreasing in $\lambda_i$. Consequently,  (\ref{Yt}) implies that, in the early steps of the iterations, the t-SNE algorithm imposes an \emph{implicit regularization} effect on the low-dimensional map $\{y_i^{(k)}\}_{1\le i\le n}$, in the sense that
\beq
\yy_\ell^{(k)}\approx n^{-1}({\bf 1}_n^\top\yy_\ell^{(0)}){\bf 1}_n+\sum_{i=2}^ne^{-kh(\alpha\lambda_i-\frac{1}{n-1})}(\bu_i^\top\yy_\ell^{(0)})\bu_i.
\eeq
Comparing to the limit (\ref{Yt.limit}) or  (\ref{y.subspace}), during the early steps of the iterations, $\yy_\ell^{(k)}$ is regularized as  a conical sum of all the eigenvector basis $\{\bu_i\}_{1\le i\le n}$, with larger weights on the eigenvectors $\bu_i$ corresponding to the smaller eigenvalues of $\bL(\bP)$, and smaller weights on those corresponding to the larger eigenvalues of $\bL(\bP)$. As the iteration goes, the contributions from the less informative eigenvectors with larger eigenvalues $\lambda_i$ such that $\alpha\lambda_i>\frac{1}{n-1}$ decrease exponentially in $k$, whereas the contributions from the more informative eigenvectors  with smaller eigenvalues $\lambda_i$ such that $\alpha\lambda_i<\frac{1}{n-1}$ increase with $k$.

Importantly, the inclusion of all the eigenvectors %and the regularization matrix $\bH_n$ 
helps to better summarize the cluster information in the original data and to avoid convergence to the trivial eigenvector $n^{-1/2}{\bf 1}_n$. Indeed, the convergence (\ref{Yt.limit}) by itself may not lead to a cluster structure in the limit, as in many applications with weakly clustered data, the graph corresponding to $\bP$ may be simply connected under finite samples, so that the null space $\text{span}(\{\bu_1,...,\bu_m\})$ is effectively the one-dimensional space spanned by $n^{-1/2}{\bf 1}$ alone. However, as our next theorem shows, the benefit of the implicit regularization, brought about by stopping early  at the exaggeration stage, can be seen in the creation of desirable clusters in $\{y_i^{(k)}\}_{1\le i\le n}$ for weakly clustered data with approximately block-structured $\bP$.
% embodied in the underlying true graph corresponding to a well-conditioned $\bP^*$. 
In particular, we make the following assumptions analogous to (T1.D) and (T2.D) in the discrete-time analysis.  
\begin{displayquote}
\noindent {\bf (T1.C)} the parameters $(\alpha,h,t)$ satisfy $\alpha \gg [n\lambda_{R+1}(\bL(\bP))]^{-1}$ and $t=o(n)$ as $n\to\infty$;

\noindent {\bf (T2.C)} there exists a symmetric and well-conditioned matrix $\bP^*\in \R^{n\times n}$ such that  $\lambda_{R+1}(\bL(\bP^*))\gg\max\{(t\alpha)^{-1},\|\bL(\bP^*-\bP)\|\}$, and  $t\alpha\|\bL(\bP^*-\bP)\|=o(1)$ as $n\to\infty$.
\end{displayquote}
Similar to the previous conditions, Conditions (T1.C) and (T2.C) concerns the approximate block structure of $\bP$, and ensures sufficient exaggeration and early stopping of the iterations.

\bet[Implicit regularization, clustering and early stopping] \label{es.thm}
Under Conditions (I1), (T1.C) and (T2.C), let $\bU_0\in {O}(n,R)$ such that its columns span the null space of $\bP^*$. Then, we have
\beq
\lim_{n\to\infty}\frac{\|\bY_\ell(t)- \bU_0\bU_0^\top\bY_\ell(t)\|_2}{\|\bY_\ell(0)\|_2}=0,\quad \ell\in\{1,2\},
\eeq
and, for $\bz_{\ell}$  defined in Theorem \ref{tsne.cor2}, there exists a permutation matrix $O\in\R^{n\times n}$ such that
\beq
\lim_{n\to\infty}\frac{\|\bY_\ell(t)-O\bz_\ell\|_2}{\|\bY_{\ell}(0)\|_2}=0,\quad \ell\in\{1,2\}.
\eeq
\eet

An immediate consequence of the above theorem is the following corollary, which arrives at the same conclusion as Theorem \ref{tsne.cor2} through a different route.

\bec \label{es.cor}
Suppose the conditions of Theorems \ref{power.thm} and \ref{es.thm} hold with $t=hk$, and $k\alpha^2 h^2\|\bL(\bP-\bH_n)\|^2\to 0$. Then the conclusion of Theorem \ref{tsne.cor2} holds.
\eec

The above theorems provide a deeper theoretical explanation of the need of  stopping early at the exaggeration stage.
On the one hand, the number of iterations should be sufficiently large so that $\{y_i^{(k)}\}_{1\le i\le n}$  moves away from the initialization and is sufficiently close to a subspace where the underlying cluster information  is properly stored. On the other hand, the iterations should be also stopped early for weakly clustered data to avoid ``overshooting," that is, convergence to the null space of the superficial Laplacian $\bL(\bP)$, which may only include the non-informative trivial eigenvector $n^{-1/2}{\bf 1}_n$ (Figure \ref{early_stop.fig} right).

%%%%%%%%%%%%%%%%%%%%%%%%%%%%%%%%%%%%%%%%%%%%%%%%%%%%%%%%		
\section{Analysis of the Embedding Stage} \label{eb.sec}
%%%%%%%%%%%%%%%%%%%%%%%%%%%%%%%%%%%%%%%%%%%%%%%%%%%%%%%%	

We have shown in Section \ref{ee.sec} that the iterations in the early exaggeration stage essentially create clusters in the  low-dimensional map $\{y_i^{(k)}\}_{1\le i\le n}$, that agree with those underlying $\{X_i\}_{1\le i\le n}$. However, as indicated by Proposition \ref{cont.prop}, so far the low-dimensional map is concentrated and localized around zero, which may not be ideal for visualization purpose. In addition, by Theorem \ref{tsne.cor2}, much information about $\{X_i\}_{1\le i\le n}$ other than its cluster membership are not reflected by the low-dimensional map. In this section, we show that, after transition to the embedding stage, the t-SNE iterations (\ref{ue1}) essentially start by amplifying and refining the existing cluster structures in the low-dimensional map and then aim at a proper embedding of the original data. %In other words, we identify two phases of the embedding stage that have distinct behaviors. 

%\subsection{Phase I: Amplification through Intercluster Repulsion}

\begin{figure}[h!]
	\centering
	\includegraphics[angle=0,width=12cm]{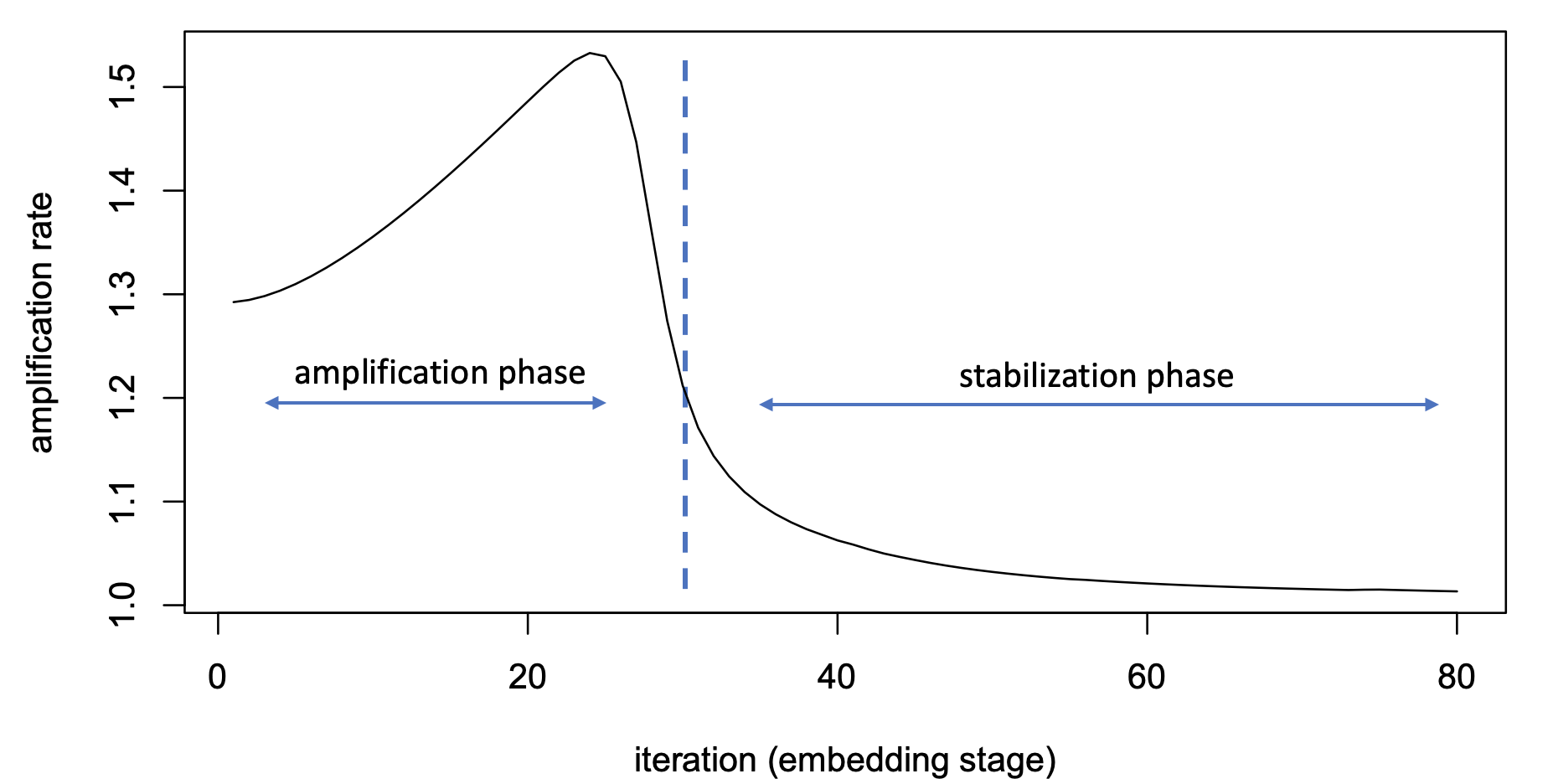}
	\caption{An illustration of the two phase of the embedding stage based on the 1600 MNIST samples described in Section \ref{real.sec}. The iterations are counted from the beginning of the embedding stage, and the amplification rate is the ratio between the diameters of two consecutive embeddings. } 
	\label{expand.fig}
\end{figure}

We show that, starting from the embedding stage, the diameter of the low-dimensional map $\{y_i^{(k)}\}_{1\le i\le n}$ grows fast and they move in clusters as inherited from the early exaggeration stage.   Importantly, over the iterations, the elements of $\{y_i^{(k)}\}_{1\le i\le n}$ belonging to different clusters would in general move away from each other, resulting to an enlarged visualization with more separated clusters.  We refer these iteration steps presenting such a drastically expansive, and intercluster-repulsive behavior of  $\{y_i^{(k)}\}_{1\le i\le n}$ as the {\it amplification phase} of the embedding stage. We also show that, after a certain point, the conditions for the fast expansion phenomenon no longer hold, which likely causes the change of behavior, into a new phase which we refer as the  {\it stabilization phase}. This is in line with the empirical observation (Figure \ref{expand.fig}) that, after a few fast expansive iterations in the amplification phase,  the speed of expansion/amplification gradually reduces towards zero, and in the stabilization phase the diameter only increases very slowly with the iterations.

Recall that the updating equation at the embedding stage is
\beq \label{ue4}
y^{(k+1)}_i=y^{(k)}_i+h'\sum_{j\ne i}S_{ij}^{(k)}(y_{j}^{(k)}-y_i^{(k)}),
\eeq
where $h'$ is the step size that may not be identical to the one in the early exaggeration stage.
To understand the behavior of t-SNE at this stage, we start with the following proposition characterizing the matrix $\bS^{(k)}=(S_{ij}^{(k)})_{1\le i,j\le n}$ over the amplification phase.

\bep \label{S.lem}
For any integer $k$, if $\textup{diam}(\{y_i^{(k)}\}_{1\le i\le n})=o(1)$ as $n\to\infty$, then, for any $i,j\in[n]$ such that $i\ne j$,
\begin{enumerate}
	\item if $\lim_{n\to\infty}n^2p_{ij}=0$, it holds that $S_{ij}^{(k)}= -\frac{1+O(\eta^{(k)})}{n(n-1)}$ as $n\to\infty$; and
	\item if $\lim_{n\to\infty}n^2p_{ij}\ge c$ for some constant $c>0$, it holds that $|S_{ij}^{(k)}|\asymp p_{ij}$ as $n\to\infty$.%; there exists some constant $C>0$ such that for all $p_{ij}\ge  Cn^{-2}$, it holds that  $S_{ij}^{(k)}\asymp p_{ij}$.
\end{enumerate}
\eep

Roughly speaking, Proposition \ref{S.lem} says that over the amplification phase, the matrix $\bS^{(k)}=(S_{ij}^{(k)})_{1\le i,j\le n}$ essentially has two types of entries, determined by the magnitude of the corresponding entries in $\bP$. Specifically, $S_{ij}^{(k)}$ is negative with magnitude $n^{-2}$ if $p_{ij}$ is much smaller than $n^{-2}$, and otherwise $S_{ij}^{(k)}$ has the same magnitude as $p_{ij}$. 
This observation leads to the next theorem, which provides important insights on the updating equation (\ref{ue4}) by partitioning the contributions of $\{y_i^{(k)}\}_{1\le i\le n}$ to an updated $y_i^{(k+1)}$ into a few major components, each corresponding to a distinct cluster in the original data. To this end, we consider again the similarity matrix $\bP$ that is only approximately well-conditioned, as characterized by the following assumption.
\begin{displayquote}
\noindent {\bf (T2.E)} There exists a symmetric and well-conditioned matrix $\bP^*\in\R^{n\times n}$ satisfying (T2.D), $\lim_{n\to\infty}n^2\|\bP-\bP^*\|_\infty=0$, and $\lim_{n\to\infty}\frac{n_r}{n}\to \gamma_r\in(0,1)$ for each $r\in[R]$.
\end{displayquote}
The existence of well-conditioned $\bP^*$ induces an equivalence class on $[n]$ characterizing the underlying cluster membership. Specifically,  for any $i,j\in[n]$, we denote $i\sim j$ if and only if the $i$-th node and the $j$-th node belong to the same graph component. Therefore, we have the partition $[n]=\cup_{r\in[R]}H_r$ for mutually disjoint sets $\{H_r\}_{1\le r\le R}$, with $H_r$ corresponding to the $r$-th equivalence class.

Next, we make assumptions on the initialization and parameters $(\alpha,h,K_0)$ in the early exaggeration stage, where $K_0=K_0(n)\to\infty$ is the total number of iterations in that stage. Specifically, we assume
\begin{displayquote}
\noindent {\bf (I3)} the initialization is chosen such that $\|\yy_{1}^{(0)}\|_2\asymp \|\yy_{2}^{(0)}\|_2$, $\max_{\ell\in[2]}\|\yy_{\ell}^{(0)}\|_\infty=o(n^{-1/2})$ as $n\to\infty$, and there exists some constant $C>1$ such that $\bz_{\ell}$ defined in Theorem \ref{tsne.cor2} satisfies $C^{-1}\le{n|z_{\ell i}-z_{\ell j}|}/{\|\yy_\ell^{(0)}\|_2}\le C$ for any $i,j\in[R]$ such that $i\ne j$, and $\ell\in\{1,2\}$; and

\noindent {\bf (T1.E)} the parameters $(\alpha,h,K_0)$ in (\ref{ue2}) satisfy (T1.D), and, for $R_n=(1-\kappa)^{K_0}+hK_0[(\alpha n \|\bP\|_\infty+1/n)\cdot\max_{\ell\in[2]}\|\yy_{\ell}^{(0)}\|_\infty^2+\alpha\|\bL(\bP^*-\bP)\|]$, we have $nR_n(1+n^2\|\bP^*\|_{\infty})=o(1)$ as $n\to\infty$.
\end{displayquote}
Condition (I3) is mild as it can be satisfied with high probability by a straightforward random initialization procedure, to be presented shortly.
Condition (T1.E) is analogous to but slightly stronger than (T1.D), by requiring a smaller cumulative approximation error $R_n$ between $\bL(\bS_\alpha^{(k)})$ and $\bL(\alpha\bP^*-\bH_n)$, that is, more distinct clusters in $\{X_i\}_{1\le i\le n}$.

Finally, for the parameters $(h',K_1)$ in embedding stage, where $K_1=K_1(n)$ is the number of iterations within the amplification phase, we make the following assumption that controls the cumulative approximation error in $\bS^{(k)}$ as suggested by Proposition \ref{S.lem}.
\begin{displayquote}
\noindent {\bf (T3.E)} $\text{diam}(\{y_i^{(K_0+K_1)}\}_{1\le i\le n})=o(1)$, and the parameter $h'$ in (\ref{ue4}) satisfies $K_1 h' (n\|\bP^*\|_\infty+1/n)=O(1)$ as $n\to\infty$.
\end{displayquote}

\bet[Intercluster repulsion] \label{net.force.thm}
Under Conditions (T1.E) (T2.E) (T3.E) and  (I3), for each $K_0\le k\le K_0+K_1$ and any $i\in[n]$, we have
\beq \label{decomp}
y^{(k+1)}_i=y^{(k)}_i+\sum_{{r\in[R]\setminus r_0}} f^{(k)}_{ir}+\epsilon_i^{(k)},
\eeq
where $r_0\in[R]$ such that $i\in H_{r_0}$, $\lim_{n\to\infty}\|\epsilon_i^{(k)}\|_2/\|f^{(k)}_{ir}\|_2=0$ for all $r\in[R]\setminus r_0$, and
\[
f_{ir}^{(k)}=\frac{h'|H_r|}{n(n-1)}\bigg(y_i^{(k)}-\frac{1}{|H_r|}\sum_{j\in H_r}y_j^{(k)}\bigg)\in\R^2.
\]
In addition, we have 
\beq 
%\sup_{k_0\le k \le k_0+k_1} \max_{r\in[R]}\big[\textup{diam}(\{y_i^{(k)}\}_{i\in H_r})\big]\ll 1/n,
\sup_{K_0\le k \le K_0+K_1}\max_{(i,j):i\sim j}\|y_i^{(k)}-y_j^{(k)}\|_2\ll n^{-1}(\|\yy_1^{(0)}\|_2+\|\yy_2^{(0)}\|_2),
\eeq
and 
\beq 
%\inf_{k_0\le k \le k_0+k_1} \min_{\substack{r_1,r_2\in[R]\\ r_1\ne r_2}} \big[\textup{dist}(H_{r_1}, H_{r_2})\big]\gtrsim 1/n.
\inf_{K_0\le k \le K_0+K_1}\min_{(i,j):i\nsim j}\|y_i^{(k)}-y_j^{(k)}\|_2\gtrsim n^{-1}(\|\yy_1^{(0)}\|_2+\|\yy_2^{(0)}\|_2).
\eeq
\eet

A few remarks about the above theorem are in order. Firstly, Conditions (T1.E) (T2.E) and (I3) concerning the initialization, parameter selection and the number of iterations in the early exaggerations are not only compatible but also sufficient for the previous results, including Theorems \ref{power.thm} and \ref{tsne.cor2}.	This suggests the above intercluster repulsive  phenomenon at the embedding stage actually relies on the properties of the outputs from the early exaggeration stage, again yielding the necessity of the early exaggeration, or equivalent techniques. 
Secondly, as indicated by the next theorem, Condition (I3)  on the initialization can be satisfied by the following simple local random initialization procedure.

\bet[Random initialization] \label{ini.thm}
For any sequence $\sigma_n\to 0$ as $n\to\infty$, let $\yy_{\ell}^{(0)}=\sigma_n \bg_{\ell}/\|\bg_{\ell}\|_2$, where $\bg_{\ell}\in\R^n$ for $\ell\in[2]$ is independently generated from a standard multivariate normal distribution. Then $\{y_i^{(0)}\}_{1\le i\le n}$ satisfies Condition (I3) with probability at least $1-\delta$ for some sufficiently small constant $\delta>0$.
\eet

Thirdly, the above theorem provides a precise characterization of the kinematics of each $y_i^{k}$ during the iterations, and its reliance on the data points $\{y_i^{(k-1)}\}$ in the previous step, as well as the cluster structure inherited from the early exaggeration stage.  Specifically, $f^{(k)}_{ir}$ summarizes the contributions from the points $\{y_i^{(k)}\}_{i\in H_r}$ in the $r$-th cluster to the new point $y^{(k+1)}_i$. 
The theorem implies that, at the amplification phase, the behavior of $\{y_i^{(k)}\}_{1\le i\le n}$ is mainly driven by the relative positions of the $R$ clusters  produced in the early exaggeration stage: for each point, a vector sum of the \emph{repulsive forces} coming from all the other clusters at their current positions  determine the direction and distance of its movement of each point in this iteration (Figure \ref{repulsion.fig}). As a consequence, after each iteration, the diameter of $\{y_i^{(k)}\}_{1\le i\le n}$ would increase, till the end of the amplification phase, that is, when Condition (T3.E), or more specifically, $\text{diam}(\{y_i^{(k)}\}_{1\le i\le n})=o(1)$ no longer holds.
This process improves the visualization quality by making the clusters more distinct and separated (Figure \ref{ees.fig} with $k=40$ and 80).

Our next result confirms the intuition that the diameter of  $\{y_i^{(k)}\}_{1\le i\le n}$ is bound to increase after each iteration in the amplification phase of the embedding stage.

\bet[Expansion] \label{expand.thm}
Suppose the conditions of Theorem \ref{net.force.thm} hold. If in addition $\|\bP^*\|_\infty\lesssim n^{-2}$, then for any $k\in \{K_0,K_0+1,...,K_1\}$, we have
\beq
\textup{diam}(\{y_i^{(k+1)}\}_{1\le i\le n})>\textup{diam}(\{y_i^{(k)}\}_{1\le i\le n}),
\eeq
where $\textup{diam}(\{y_i^{(k+1)}\}_{1\le i\le n})-\textup{diam}(\{y_i^{(k)}\}_{1\le i\le n})\gtrsim \frac{h'}{n^2}\min_{\ell=1,2}\|\yy_{\ell}^{(0)}\|_2$.
\eet

Once the diameter of $\{y_i^{(k)}\}_{1\le i\le n}$ exceeds certain threshold, that is, when $\text{diam}(\{y_i^{(k)}\}_{1\le i\le n})$ is at least of constant order, we arrive at the final stabilization phase. In this phase, the condition of Proposition \ref{S.lem} is violated, and, unlike what is claimed in part one of Proposition \ref{S.lem},  the entries of the matrix $\bS^{(k)}$ corresponding to the smaller entries in $\bP$, that is, $p_{ij}$'s with $p_{ij}\ll n^{-2}$,  no longer remain an almost constant value $1/n(n-1)$. In particular,  the sign of $S_{ij}^{(k)}$ would generally rely on the relative magnitudes between $p_{ij}$ and $q_{ij}$.

We rewrite (\ref{ue4}) as
\beq 
y^{(k+1)}_i=y^{(k)}_i+h'\sum_{j\ne i}\frac{p_{ij}-q_{ij}^{(k)}}{1+d_{ij}^{(k)}}(y_j^{(k)}-y_i^{(k)}).
\eeq
In the stabilization phase, the new position for $y^{(k+1)}_i$ is determined by the starting point $y_i^{(k)}$, and the averaged contributions from each of the other data points $\{y_j^{(k)}\}_{j\ne i}$. The contribution from $y_j^{(k)}$ to $y_i^{(k+1)}$ is either in or against the direction of $(y_j^{(k)}-y_i^{(k)})$, depending on  $\text{sign}(p_{ij}-q_{ij})$. If $\text{sign}(p_{ij}-q_{ij})=-1$, or, the similarity between $y_i^{(k)}$ and $y_j^{(k)}$  as measured by $q_{ij}$ is greater than the similarity between $X_i$ and $X_j$ as measured by $p_{ij}$,  the contribution from $y_j^{(k)}$ to $y_i^{(k+1)}$ is in the direction of $y_i^{(k)}-y_j^{(k)}$, resulting to a repulsive force that enlarges the distance between $y_i^{(k+1)}$ and $y_j^{(k+1)}$ after the iteration. Similarly, if $\text{sign}(p_{ij}-q_{ij})=1$, it means the similarity between $y_i^{(k)}$ and $y_j^{(k)}$  is smaller than their counterparts in $\{X_i\}_{1\le i\le n}$, so the contribution $y_j^{(k)}$ to $y_i^{(k+1)}$ is in the opposite direction $y_j^{(k)}-y_i^{(k)}$, resulting to an attractive force that reduces the distance between  $y_i^{(k+1)}$ and $y_j^{(k+1)}$ after the iteration. The iterations over the stabilization phase aim to locally adjust the relative positions of the low-dimensional map to make the final visualization more reliable and faithful.

\begin{remark} \label{em.remark}
{\rm In practice, expansion and repulsion effects help make clusters identified from the early exaggeration step more salient in the final visualization, and possibly more informative in terms of the local structures within the clusters. This is especially helpful if two clusters are positioned too close to each other at the end of the early exaggeration stage, as an artifact of the random initialization (e.g., the middle column of Figure \ref{early_stop.fig}).
Moreover, the intercluster repulsion phenomenon explains the occasional appearance of false clusters in the t-SNE visualization \citep{kobak2021initialization}. Specifically, our theory indicates that false clustering may appear due to an incidental combination of overlapped clusters from the early exaggeration stage with random initialization, and the intercluster repulsion from the embedding stage (Figures \ref{early_stop.fig} and \ref{initialize.fig}). This leads to our third  general advice on practice at the end of Section \ref{contr.sec}. 
}\end{remark}

%%%%%%%%%%%%%%%%%%%%%%%%%%%%%%%%%%%%%%%%%%%%%%%%%%%%%%%%	
\section{Application I: Visualizing Model-Based Clustered Data} \label{examples.sec}
%%%%%%%%%%%%%%%%%%%%%%%%%%%%%%%%%%%%%%%%%%%%%%%%%%%%%%%%	

In the previous sections, we established the theoretical properties for the basic t-SNE algorithm under general conditions on the parameters $(\alpha,h,h',K_1,K_2)$, the initialization, and the similarity matrix $\bP$ constructed from the original data. In this section, we apply our general theory in two concrete examples of clustered data, one generated from a Gaussian mixture model and another from a noisy nested sphere model.

\subsection{Gaussian Mixture Model}

Consider the Gaussian mixture model 
\beq
X_i|z_i=r \sim N(\mu_r,\Sigma),\quad z_i\stackrel{i.i.d.}{\sim} \text{Multinomial}(\pi_1,...,\pi_R),\quad\text{for $i\in[n]$,}
\eeq
where $\mu_r\in\R^p$ and $\sum_{r=1}^R \pi_r=1$. We make the following assumptions.
\begin{displayquote}
\noindent {\bf (C1)} The mixing proportions $\{\pi_r\}_{1\le r\le R}$ satisfy  $\min_r\pi_r\ge c>0$.

\noindent {\bf (C2)} There exists some large constant $C'>0$ such that $\rho^2=\min_{1\le j\ne k\le R}\|\mu_j-\mu_k\|^2_2\ge C'\max\{p,\log n\}$.

\noindent {\bf (C3)} There exists some constant $C>0$ such that the population covariance matrix $\Sigma\in\R^{p\times p}$ satisfies  $C^{-1}\le \lambda_1(\Sigma)\le \lambda_p(\Sigma)\le C$ and $\text{tr}(\Sigma)/p\le C$.
\end{displayquote}
Under the above Gaussian mixture model, we  obtain the following corollary that provides the  conditions for the theoretical results presented in the previous sections. %The result implies the range of appropriate tuning parameters for visualizing the model-generated  data   with t-SNE.

\bec \label{gmm.cor}
Suppose Conditions (C1) (C2) and (C3) hold, and $\tau_i^2\asymp \max\{p,\log n\}$. If $\alpha\gg 1$,  $K_0h=o(n)$, $h\alpha\asymp n$, $1\ll K_0\ll \exp\{\frac{\rho^2}{\max\{p,\log n\}}\}$ and $K_0h\alpha\sigma_n^2\log n=o(n^2)$, then Conditions (T1.D) and (T2.D) hold.
If in addition $\log n\ll K_0\ll n^{-1}\exp\{\frac{\rho^2}{\max\{p,\log n\}}\}$, $K_0h\alpha\sigma_n^2\log n=o(n)$, and $K_1h'=O(n)$, then Conditions (T1.E) (T2.E) and (T3.E) hold.
\eec

As a consequence, suitable choices of the tuning parameters $(\alpha,h,h',K_0,K_1)$ under the Gaussian mixture model can be determined efficiently. For example, if $\rho^2\gtrsim \log n\cdot \max\{p,\log n\}$, one could choose $K_0=\lfloor (\log n)^2\rfloor$, $\sigma_n=(\log n)^{-2}$, $h=h'=n^{\delta}$ and $\alpha=n^{1-\delta}$ for any constant $\delta\in(0,1)$. By Corollary \ref{gmm.cor}, Conditions (T1.D) and (T2.D) hold, so the conclusions of Theorem \ref{tsne.cor2} follows for $k=K_0$; meanwhile, Conditions (T1.E) (T2.E) and (T3.E) also hold, so the conclusions of Theorem \ref{net.force.thm} hold for each $K_1$ with $K_0\le K_1\le n^{1-\delta}$. Note that the above results apply to both low-dimensional settings where $p=o(n)$ and high-dimensional settings where $p\gtrsim n$. 

To demonstrate the effectiveness of the theoretical guidance, we generate $n=1500$ samples of dimension $p=100$ from a Gaussian mixture model with $r=6, \rho^2=p, \Sigma={\bf I}_p$, and the cluster proportion vector $(0.1, 0.1, 0.1, 0.15, 0.25, 0.3)$. We use the above tuning parameters with various $\delta\in\{1/2, 1/3\}$ and  \texttt{perplexity=30} (default). The t-SNE embeddings at the  end of the early exaggeration stage $k=K_0=\lfloor (\log n)^2\rfloor=53$ and at $k=1000$ are included in Figure \ref{simu.fig} below and Figure \ref{simu-supp.fig} in Appendix \ref{supp.fig.sec}, confirming the theoretical predictions. Moreover, Figure \ref{simu-supp-rho.fig} in Appendix \ref{supp.fig.sec} shows that when the above separation condition (C2) is slightly violated (e.g., $\rho^2= p^{4/5}$), t-SNE is still able to visualize clusters, which demonstrates the robustness of t-SNE with respect to the separation condition.

 	\begin{figure}[h!]
 	\centering
 	\includegraphics[angle=0,width=12cm]{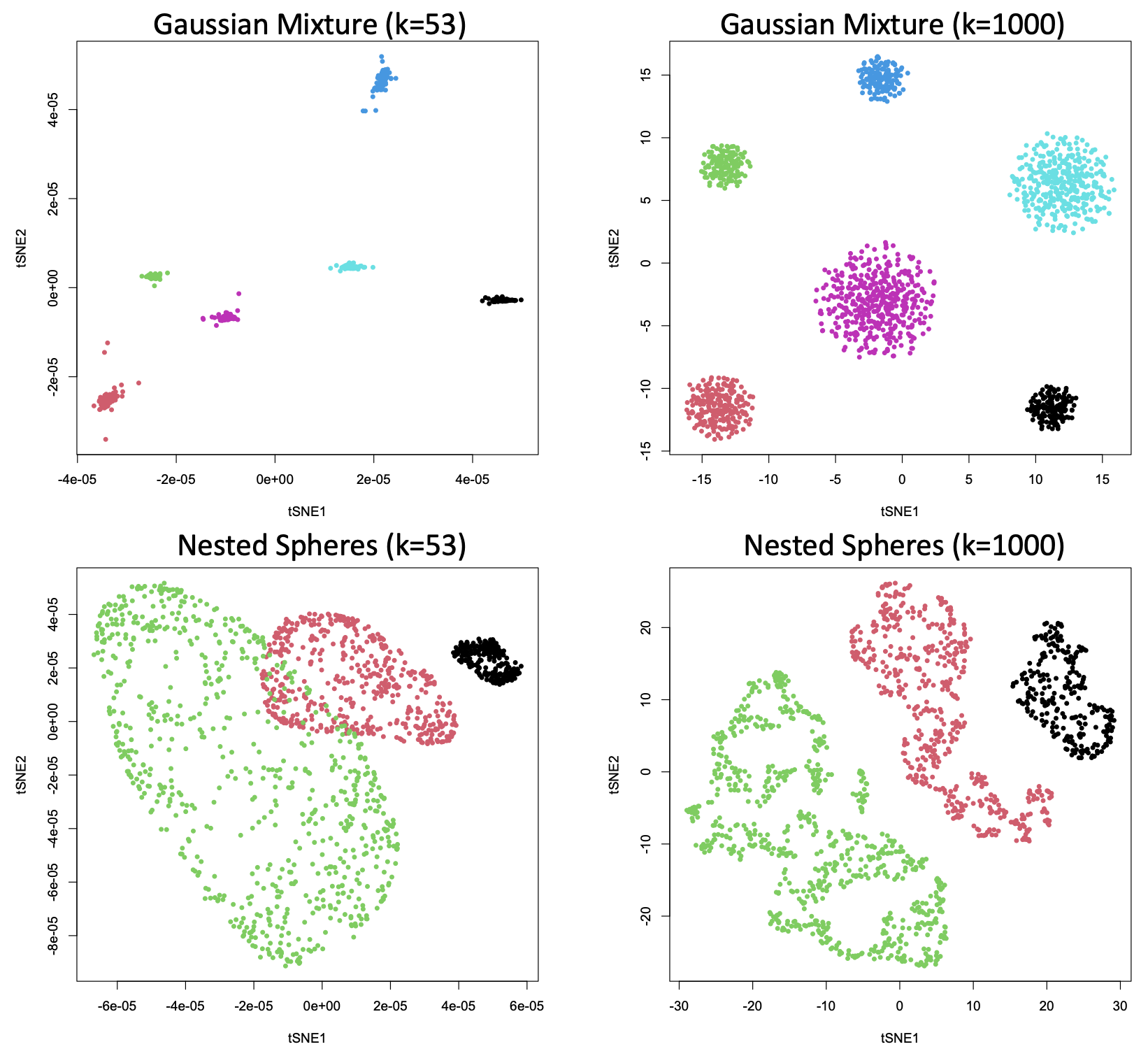}
 	\caption{t-SNE visualizations of the model-generated samples as described in Section \ref{examples.sec}, using the theory-guided tuning parameters with $\delta=1/3$ (see Figure \ref{simu-supp.fig} for similar results with $\delta=1/2$). The left column shows outputs from the early exaggeration stage, whereas the right column are the corresponding final embeddings.} 
 	\label{simu.fig}
 \end{figure}

\begin{remark}
{\rm \cite{arora2018analysis} analyzed the early exaggeration stage of t-SNE based on a slightly different theoretical framework under the  Gaussian mixture model with a mean separation condition $\rho\gtrsim p^{1/4}$, and under the mixture model of log-concave distributions with a separation condition $\rho\gtrsim p^{5/12}$. Compared to these results, our  separation condition $\rho\gtrsim p^{1/2}$ in (C2) is strong, and it is unclear to us if such a restriction is intrinsic to our theoretical framework or an artifact from our proof strategy. Nevertheless, nailing down the sharp information threshold for t-SNE visualization is an important and fundamental problem -- we plan to have a more systematic treatment of this in a subsequent work.
}\end{remark}

\subsection{Noisy Nested Sphere Model}

Consider the model of nested spheres with radial noise \citep{amini2021concentration}, where for $i\in[n]$, we have 
\beq \label{nns.model}
X_i=\mu_i+\frac{\mu_i}{\|\mu_i\|_2}\xi_i,\quad \xi_i\stackrel{i.i.d.}{\sim} N(0,\sigma^2)
\eeq
and 
\beq
\mu_i|z_i=r\sim P_k,\quad z_i\stackrel{i.i.d.}{\sim}\text{Multinomial}(\pi_1,...,\pi_R),
\eeq
with $\sum_{r=1}^R \pi_r=1$ and $\{P_k\}$ being uniform distributions on nested spheres in $\R^p$ of various radii $\rho_{\min}=\rho_1<\rho_2<...<\rho_R=\rho_{\max}$. 
We make the following assumptions concerning the separation distances between the underlying nested spheres.
\begin{displayquote}
\noindent {\bf(C4)} There exists some $\gamma$ such that $\max\{n^{-1}, \sigma^2\rho^{-2}_{\min}\}\log n\ll \gamma\ll 1$ and $\max_{r\in [R-1]}\frac{\rho_{r}}{\rho_{r+1}}\ll 1-C\sqrt{\gamma\log\gamma}$ 
%{\red $\leftarrow$ It is better to take logarithm on both sizes and simplify the condition in terms of $\gamma$. Some of the conditions in the following corollary should also be made cleaner.} 
for some sufficiently large constant $C>0$.

\noindent {\bf (C5)} There exists some small constant $c>0$ such that $c\min|\rho_{r+1}-\rho_r|\ge \sigma\sqrt{\log n}$.
\end{displayquote}
In Condition (C4), the separation distance is characterized by the ratio $\rho_r/\rho_{r+1}$ whereas in Condition (C5) the distance is characterized by the difference $\rho_{r+1}-\rho_r$.
The following corollary provides a sufficient condition for the results presented in the previous sections.% under the noisy nested sphere model.

\bec \label{nns.cor}
Suppose Assumptions (C1) (C4) and (C5) hold, and $\tau_i^2\asymp\gamma\rho_{z_i}^2$. If  $K_0h=o(n)$,  $\alpha h=O(\gamma n)$, $h\alpha \lambda_{R+1}(\bL(\bP^*))\ge \kappa$ for some constant $\kappa\in(0,1)$, $K_0\gg 1$,  $K_0h(\alpha/\gamma +1)\sigma_n^2\log n=o(n^2)$, and $\log \frac{K_0h\alpha}{n}\ll \gamma^{-1}(1-\max_{r\in[R-1]}\frac{\rho_{r}}{\rho_{r+1}})^2+\log \gamma$, then Conditions (T1.D) and (T2.D) hold.
If in addition $K_0\gg \log n$, $K_0h(\alpha /\gamma+1)\sigma_n^2\log n=o(n)$, $\log K_0h\alpha\ll \gamma^{-1}(1-\max_{r\in[R-1]}\frac{\rho_{r}}{\rho_{r+1}})^2+\log \gamma$, and $K_1h'=O( \gamma n)$, then Conditions (T1.E) (T2.E) and (T3.E) hold.
\eec

Again, suitable choices of the tuning parameters $(\alpha,h,h',K_0,K_1)$ under the noisy nested sphere model can be determined efficiently.
For example, let's consider the case where $\rho_{r+1}-\rho_r=\Delta$ for all $r\in[R-1]$. Specifically, suppose there exists some small constant $c>0$ such that $\Delta\ge c\rho_R$, and that $\gamma=c(\log n)^{-1}$ satisfies (C4) and $\lambda_{R+1}(\bL(\bP^*))\gtrsim \frac{1}{\gamma n}$ in probability. Then, by Corollary \ref{nns.cor}, the desired visualization properties  such as  those in Theorems \ref{tsne.cor2} and \ref{net.force.thm} would hold with high probability, as long as  we choose $K_0=\lfloor (\log n)^2\rfloor$,  $K_1\le n^{1-\delta}/\log n$, $\sigma_n=(\log n)^{-2}$, $h=h'=n^{\delta}$ and $\alpha=\gamma n^{1-\delta}$ for any constant $\delta\in(0,1)$.  Figures \ref{simu.fig} and  \ref{simu-supp.fig} in the Appendix show the t-SNE embeddings of $n=1500$ samples of dimension $p=50$, at the  end of the early exaggeration stage $k=K_0=\lfloor (\log n)^2\rfloor=53$ and at $k=1000$, generated from Model (\ref{nns.model}) with $r=3, \sigma=1$, $(\rho_1,\rho_2,\rho_3)=(10,25, 50)$ and  cluster proportion $(0.17, 0.33,0.5)$. For the tuning parameters, the above analytical values with $\gamma=0.5$ and $\delta\in\{1/3, 1/2\}$ are used. As a result, clusters of three nested spheres are visible in all t-SNE embeddings, confirming our theoretical predictions.

%%%%%%%%%%%%%%%%%%%%%%%5
\section{Application II: Visualizing Real-World Clustered Data} \label{real.sec}
%%%%%%%%%%%%%%%%%%%%%%%%%

Finally, we demonstrate our theory by applying t-SNE to the MNIST\footnote{http://yann.lecun.com/exdb/mnist/} dataset, which contains images of hand-written digits. Specifically, we focus on $n=4N=1600$ images of hand-written digits ``2," ``4," ``6" and ``8,"  with each digit having $N=400$ images. Each image contains $28\times28$ pixels and was treated as a $784$-dimensional vector.  
Based on our theoretical analysis, we set the tuning parameters 
\beq \label{tuning}
\alpha=n^{1-\delta},\quad h=h' = n^{\delta}, \quad K_0=\lfloor (\log n)^2 \rfloor
\eeq
 with $\delta=2/3$. Again, we use the default perplexity (=30), leading to an approximate block matrix $\bP$, with block structure corresponding to the cluster membership (Figure \ref{block-supp.fig}).

\begin{figure}[h!]
	\centering
	\includegraphics[angle=0,width=15cm]{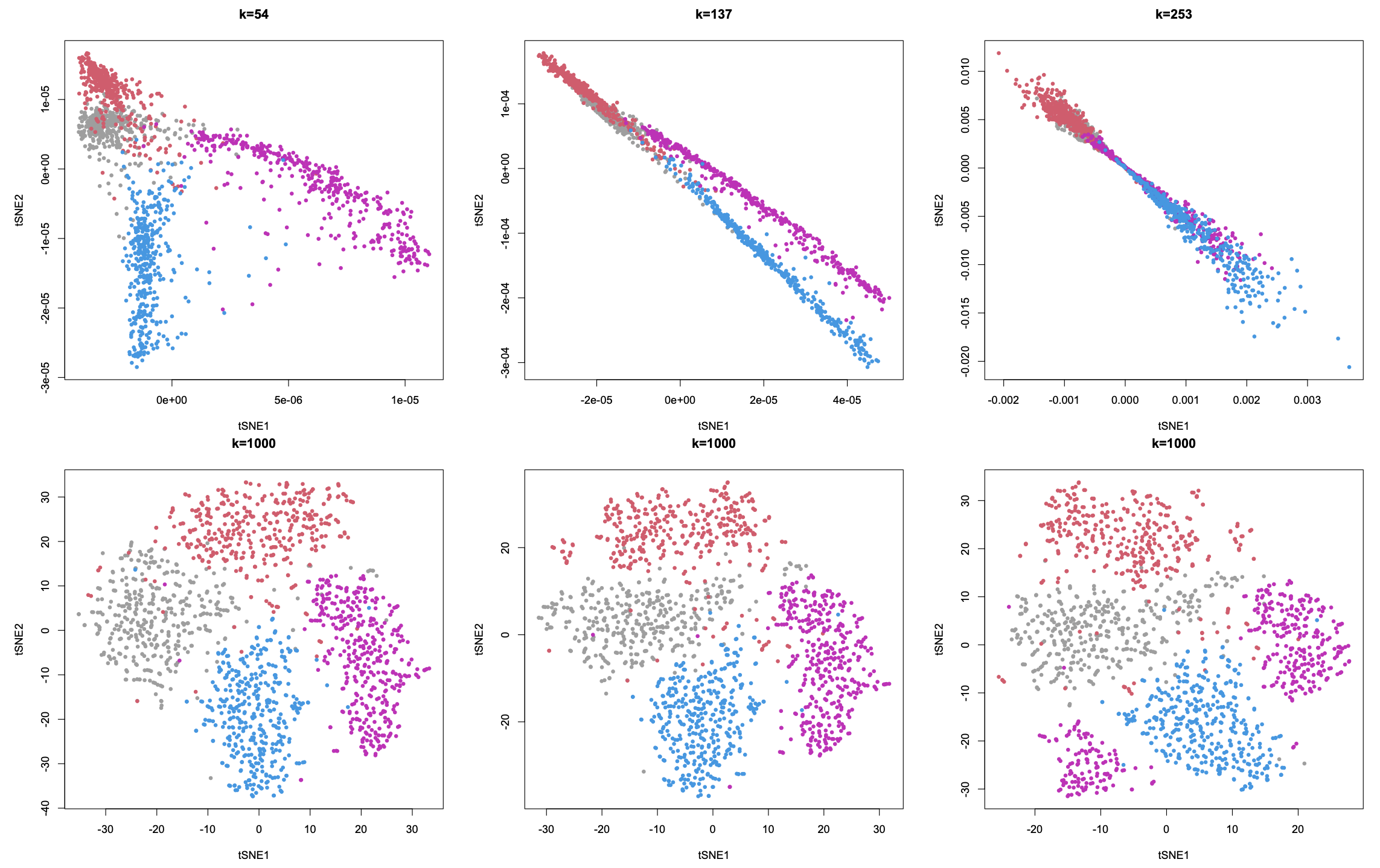}
	\caption{Illustration of t-SNE embeddings of 1600 MNIST samples at the end of embedding stage (bottom row), and their corresponding outputs from the early exaggeration stage (top row). Different columns have identical initializations and tuning parameters, but distinct number of iterations for the early exaggeration stage. The colors of the dots indicate the underlying four clusters.} 
	\label{early_stop.fig}
\end{figure} 

To demonstrate the necessity of stopping early at the early exaggeration stage, in Figure \ref{early_stop.fig}, we show the t-SNE embeddings at the end of embedding stage (bottom row), and their corresponding outputs from the early exaggeration stage (top row). Different columns have identical initialization and tuning parameters, but distinct  numbers of iterations for the early exaggeration stage, namely $K_0= \lfloor (\log n)^2\rfloor=54$ (left), $K_0=  \lfloor n^{2/3}\rfloor = 137$ (middle), and $ K_0=  \lfloor n^{3/4}\rfloor = 253$ (right). Comparing the top three plots in Figure \ref{early_stop.fig}, we can clearly see that when $K_0$ far exceeds our theory-guided value $\lfloor (\log n)^2\rfloor$, the cluster patterns is no longer visible, which, in the case of $K_0=253$, led to false clustering in the final visualization.
Moreover, the middle column of Figure \ref{early_stop.fig} also demonstrates the importance of the embedding stage, especially its underlying intercluster repulsion and  expansion effects, as to making the cluster patterns more salient in the final visualization.

\begin{figure}[h!]
	\centering
	\includegraphics[angle=0,width=15cm]{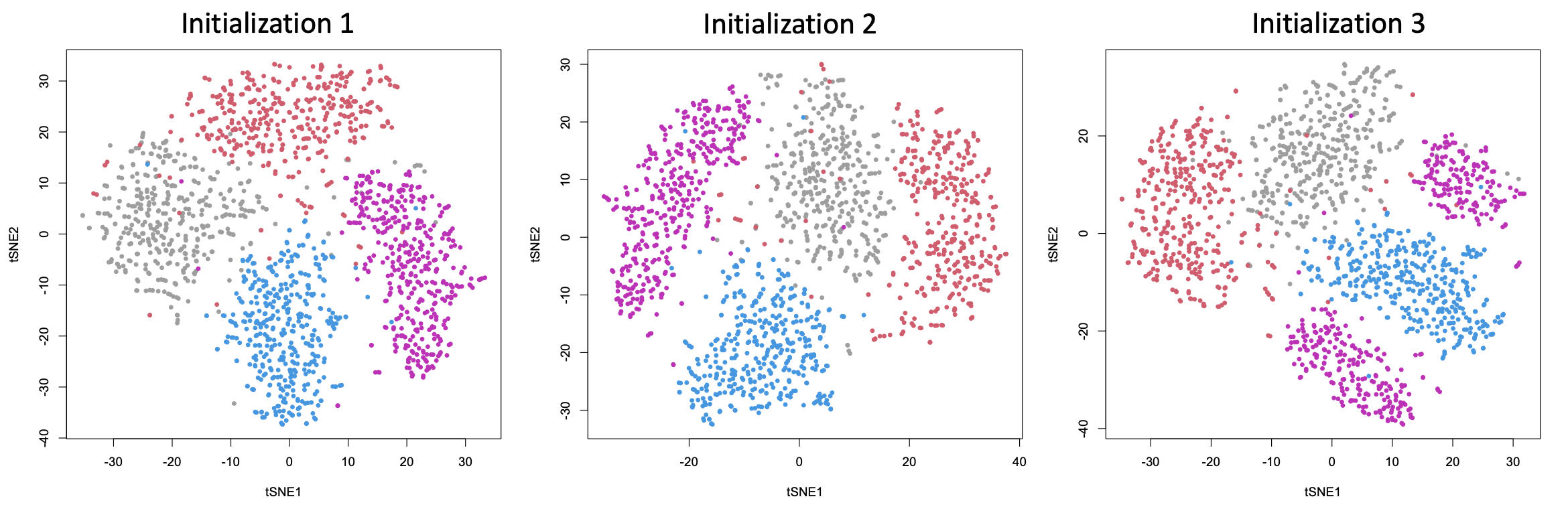}
	\caption{t-SNE visualizations of 1600 MNIST samples based on three different random initializations and identical tuning parameters in (\ref{tuning}).} 
	\label{initialize.fig}
\end{figure}

Next we assess the effects and artifacts of the random initialization. In Figure \ref{initialize.fig}, we fix all the tuning parameters as in (\ref{tuning}) and generate t-SNE visualizations from three different random initializations. Comparing the first two plots, we observe that the relative positions of  the clusters vary with the initialization. For example,  the purple cluster and red cluster are neighbors in the left panel but not in the middle panel. This echos our theoretical prediction (discussion after Theorem \ref{tsne.cor2}) and justifies our second practical advice in Section \ref{contr.sec}. On the other hand, on the right panel of Figure \ref{initialize.fig}, we find that even with a proper choice of the tuning parameters, false clustering may still appear as an artifact of the random initialization (cf. Remark \ref{em.remark} and the third practical advice in Section \ref{contr.sec}).

Finally, we point out that our theory-guided values for the tuning parameters are flexible, robust and adaptive to the sample size. For example, in Figure \ref{delta.fig}, we present three more visualizations of $n=2400$ ($N=600$ for each digit) MNIST samples, by using the tuning parameters in (\ref{tuning}) with various $\delta\in \{1/3, 1/2, 2/3\}$, and an identical random initialization. The cluster patterns are visible and similar in all the cases, showing the effectiveness of our tuning parameters and the insensitivity to the choice of $\delta$.

%%%%%%%%%%%%%%%%%%%%%%%%%%%%%%%%%%%%%%%%%%%%%%%%%%%%%%%%	
\section{Discussion} \label{dis.sec}
%%%%%%%%%%%%%%%%%%%%%%%%%%%%%%%%%%%%%%%%%%%%%%%%%%%%%%%%	

The present paper provides theoretical foundations of t-SNE for visualizing clustered data and obtains insights about its theoretical properties and interpretations. We believe that some of the conditions may be relaxed by adopting more advanced technical tools. For example, the current analysis of the early exaggeration stage relies on the well-celebrated Davis-Kahan matrix perturbation inequality (cf. Section \ref{A3.sec}), which may be further improved by leveraging advanced results in Random Matrix Theory, such as \cite{benaych2012singular} and \cite{bao2021singular}.

There are still many interesting questions that remain  to be explored. For instance, what is the limiting behavior of the low-dimensional map $\{y_i^{(k)}\}_{1\le i\le n}$ towards the end of the embedding stage, after transition to the stabilization phase? How to interpret the local structure within a cluster \citep{depavia2020spectral,robinson2020tree}? How many iterations are needed for the embedding stage? How to determine the bandwidth $\{\tau_i\}$ in a data-driven and adaptive manner \citep{ding2022learning}? {The present work is a first step towards answering these important questions.} %{\red I am not sure about this sentence.  It seems to imply that a little more work will lead to answers to these questions. Is that true?}

Moreover, our theoretical framework is  generic and can be generalized to study other algorithms that are closely related to or share similar features with t-SNE. For example, in addition to the variants of t-SNE mentioned in  Section \ref{intro.sec}, many dimension reduction and data visualization methods, such as multidimensional scaling \citep{kruskal1978multidimensional}, kernel principal component analysis \citep{scholkopf1997kernel}, and Laplacian eigenmap \citep{belkin2003laplacian}, start with a similarity matrix summarizing the pairwise distances within a dataset, and then proceed by either explicitly or implicitly exploiting the spectral properties of the similarity matrix. In this connection, the general ideas behind our theoretical analysis, such as identifying the underlying structured graph and properties of its adjacency or Laplacian matrix (Sections \ref{graph.sec} and \ref{power.sec}), studying the gradient flow associated with the discrete algorithm (Section \ref{cont.sec}), and the mechanical/kinematic view of the updating equation (Section \ref{eb.sec}), can be adopted to uncover the underlying mechanism and the properties of these methods.

It is also interesting to explore the fundamental limit for data visualization and dimension reduction. For example, what are the necessary conditions for the data $\{X_i\}_{1\le i\le n}$ to guarantee the existence of a low-dimensional map $\{y_i\}_{1\le i\le n}$ being  a metric embedding of it? {Whether t-SNE has to sacrifice some global structures in order to locally embed the data well \citep{,chari2021specious}?} %{\red This sentence is unclear to me. Rephrase?} 
These problems are left for future investigation.

\section*{Acknowledgement}

The authors are grateful to the editors and four anonymous referees for their comments and suggestions which have significantly improved the results and presentation of the paper. The research of Tony Cai was supported in part by NSF grant DMS-2015259 and NIH grant R01-GM129781. The research of Rong Ma was supported by Professor David Donoho at Stanford University. This work was partially completed while Rong Ma was a PhD candidate in Biostatistics at the University of Pennsylvania. Rong Ma would like to thank Mingyao Li for introducing the subject, and Micha\"el Aupetit, David Donoho, Jeyong Lee, Stefan Steinerberger, Yiqiao Zhong and James Zou for helpful discussions.

\vskip 0.2in
\bibliography{reference}

\newpage
\appendix

\section{Discrete-Time Analysis of the Early Exaggeration Stage} \label{proof.sec}

\subsection{Proof of Theorem \ref{graph.prop}}

For simplicity, we ignore the superscript $(k)$ in $y_i^{(k)},q_{ij}^{(k)},\eta^{(k)}$ and $S_{ij}^{(k)}(\alpha)$.
Since $
S_{ij}(\alpha)=\frac{\alpha p_{ij}-q_{ij}}{1+\|y_i-y_j\|_2^2},
$
if we denote $d_{ij}=\|y_i-y_j\|_2^2$, we have
\begin{align*}
\bigg|S_{ij}(\alpha)-{\alpha p_{ij}}+\frac{1}{n(n-1)}\bigg|&\le |{\alpha p_{ij}}-\frac{\alpha p_{ij}}{1+d_{ij}}|+ \bigg|\frac{q_{ij}}{1+d_{ij}}-\frac{1}{n(n-1)}\bigg|\\
&=d_{ij}\alpha p_{ij}+\bigg|\frac{1}{Z(1+d_{ij})^2}-\frac{1}{n(n-1)}\bigg|,
\end{align*}
where
\[
Z=\sum_{i\ne j} (1+d_{ij})^{-1}=\sum_{i\ne j}\bigg(1-\frac{d_{ij}}{1+d_{ij}}  \bigg)=n(n-1)-\sum_{i\ne j}\frac{d_{ij}}{1+d_{ij}}\equiv n(n-1)-\Delta.%\ge n(n-1)(1+\eta)^{-1}.
\]
Now since $\Delta\le n(n-1)\min\{\eta,1\}=n(n-1)\eta$, we have
\begin{align}
&\bigg|\frac{1}{Z(1+d_{ij})^2}-\frac{1}{n(n-1)}\bigg|=\bigg|\frac{(1+d_{ij})^{-2}n(n-1)-n(n-1)+\Delta}{n^2(n-1)^2-n(n-1)\Delta}\bigg|\nonumber \\
&\le \frac{|[(1+d_{ij})^{-2}-1]n(n-1)+\Delta|}{n^2(n-1)^2(1-\eta)}\le \frac{2\eta}{n(n-1)(1-\eta)} \label{q.bnd}
\end{align}
Hence
\[
\bigg|S_{ij}(\alpha)-{\alpha p_{ij}}+\frac{1}{n(n-1)}\bigg|\le \eta\alpha p_{ij}+\frac{2\eta}{n(n-1)(1-\eta)}.
\]
For the second statement,  %if we define
%\[
%	\bH_n=\frac{1}{n(n-1)}({\bf 11}^\top-{\bf I}_n),
%\]
we note that
\beq \label{bH.norm}
\|\bH_n\|\le \frac{1}{n-1}+\frac{1}{n(n-1)}\lesssim \frac{1}{n}.
\eeq
Then as long as $\|\alpha\bP\|\gg \frac{1}{n}$,
we have
$
\|\alpha\bP-\bH_n\|\ge \|\alpha\bP\|-\frac{1}{n}\asymp \|\alpha\bP\|.
$
Therefore, under the condition that $	\eta\ll {\frac{\|\bP\|}{n\|\bP\|_\infty}}\le 1$,
\begin{align*}
\frac{\|\bS_\alpha-(\alpha \bP-\bH_n)\|}{\|\alpha \bP-\bH_n\|}&\lesssim \frac{n\|\bS_\alpha-(\alpha \bP-\bH_n)\|_\infty}{\|\alpha \bP\|}\lesssim \frac{\alpha  n\|\bP\|_\infty\eta+\eta/n}{\|\alpha \bP\|}
\end{align*}
Then, the first term ${\eta n\|\bP\|_\infty}/{\|\bP\|}\to 0$ as $\eta\ll \|\bP\|/(n\|\bP\|_\infty)$; the second term $\frac{\eta}{n\alpha\|\bP\|}\lesssim \frac{1}{n\alpha\|\bP\|}\to 0$ as $\alpha\gg (n\|\bP\|)^{-1}$. %The final result follows from
%\[
%\frac{\|\bS_\alpha-\alpha \bP\|}{\|\alpha \bP\|}\le \frac{\|\bS_\alpha-(\alpha \bP-\bH_n)\|}{\|\alpha \bP\|}+\frac{\|\bH_n\|}{\|\alpha \bP\|},
%\]%
%and (\ref{bH.norm}).

\subsection{Proof of Proposition \ref{cont.prop}}

Note that $
y_{\ell i}^{(k+1)}\le \|[{\bf I}-h\bL(\bS^{(k)}_\alpha )]_{i.}\|_1\|\yy_\ell^{(k)}\|_\infty$ for any $k\ge 0$,
where
\begin{align*}
\|[{\bf I}-h\bL(\bS^{(k)}_\alpha )]_{i.}\|_1&=\bigg|1-h\sum_{j=1}^n S_{ij}^{(k)}(\alpha)\bigg|+h\sum_{j\ne i}|S_{ij}^{(k)}(\alpha)|\le 1+2h\sum_{j=1}^n |S_{ij}^{(k)}(\alpha)|.
\end{align*}
For the last term, we have
\[
h\sum_{j=1}^n |S_{ij}^{(k)}(\alpha)|\le hn\|\bS_{\alpha}^{(k)}\|_\infty \le nh(\alpha\|\bP\|_\infty+\|\bQ^{(k)}\|_\infty)\le nh\alpha\|\bP\|_\infty+\frac{h(1+\eta^{(k)})}{n-1}
\]
where the last inequality follows from (\ref{q.bnd}), that is,
$\|\bQ^{(k)}\|_\infty\le 1/Z$, $Z\ge n(n-1)/(1+\eta^{(k)})$, so that $\|\bQ^{(k)}\|_\infty\le (1+\eta^{(k)})/n(n-1)$. Then we have
\[
y_{\ell i}^{(k+1)}\le  \bigg(1+ 2nh\alpha\|\bP\|_\infty+\frac{2h(1+\eta^{(k)})}{n-1}\bigg)\|\yy_{\ell}^{(k)}\|_\infty,
\]
or
\[
\|\yy_{\ell}^{(k+1)}\|_\infty\le  \bigg(1+ 2nh\alpha\|\bP\|_\infty+\frac{2h(1+\eta^{(k)})}{n-1}\bigg)\|\yy_{\ell}^{(k)}\|_\infty.
\]
Whenever $\eta^{(k)}$ and $\max\{\|\yy_{1}^{(k)}\|_\infty,\|\yy_{2}^{(k)}\|_\infty\}$ are bounded by an absolute constant, by setting $r_n=nh\alpha\|\bP\|_\infty+\frac{h}{n}$ and assuming $r_n=O(1)$ (by Condition (T1)), we have
\beq\label{it.eq}
\|\yy_{\ell}^{(k+1)}\|_\infty\le(1+Cr_n)\|\yy_{\ell}^{(k)}\|_\infty,
\eeq
and
\begin{align*}
&\eta^{(k+1)}\le 4\max_{i\in[n], \ell\in[2]} |y_{i\ell}^{(k)}|^2\le 8\max\{\|\yy_{1}^{(k+1)}\|^2_\infty,\|\yy_{2}^{(k+1)}\|^2_\infty\}\\
&\le 8(1+Cr_n)\max\{\|\yy_{1}^{(k)}\|^2_\infty,\|\yy_{2}^{(k)}\|^2_\infty\}=O(1)
\end{align*}
In other words, we have shown that for any $k$ such that $\eta^{(k)}$ and $\max\{\|\yy_{1}^{(k)}\|_\infty,\|\yy_{2}^{(k)}\|_\infty\}$ are bounded, then (\ref{it.eq}) holds, and $\eta^{(k+1)}$ and $\max\{\|\yy_{1}^{(k+1)}\|_\infty,\|\yy_{2}^{(k+1)}\|_\infty\}$ are bounded.

Now  Condition (I1) says that $\max\{\|\yy_{1}^{(0)}\|_\infty,\|\yy_{2}^{(0)}\|_\infty\}=O(1)$ and $\eta^{(0)}\le 4\max_{\ell\in[2]} \|\yy_{\ell}^{(0)}\|_\infty^2=O(1)$. By induction, we know that $\eta^{(k)}$ and $\max\{\|\yy_{1}^{(k+1)}\|_\infty,\|\yy_{2}^{(k+1)}\|_\infty\}$ are bounded and (\ref{it.eq}) holds for all $k\ge 1$. Applying (\ref{it.eq}) iteratively, we have for any $k\ge 1$,
\beq
\|\yy_{\ell}^{(k)}\|_\infty\le(1+Cr_n)^k\|\yy_{\ell}^{(0)}\|_\infty.
\eeq
Therefore, as long as $k=k(n)$ such that {$kr_n=O(1)$} (by Condition (T1)), we have 
$
{\|\yy_{\ell}^{(k)}\|_\infty}/{\|\yy_{\ell}^{(0)}\|_\infty}=O(1),
$
or
\[
\frac{ \text{diam}(\{y_i^{(k)}\}_{1\le i\le n})}{\max_{i\in[n], \ell\in[2]} |y_{i\ell}^{(0)}|}\le \frac{\max_{i\in[n], \ell\in[2]} |y_{i\ell}^{(k)}|}{\max_{i\in[n], \ell\in[2]} |y_{i\ell}^{(0)}|}=O(1).
\]

\subsection{Proof of Theorem \ref{power.thm}}

The results concerning (\ref{graph}) and (\ref{cont.ineq})  follows directly from Theorem \ref{graph.prop} and Proposition \ref{cont.prop}. To see that (\ref{pm.eq}) holds,  we first prove the following proposition. 

\bep \label{graph.cor}
Let $\bE_\alpha^{(k)}=\bS_\alpha^{(k)}-(\alpha\bP-\bH_n)$ and $\zeta= \sup_{k\ge 0} \|\bL(\bE_\alpha^{(k)})\|$. Suppose the initialization satisfies $\|\yy_\ell^{(0)}\|\ne 0$ for $\ell=1,2$, and $(\alpha, h,K)$ satisfies $h\|\bL(\alpha\bP)\|< 2$, $Kh\zeta=O(1)$ and $Kh=O(n)$ as $n\to\infty$. Then for $\ell\in\{1,2\}$, it holds that
\beq
\sup_{1\le k\le K}\frac{\|\yy_\ell^{(k)} -[{\bf I}-h\bL(\alpha\bP-\bH_n)]^{k}\yy_\ell^{(0)} \|_2}{\|\yy_\ell^{(0)} \|_2 }=O(Kh\zeta).
\eeq
Consequently, for $(\alpha,h,k)$ such that $h\|\bL(\alpha\bP)\|< 2$, $kh\zeta=o(1)$ and $kh=O(n)$, we have
\beq
\lim_{n\to\infty}\frac{\|\yy_\ell^{(k)} -[{\bf I}-h\bL(\alpha\bP-\bH_n)]^{k}\yy_\ell^{(0)} \|_2}{\|\yy_\ell^{(0)} \|_2 }=0.
\eeq 
\eep

\begin{proof}
	By linearity of the Laplacian operator, we have
	\[
	\bL(\bE_\alpha^{(k)})=	\bL(\bS_\alpha^{(k)})-\bL(\alpha\bP-\bH_n)=	\bL(\bS_\alpha^{(k)})-\bL(\alpha\bP-\bH_n)
	\]
	For $\ell=1,2,$ the updating equation 
	$
	\yy_\ell^{(k+1)}=({\bf I}-h\bL(\bS_\alpha^{(k)}))\yy_\ell^{(k)}
	$
	can be written as
	\begin{align*}
	\yy_\ell^{(k+1)}&=({\bf I}-h\bL(\bS_\alpha^{(k)}))({\bf I}-h\bL(\bS_\alpha^{(k-1)}))...({\bf I}-h\bL(\bS_\alpha^{(0)}))\yy_\ell^{(0)}\\
	&=({\bf I}-h\bL(\alpha\bP-\bH_n)-h\bL(\bE_\alpha^{(k)}))({\bf I}-h\bL(\alpha\bP-\bH_n)-\bL(\bE_\alpha^{(k-1)}))...\\
	&\quad\times({\bf I}-h\bL(\alpha\bP-\bH_n)-\bL(\bE_\alpha^{(0)}))\yy_\ell^{(0)}\\
	&=({\bf I}-h\bL(\alpha\bP-\bH_n))^{k+1}\yy_\ell^{(0)} + \bepsilon^{(k)},
	\end{align*}
	where 
	\begin{align*}
	\|\bepsilon^{(k)}\|_2&\le \|\yy_\ell^{(0)}  \|_2\bigg[{k+1\choose 1}h\zeta\lambda^k+{k+1\choose 2}(h\zeta)^2\lambda^{k-1}+...+{k+1\choose k+1}(h\zeta)^{k+1}\lambda^0\bigg]\\
	&\le \|\yy_\ell^{(0)}\|_2  [ (h\zeta+\lambda)^{k+1}-\lambda^{k+1} ]\\
	&\le \|\yy_\ell^{(0)}\|_2  \lambda^{k+1}[ ({h\zeta}/{\lambda}+1)^{k+1}-1 ]
	\end{align*}
	where $\lambda=\| {\bf I}-h\bL(\alpha\bP-\bH_n)\|$ and $\zeta = \sup_{k\ge 0} \|\bL(\bE_\alpha^{(k)})\|$. 
	We need the following lemma.% concerning the eigenvalues of ${\bf I}-h\bL(\alpha\bP-\bH_n)$.
	
	\bel \label{lem.19}
	If $\|h\bL(\alpha\bP)\|< 2$, then $1\le \|{\bf I}-h\bL(\alpha\bP-\bH_n)\|\le 1+\frac{h}{n-1}$.
	\eel
	
	The above lemma implies
	\[
	\|\bepsilon^{(k)}\|_2\le \|\yy_\ell^{(0)}\|_2  (1+Ch/n))^{k+1}[ ({h\zeta}+1)^{k+1}-1. 
	\]
	By the binomial identity,
	\begin{align*}
	(1+x_n)^k&=1+kx_n+\frac{k(k-1)}{2}x_n^2+\frac{k(k-1)(k-2)}{3!}x_n^3+...+x_n^k\\
	&\le 1+kx_n+\frac{k^2}{2}x_n^2+\frac{k^3}{3!}x_n^3+...+\frac{k^k}{k!}x_n^k\\
	&\le 1+kx_n\bigg(1+kx_n+\frac{k^2x_n^2}{2!}+...+\frac{k^{k-1}x_n^{k-1}}{(k-1)!}\bigg)\\
	&\le 1+kx_ne^{kx_n}.
	\end{align*}
	Then, as long as $kh\zeta=O(1)$ as $n\to \infty$, there exits some universal constant $C>0$ such that, for all $n$,
	\beq
	(1+h\zeta)^{k+1}-1\le Ckh\zeta.
	\eeq
	Similarly, whenever $kh/n=O(1)$, we have $ (1+Ch/n))^{k+1}\le C$.
	Hence, we have
	$
{\|\bepsilon^{(k)}\|_2}/{\|\yy_\ell^{(0)}\|_2}\le Ckh\zeta.
	$
	This proves  the theorem.
\end{proof}

By the above proposition, it suffices to reduce the following full list of conditions --  $\alpha\gg \frac{1}{n\|\bP\|}$, $kh=O(n)$, $nh\alpha\|\bP\|_\infty<1$, $\max_{i\in[n], \ell\in[2]} |y_{i\ell}^{(0)}|^2=o(\|\bP\|/(n\|\bP\|_\infty))$, $\|\yy_\ell^{(0)}\|\ne 0$ for $\ell\in\{1,2\}$, $k(nh\alpha\|\bP\|_\infty+h/n) =O(1)$, and $k(nh\alpha\|\bP\|_\infty+h/n) \cdot\max_{i\in[n], \ell\in[2]} |y_{i\ell}^{(0)}|^2=O(1)$ -- to those in (I1) (I2) and (T1.D).
To see this, note that 	$k\big(nh\alpha\|\bP\|_\infty +{h}/{n}\big)\cdot\max_{i\in[n], \ell\in[2]} |y_{i\ell}^{(0)}|^2\to 0$, can be implied by
\[
k\big(nh\alpha\|\bP\|_\infty +{h}/{n}\big)=o(1), \quad \max_{i\in[n], \ell\in[2]} |y_{i\ell}^{(0)}|^2\ll \|\bP\|/(n\|\bP\|_\infty)\le 1.
\]
In addition, $kh=O(n)$ and $\alpha hn\|\bP\|_\infty<1$ can be implied by the above first inequality.

\subsection{Proof of Theorem \ref{power.thm2}}

Since
$
\bL(\bH_n)=\frac{1}{n-1}{\bf I}_n-\frac{1}{n(n-1)}{\bf 1}_n{\bf1}_n^\top,
$
it follows that
\beq
{\bf I}_n-h\bL(\bP-\bH_n)={\bf I}_n-h\bL(\bP)+\frac{h}{n-1}{\bf I}_n-\frac{h}{n(n-1)}{\bf 1}_n{\bf1}_n^\top.
\eeq
Without loss of generality, we assume $R\ge 2$, as the case for $R=1$ follows similarly.
Let $\bu_i$ be the $i$-th column of $\bU'\in O(n,R-1)$, which consists of the eigenvectors corresponding to the eigenvalue $0$ of $\bL(\bP)$ other than the trivial eigenvector $n^{-1/2}{\bf 1}_n$, and let $\bU$ be the matrix that binds an additional column $n^{-1/2}{\bf 1}_n$ to $\bU'$. Let $\lambda_1\le ...\le\lambda_n$ be the eigenvalues of $\bL(\bP)$, with $\lambda_1=...=\lambda_R=0$. Then it follows that
\begin{align*}
&{\bf I}-h\bL(\bP-\bH_n)\\
&=[\bU' \quad n^{-1/2}{\bf 1}_n\quad \bU_{\perp}]
{\tiny	\begin{bmatrix}
	1+\frac{h}{n-1}-h\lambda_1 &&&&&&\\
	&\ddots&&&&&\\
	&&1+\frac{h}{n-1}-h\lambda_{R-1}&&&&\\
	&&&1+\frac{h}{n-1}-h\lambda_{R}&&&\\
	&&&&1+\frac{h}{n-1}-h\lambda_{R+1}&&\\
	&&&&&\ddots&\\
	&&&&&&	1+\frac{h}{n-1}-h\lambda_n
	\end{bmatrix}}\\
&\quad\times
[\bU' \quad n^{-1/2}{\bf 1}\quad \bU_{\perp}]^\top-\frac{h}{n(n-1)}{\bf 11}^\top\\
&=[\bU' \quad n^{-1/2}{\bf 1}\quad \bU_{\perp}]
{\tiny\begin{bmatrix}
	1+\frac{h}{n-1} &&&&&&\\
	&\ddots&&&&&\\
	&&1+\frac{h}{n-1}&&&&\\
	&&&1&&&\\
	&&&&1+\frac{h}{n-1}-h\lambda_{R+1}&&\\
	&&&&&\ddots&\\
	&&&&&&	1+\frac{h}{n-1}-h\lambda_n
	\end{bmatrix}}[\bU' \quad n^{-1/2}{\bf 1}\quad \bU_{\perp}]^\top
\end{align*}
Then if we denote $\bu_{R+1}, ..., \bu_n$ as the columns of $\bU_{\perp}$, we have
\[
(	{\bf I}-h\bL(\bP-\bH_n))^k=\sum_{i=1}^{R-1}\bigg( 1+\frac{h}{n-1}\bigg)^k \bu_i\bu_i^\top+n^{-1}{\bf 11}^\top+\sum_{i=R+1}^{n}\bigg(1+\frac{h}{n-1}-h\lambda_{i}\bigg)^k\bu_i\bu_i^\top.
\]
Hence
\begin{align*}
&\bigg\|({\bf I}-h\bL)^k\yy-\bigg( 1+\frac{h}{n-1}\bigg)^k\sum_{i=1}^{R-1}\bu_i\bu_i^\top\yy-n^{-1}{\bf 11}^\top\yy\bigg\|_2= \bigg\| \sum_{i=R+1}^n\bigg(1+\frac{h}{n-1}-h\lambda_{i}\bigg)^k\bu_i\bu_i^\top\yy \bigg\|_2\\
&= \sqrt{\sum_{i=R+1}^n\bigg(1+\frac{h}{n-1}-h\lambda_{i}\bigg)^{2k}(\bu_i^\top\yy)^2}\le \bigg(1+\frac{h}{n-1}-h\lambda_{R+1}\bigg)^k\|\yy\|_2.
\end{align*}
The final result follows by noting that 
$
1+\frac{h}{n-1}-h\lambda_{R+1}<1-\kappa/2<1.
$

\subsection{Proof of Proposition \ref{eigen}}

Firstly, since $\bA$ is nonnegative, by the Ger{\v{s}}gorin circle theorem \citep{varga2010gervsgorin}, $\bL(\bA)$ is positive semi-definite.
For any $\btheta_r$, $r\in\{1,...,R\}$, it holds that
$
\bL(\bP)\btheta_r=0.
$
It follows that $\{\btheta_r\}_{r=1}^R$ is a set of eigenvectors corresponding to the smallest eigenvalue $0$. In addition, since the graph corresponding to the weighted adjacency matrix $\bP$  has $R$ connected components, by the spectral property of the Laplacian matrix (see, for example, Theorem 3.10 of \cite{marsden2013eigenvalues}),  the null space of $\bL(\bP)$ has dimension $R$. This implies that the eigenvalue $0$ of  $\bL(\bP)$ has multiplicity $R$. Lastly, as $\{\btheta_k\}_{k=1}^K$  are linearly independent, the eigen subspace associated with the eigenvalue $0$ is spanned by $\{\btheta_r\}_{r=1}^R$.

\subsection{Proof of Theorem \ref{tsne.cor2}}

Let ${\bE'}_\alpha^{(k)}=\bS_{\alpha}^{(k)}-\alpha\bP^*+\bH_n$ and $\zeta'=\sup_{k\ge 1}\|\bL({\bE'}_{\alpha}^{(k)})\|$. Then similar arguments as in the proof of Theorem \ref{power.thm} imply that
$
{\bepsilon'}^{(k)}=\yy_{\ell}^{(k+1)}-({\bf I}-h\bL(\alpha\bP^*-\bH_n))^{k+1}\yy_{\ell}^{(0)}
$
satisfies
\[
\frac{\|{\bepsilon'}^{(k)}\|_2}{\|\yy_{\ell}^{(0)}\|_2}\le (1+Ch/n))^{k+1}[(1+h\zeta')^{k+1}-1].
\]
As a result, as long as $h\zeta'\ll 1/k$, we have 
\[
\lim_{(n,k)\to \infty}\frac{\|\yy_{\ell}^{(k)}-[{\bf I}-h\bL(\alpha\bP^*-\bH_n)]^k \yy_{\ell}^{(0)}\|_2}{\|\yy_{\ell}^{(0)}\|_2}=0.
\]
Now by the inequality 
$
\|\bL(\bA)\|\le   \|\bD(\bA)\|+ \|\bA\|\le 2n\|\bA\|_\infty,
$
and the bounded initialization,  we have
\begin{align}
\|\bL({\bE'}_{\alpha}^{(k)})\|&\le \|\bL(\bS_{\alpha}^{(k)}-\alpha\bP+\bH_n)\|+ \alpha\|\bL(\bP^*-\bP)\|\nonumber \\
&\le 2n\|\bS_{\alpha}^{(k)}-\alpha\bP+\bH_n\|_\infty+\alpha\|\bL(\bP^*-\bP)\|\nonumber \\
&\le (\alpha n\|\bP\|_\infty+1/n)\cdot\max_{i\in[n], \ell\in[2]} |y_{i\ell}^{(0)}|^2+\alpha\|\bL(\bP^*-\bP)\| \label{LE'}
\end{align}
where the last inequality follows from Proposition \ref{graph.prop}. Thus,
%	\begin{align*}
%	\|\bL({\bE'}_{\alpha}^{(k)})\|&\le 2n\|\bS_{\alpha}^{(k)}-\alpha\bP\|_\infty+2n\alpha\|\bP-\bP^*\|_{\infty}\\
%	&\lesssim \alpha n\eta^{(k)}\|\bP\|_\infty+\frac{1}{n}+n\alpha\|\bP^*-\bP\|_\infty\\
%	&\lesssim \alpha n\eta^{(k)}\|\bP^*\|_\infty+\frac{1}{n}+n\alpha\|\bP^*-\bP\|_\infty,
%	\end{align*}
the condition $h\zeta'\ll 1/k$ can be implied by 
\[
k[(nh\alpha\|\bP\|_\infty+h/n)\cdot\max_{i\in[n], \ell\in[2]} |y_{i\ell}^{(0)}|^2+\alpha h\|\bL(\bP^*-\bP)\|]=o(1),
\]
which holds under Conditions (I2) (T1.D) and (T2.D).
Now, if we further assume $\kappa<h\lambda_{R+1}(\bL(\alpha\bP^*))\le h\|\bL(\alpha\bP^*)\|\le 1$ and $k\to\infty$, by Theorem \ref{power.thm2}, we also have
\[
\lim_{(n,k)\to \infty}\frac{\|\yy_{\ell}^{(k)}-\bU\bU^\top \yy_{\ell}^{(0)}\|_2}{\|\yy_{\ell}^{(0)}\|_2}=0,
\]
where the columns of $\bU\in O(n,R-1)$ span the null space of $\bL(\bP^*)$. 
By Proposition \ref{eigen}, we know that the matrix
$
\bU\bU^{\top}\begin{bmatrix}
\yy_{1}^{(0)}&\yy_{2}^{(0)}\end{bmatrix}\in \R^{n\times 2}
$
has at most $R$ distinct rows, and any two rows corresponding to the same graph component in $G$ have the identical values.
Then, the final results follow by setting $\{z_1,...,z_R\}$ such that $z_r=(z_{1r},z_{2r})$ is the same as the rows in $\bU\bU^{\top}\begin{bmatrix}
\yy_{1}^{(0)}&\yy_{2}^{(0)}\end{bmatrix}$ corresponding to the $r$-th graph component.

\section{Continuous-Time Analysis of the Early Exaggeration Stage}

\subsection{Proof of Proposition \ref{cont.thm}}

Note that the algorithm (\ref{y.tilde.alg}) is in fact the Euler scheme for solving the differential equation (\ref{GF}).
We can apply the standard differential equation theory to obtain the global approximation error for the Euler scheme. By taking derivative on both sides of the differential equation (\ref{GF}), we have
\[
\ddot{\bY}_\ell(t)=\bL(\alpha\bP-\bH_n)\dot{\bY}_\ell(t)=\bL^2(\alpha\bP-\bH_n)\bY_\ell(t).
\]
Since
$
\|\ddot{\bY}_\ell(t)\|_2\le \|\bL(\alpha\bP-\bH_n)\|^2\|\bY_\ell(t)\|_2,
$
by Theorem 212A of \cite{butcher2008numerical}, we have
\[
\sup_{t\in[0,T]}\frac{\|{\yy}_{\ell,h}(t)-\bY_{\ell}(t)\|_2}{\|\bY_\ell(t)\|_2}\le Th\|\bL(\alpha\bP-\bH_n)\|^2.
\]
Consequently, for any $k\le T/h$, if we set $t=kh\le T$, then
\[
\frac{\|\tilde{\yy}_{\ell}^{(k)}-\bY_{\ell}(kh)\|_2}{\|\bY_\ell(kh)\|_2}=\frac{\|{\yy}_{\ell,h}^{(t)}-\bY_{\ell}(t)\|_2}{\|\bY_\ell(t)\|_2}\le Th\|\bL(\alpha\bP-\bH_n)\|^2.
\]
This proves the second statement of the theorem.

\subsection{Proof of Proposition \ref{Yt.thm}}

By standard theory of ODE, we have
$
\bY_\ell(t)=\sum_{i=1}^ne^{-t\sigma_i}(\bu_i^\top\yy_\ell^{(0)})\bu_i,
$
where $\bL(\alpha\bP-\bH_n)=\sum_{i=1}^n\sigma_i\bu_i\bu_i^\top$  is the eigendecomposition of $\bL(\alpha\bP-\bH_n)$. The final result follows from the fact that
$
\bL(\alpha\bP-\bH_n)=\bL(\alpha\bP)-\frac{1}{n-1}{\bf I}+\frac{1}{n(n-1)}{\bf 11}^\top,
$
so that $ \bL(\alpha\bP-\bH_n)$ and $\bL(\alpha\bP)$ share the same set of eigenvectors, and
\[
\sigma_i=\alpha\lambda_i-\frac{1}{n-1},\quad i\in\{2,...,n\},
\]
and $\sigma_1=\alpha\lambda_1$.

\subsection{Proof of Theorem \ref{es.thm}} \label{A3.sec}

Let $\bU_0\in O(n,R)$ be the matrix whose columns span the null space of $\bL(\bP^*)$, and the first column of $\bU_0$ is $n^{-1/2}{\bf 1}$. Let $\bU\in O(n,R)$ be the matrix whose columns correspond to the smallest $R$ eigenvalues of $\bL(\bP)$. By standard Davis-Kahan matrix perturbation inequality, we have
\beq
\|\bU_{0\perp}^\top\bU\|\le \frac{\|\bL(\bE)\|}{\lambda_{R+1}(\bL(\bP^*))},
\eeq
where $\bE=\bP-\bP^*$, and $\lambda_{R+1}(\bL(\bP^*))$ is the smallest nonzero eigenvalue of $\bL(\bP^*)$. Note that if we define $\bbeta_t=(\beta_{t,1},...,\beta_{t,n})\in \R^n$ such that $\beta_{t,i}=e^{-t(\alpha\lambda_i-\frac{1}{n-1})}(\bu_i^\top \yy_\ell^{(0)})$ for $i=2,...,n$ and $\beta_{t,1}=n^{-1/2}{\bf 1}_n^\top \yy_\ell^{(0)}$, we can write
$
\bY_\ell(t)=\bU\bbeta_{t,1:R}+\bU_{\perp}\bbeta_{t,(R+1):n}.
$
In particular, we have
\[
\|	\bY_\ell(t)\|_2=\|\bU\bbeta_{t,1:R}\|_2+\|\bU_{\perp}\bbeta_{t,(R+1):n}\|_2.
\]
Since $\|\bY_{\ell}(t)\|_2\lesssim \|\bY_{\ell}(0)\|_2$, it follows that
\begin{align*}
&\quad\frac{\| \bU_0\bU_0^\top\bY_\ell(t)-\bY_\ell(t)\|_2}{\|\bY_\ell(0)\|_2}\\
&=\frac{\| \bU_0\bU_0^\top\bU\bbeta_{t,1:R}-\bU\bbeta_{t,1:R}\|_2}{\|\bY_\ell(t)\|_2}\frac{{\|\bY_\ell(t)\|_2}}{{\|\bY_\ell(0)\|_2}}+\frac{\| \bU_0\bU_0^\top\bU_{\perp}\bbeta_{t,(R+1):n}-\bU_{\perp}\bbeta_{t,(R+1):n}\|_2}{\|\bY_\ell(0)\|_2}\\
&\lesssim \| (\bU_0\bU_0^\top-{\bf I})\bU\bU^\top\|\cdot\frac{ \|\bU\bbeta_{t,1:R}\|_2}{\|\bU\bbeta_{t,1:R}\|_2+\|\bU_{\perp}\bbeta_{t,(R+1):n}\|_2}+\| (\bU_0\bU_0^\top-{\bf I})\bU_{\perp}\bU_{\perp}^\top\|\cdot \frac{\|\bU_{\perp}\bbeta_{t,(R+1):n}\|_2}{\|\bY_\ell(0)\|_2}\\
&\le   \|(\bU_0)_{\perp}^\top\bU\|+\|(\bU_0)_{\perp}^\top\bU_{\perp}\|\cdot\frac{\|\bU_{\perp}\bbeta_{t,(R+1):n}\|_2}{\|\bY_\ell(0)\|_2}\\
&\le \|(\bU_0)_{\perp}^\top\bU\|+\frac{\|\bU_{\perp}\bbeta_{t,(R+1):n}\|_2}{\|\bY_\ell(0)\|_2}\\
&\le \frac{\|\bL(\bE)\|}{\lambda_{R+1}(\bL(\bP^*))}+\frac{e^{-t(\alpha\lambda_{R+1}(\bL(\bP))-\frac{1}{n-1})}\|\bU_{\perp}\yy_\ell^{(0)}\|_2}{\|\yy_\ell^{(0)}\|_2}\\
&\le \frac{\|\bL(\bE)\|}{\lambda_{R+1}(\bL(\bP^*))}+e^{-t(\alpha\lambda_{R+1}(\bL(\bP))-\frac{1}{n-1})}.
%	&\le \frac{\|\bL(\bE)\|}{\lambda_{R+1}(\bL(\bP^*))}+\frac{e^{-t\alpha\lambda_{R+1}(\bL(\bP^*))}e^{-t\alpha\lambda_{n}(\bL(\bE))}\|\bU_{\perp}\yy_\ell^{(0)}\|_2}{e^{-t\alpha\lambda_{R}(\bL(\bP))}\|\bU\yy_\ell^{(0)}\|_2},
\end{align*}
Therefore,
whenever
\beq \label{ir.cond.1}
\|\bL(\bE)\|\ll \lambda_{R+1}(\bL(\bP^*)),\quad \alpha\lambda_{R+1}(\bL(\bP))\gg \frac{1}{n},\quad t\alpha\lambda_{R+1}(\bL(\bP))\to\infty,
\eeq
we have
\[
\lim_{(t,n)\to\infty}\frac{\| \bU_0\bU_0^\top\bY_\ell(t)-\bY_\ell(t)\|_2}{\|\bY_\ell(0)\|_2}=0.
\]
On the other hand, note that
$
\bY_\ell(t)=\bU\bGam_1(t)\bU^\top \yy_{\ell}^{(0)}+\bU_{\perp}\bGam_2(t)\bU_{\perp}^\top \yy_{\ell}^{(0)},
$
where $\bGam_1(t)=\text{diag}(1, ..., e^{-t(\alpha\lambda_R(\bL(\bP))-\frac{1}{n-1})})$, and $\bGam_2(t)=\text{diag}(e^{-t(\alpha\lambda_{R+1}(\bL(\bP))-\frac{1}{n-1}) }, ..., e^{-t(\alpha\lambda_n(\bL(\bP))-\frac{1}{n-1})})$. We have
\begin{align*}
\frac{\|\bU_0\bU_0^\top \bY_\ell(t)-\bU_0\bU_0^\top \bY_{\ell}(0)\|_2}{\|\bY_{\ell}(0)\|_2}&= 	\frac{\|\bU\bGam_1(t)\bU^\top \bY_{\ell}(0)+\bU_{\perp}\bGam_2(t)\bU_{\perp}^\top\bY_{\ell}(0)-\bU_0\bU_0^\top \bY_{\ell}(0)\|_2}{\|\bY_{\ell}(0)\|_2}\\
&\le \frac{\|\bU\bGam_1(t)\bU^\top \bY_{\ell}(0)-\bU_0\bU_0^\top \bY_{\ell}(0)\|_2}{\|\bY_{\ell}(0)\|_2} +\frac{\|\bU_{\perp}\bGam_2(t)\bU_{\perp}^\top \bY_{\ell}(0)\|_2}{\|\bY_{\ell}(0)\|_2}\\
&\le \|\bU(\bGam_1(t)-{\bf I}_R)\bU^\top+\bU\bU^\top-\bU_0\bU_0^\top\|+\|\bU_{\perp}\bGam_2(t)\bU_{\perp}^\top\|\\
&\le \|\bGam_1(t)-{\bf I}_R\|+2\|\bU^\top\bU_{0\perp}\|+\|\bGam_2(t)\|\\
&\le |e^{-t(\alpha\lambda_R(\bL(\bP))-\frac{1}{n-1}) }-1|+\frac{2\|\bL(\bE)\|}{\lambda_{R+1}(\bL(\bP^*))}+e^{-t(\alpha\lambda_{R+1}(\bL(\bP))-\frac{1}{n-1}) },
\end{align*}
where we used the inequality $\|\bU\bU^\top-\bU_0\bU_0\|\le 2\|\bU^\top\bU_{0\perp}\|$ from Lemma 1 of \cite{cai2018rate}.
Hence, whenever
\beq \label{ir.cond.2}
\|\bL(\bE)\|\ll \lambda_{R+1}(\bL(\bP^*)),\quad \alpha\lambda_{R+1}(\bL(\bP))\gg \frac{1}{n},\quad t\alpha\lambda_{R+1}(\bL(\bP))\to\infty,\quad t\bigg(\alpha\lambda_R(\bP)-\frac{1}{n-1}\bigg)\to 0,
\eeq
we also have
\[
\lim_{(n,t)\to\infty}\frac{\|\bU_0\bU_0^\top \bY_\ell(t)-\bU_0\bU_0^\top \yy_{\ell}^{(0)}\|_2}{\|\yy_{\ell}^{(0)}\|_2}=0.
\]
The second statement can be obtained by noticing that
\begin{align*}
\frac{\|\bY_\ell(t)-O\bz_\ell\|_2}{\|\bY_{\ell}(0)\|_2}&\le \frac{\|\bY_\ell(t)-\bU_0\bU_0^\top\bY_\ell(t)\|_2}{\|\bY_{\ell}(0)\|_2}+\frac{\|\bU_0\bU_0^\top\bY_\ell(t)-\bU_0\bU_0^\top\bY_\ell(0)\|_2}{\|\bY_{\ell}(0)\|_2},
\end{align*}
and that $\bU_0\bU_0^\top \bY_\ell(0)=O\bz_\ell$. To see the above conditions hold under Conditions (I1) (T1.C) and (T2.C), we note that by Weyl's inequality, 
$
\max_{i}|\lambda_i(\bL(\bP))-\lambda_i(\bL(\bP^*))|\le \|\bL(\bE)\|.
$ 
Since
$
\lambda_{R+1}(\bL(\bP))\ge\lambda_{R+1}(\bL(\bP^*))-\|\bL(\bE)\|,
$
the conditions in (\ref{ir.cond.1}) can be implied by
\[
\|\bL(\bE)\|\ll\lambda_{R+1}(\bL(\bP^*)),\quad\alpha\lambda_{R+1}(\bL(\bP))\gg \frac{1}{n},\quad t\alpha \lambda_{R+1}(\bL(\bP^*))\to\infty.
\]
On the other hand, the conditions in (\ref{ir.cond.2}) can be implied by
\begin{align*}
\|\bL(\bE)\|\ll\lambda_{R+1}(\bL(&\bP^*)),\quad\alpha\lambda_{R+1}(\bL(\bP))\gg \frac{1}{n},\quad t\alpha \lambda_{R+1}(\bL(\bP^*))\to\infty,\\
&t\alpha\|\bL(\bE)\|\to 0,\quad t/n\to 0.
\end{align*}
These are ensured by the conditions of the theorem.

\section{Analysis of the Embedding Stage}

\subsection{Proof of Proposition \ref{S.lem}}

Note that
$
\frac{q_{ij}^{(k)}}{1+d_{ij}^{(k)}}=\frac{1}{Z^{(k)}(1+d_{ij}^{(k)})^2}=\frac{(1+d_{ij}^{(k)})^{-2}}{\sum_{i\ne j}(1+d_{ij}^{(k)})^{-1}},
$
where $(1+\eta^{(k)})^{-2}\le(1+d_{ij}^{(k)})^{-2}\le 1$ for all $i\ne j$. It holds that 
\begin{align*}
\bigg|\frac{{q_{ij}^{(k)}}/(1+d_{ij}^{(k)})}{{1}/{n(n-1)}}-1\bigg|&=\bigg|  \frac{n(n-1)(1+d_{ij}^{(k)})^{-2}}{\sum_{i\ne j}(1+d_{ij}^{(k)})^{-1}}-1\bigg|\\
&\le  \frac{|(1+d_{ij}^{(k)})^{-2}n(n-1)-\sum_{i\ne j}(1+d_{ij}^{(k)})^{-1}|}{\sum_{i\ne j}(1+d_{ij}^{(k)})^{-1}}\\
&\le \frac{|(1+d_{ij}^{(k)})^{-2}n(n-1)-\sum_{i\ne j}(1+d_{ij}^{(k)})^{-1}|}{n(n-1)/(1+\eta^{(k)})}\\
&\le \max\bigg\{ \frac{|n(n-1)-\sum_{i\ne j}(1+d_{ij}^{(k)})^{-1}|}{n(n-1)/(1+\eta^{(k)})},\frac{|(1+\eta^{(k)})^{-2}n(n-1)-\sum_{i\ne j}(1+d_{ij}^{(k)})^{-1}|}{n(n-1)/(1+\eta^{(k)})}   \bigg\}\\
&\le \max\bigg\{ \frac{|(1+\eta^{(k)})^{-2}n(n-1)-n(n-1)|}{n(n-1)/(1+\eta^{(k)})}, \frac{|n(n-1)-n(n-1)(1+\eta^{(k)})^{-1}|}{n(n-1)/(1+\eta^{(k)})},\\
&\quad\frac{|(1+\eta^{(k)})^{-2}n(n-1)-n(n-1)(1+\eta^{(k)})^{-1}|}{n(n-1)/(1+\eta^{(k)})}   \bigg\},
\end{align*}
where the last two inequalities follows from the fact that 
$
|a-b|\le \max\{|c-b|,|d-b|\}
$ for any $c<a<d$.
Therefore, by $\eta^{(k)}=o(1)$, we have
\[
\bigg|\frac{{q_{ij}^{(k)}}/(1+d_{ij}^{(k)})}{{1}/{n(n-1)}}-1\bigg|\lesssim |(1+\eta^{(k)})^{-2}-1|+|(1+\eta^{(k)})^{-1}-1|\lesssim \eta^{(k)}.
\]
Now since $p_{ij}\ge 0$, we have
$
S_{ij}^{(k)}=\frac{p_{ij}-q_{ij}^{(k)}}{1+d_{ij}^{(k)}} \ge \frac{p_{ij}}{1+\eta}-\frac{1+O(\eta^{(k)})}{n(n-1)}.
$
Similarly, we can obtain
$
S_{ij}^{(k)}=\frac{p_{ij}-q_{ij}^{(k)}}{1+d_{ij}^{(k)}} \le {p_{ij}}-\frac{1+O(\eta^{(k)})}{n(n-1)}.
$
Hence, if $p_{ij}=o(n^{-2})$ and $\eta=o(1)$, we have
\[
S_{ij}^{(k)}\asymp -\frac{1+O(\eta^{(k)})}{n(n-1)}.
\]
This proves the first statement of the lemma. The second statement can be obtained from the similar argument.

\subsection{Proof of Theorem \ref{net.force.thm}}

The proof is divided into two parts.
\paragraph{Case I: $k=K_0$.} By (\ref{ue4}), summing up the contribution of all the points with index $j\ne i$ such that $j\sim i$, we have
\begin{align*}
\bigg\|\sum_{j:j\sim i}S_{ij}^{(k)}(y_j^{(k)}-y_i^{(k)})\bigg\|_2&\lesssim (n^{-2}+\|\bP^*\|_\infty+\|\bE\|_\infty)\cdot \sum_{j:j\sim i}\|y_j^{(k)}-y_i^{(k)}\|_2\\
&\lesssim \sqrt{n}(n^{-2}+\|\bP^*\|_\infty) R_n(\|\yy_1^{(0)}\|_2+\|\yy_2^{(0)}\|_2)
\end{align*}
where  $R_n=(1-\kappa)^{k+1}+hk((\alpha n\|\bP\|_\infty+1/n)\cdot \max_{i\in[n],\ell\in[2]}|y_{i\ell}^{(0)}|^2+\alpha\|\bL(\bP^*-\bP)\|)$, the first inequality follows from Proposition \ref{S.lem}, and the second inequality follows from the fact that, 
\begin{align}
&\sum_{j:j\sim i}\|y_j^{(k)}-y_i^{(k)}\|_2\lesssim \sum_{\ell=1}^2 \|\yy_\ell^{(k)}-O\bz_\ell\|_1\le\sqrt{n} \sum_{\ell=1}^2 \|\yy_\ell^{(k)}-O\bz_\ell\|_2\nonumber\\
&\le\sqrt{n} \sum_{\ell=1}^2[\|\yy_\ell^{(k)}-({\bf I}-h\bL(\alpha\bP^*-\bH_n))^k\yy_\ell^{(0)}\|_2+\|({\bf I}-h\bL(\alpha\bP^*-\bH_n))^k\yy_\ell^{(0)}-O\bz_\ell\|_2]\nonumber \\
&\lesssim \sqrt{n}[(1-\kappa)^{k+1}+hk\sup_{s< k}\|\bL(\bS^{(s)}_\alpha-\alpha\bP^*+\bH_n)\|](\|\yy_1^{(0)}\|_2+\|\yy_2^{(0)}\|_2)\label{3rd.ineq} \\
&\lesssim \sqrt{n}[(1-\kappa)^{k+1}+hk((\alpha n\|\bP\|_\infty+1/n)\cdot \max_{i\in[n],\ell\in[2]}|y_{i\ell}^{(0)}|^2+\alpha\|\bL(\bP^*-\bP)\|)](\|\yy_1^{(0)}\|_2+\|\yy_2^{(0)}\|_2)\nonumber\\
&=\sqrt{n}R_n(\|\yy_1^{(0)}\|_2+\|\yy_2^{(0)}\|_2)\label{bnd.eq}
\end{align}
where (\ref{3rd.ineq}) follow from Theorems \ref{power.thm} and \ref{power.thm2}, by assuming { $\kappa+\frac{h}{n-1}<h\lambda_{R+1}(\bL(\alpha\bP^*))\le h\lambda_{n}(\bL(\alpha\bP^*))\le 1+\frac{h}{n-1}$}, and the last inequality follows from (\ref{LE'}), by assuming { $kh[(n\alpha\|\bP\|_\infty+h1/n)\max_{i\in[n],\ell\in[2]}|y_{i\ell}^{(0)}|^2+\alpha h\|\bL(\bP^*-\bP)\|]=o(1)$}. 

On the other hand, we consider the contribution of all the other points $j\nsim i$. %Note that under Condition (A2) we have $R'_n\lesssim R_n$.
Suppose $i\in H_r$. Then for any $s\ne r$, by the fact that $S_{ij}^{(k)}=\frac{1+O(\eta^{(k)})}{n(n-1)}$ for $i\nsim j$, we have
\begin{align*}
&\bigg\|h'\sum_{j\in H_s}S_{ij}^{(k)}(y_j^{(k)}-y_i^{(k)})-f_{is}^{(k)}\bigg\|_2=h'\bigg\|\sum_{j\in H_s}S_{ij}^{(k)}(y_j^{(k)}-y_i^{(k)})-\frac{1}{n(n-1)}\sum_{j\in H_s}(y_i^{(k)}-y_j^{(k)})\bigg\|_2\\
&\lesssim \frac{h'\eta^{(k)}}{n^2}\bigg\|\sum_{j\in H_s}(y_i^{(k)}-y_j^{(k)})\bigg\|_2=\frac{h'\eta^{(k)}}{n^2}\| [\bt^\top \yy_{1}^{(k)},  \bt^\top \yy_{2}^{(k)}]\|_2\le \frac{h'n_s\eta^{(k)}}{n^2}\max_{\ell\in[2]}\|\yy_{\ell}^{(k)}\|_2,
\end{align*}
%					\begin{align*}
%		&\bigg\|h\sum_{j\in H_s}S_{ij}^{(k)}(y_j^{(k)}-y_i^{(k)})-f_{is}^{(k)}\bigg\|_2=h\bigg\|\sum_{j\in H_s}S_{ij}^{(k)}(y_j^{(k)}-y_i^{(k)})-\frac{1}{n(n-1)}\sum_{j\in H_s}(y_i^{(k)}-y_j^{(k)})\bigg\|_2\\
%		&\lesssim \frac{h\eta^{(k)}}{n^2}\bigg\|\sum_{j\in H_s}(y_i^{(k)}-y_j^{(k)})\bigg\|_2 \lesssim \frac{h\eta^{(k)}\sqrt{n_s\eta^{(k)}}}{n^2}\lesssim \frac{h\eta^{(k)}\sqrt{n_s}}{n^2}\max_{i\ne j}\|y_{\ell i}^{(k)}-y_{\ell j}^{(k)}\|_2,
%		\end{align*} %This may not be good since the diameter is bounded by l_infty norm of the initialization
where $\bt=(t_1,...,t_n)^\top$ such that $t_i=|H_s|=n_s$ and $t_j=- 1\{j\in  H_s\}$, and the last inequality follows from $\|\bt\|_2\lesssim n_s$.  In particular, since
$
\|\yy_{\ell}^{(k)}\|_2\le \|{\bf I}-h\bL(\bS_{\alpha}^{(k-1)})\|\cdot \|\yy_{\ell}^{(k-1)}\|_2\le \|\yy_{\ell}^{(k-1)}\|_2,
$
whenever $h\|\bS_{\alpha}^{(k-1)}\|<2$, or $\|{\bf I}-h\bL(\bS_{\alpha}^{(k-1)})\|\le 1$, we have
\[
\|\yy_{\ell}^{(k)}\|_2\le 	\|\yy_{\ell}^{(0)}\|_2,
\]
whenever $\sup_{s< k}h\|\bS_{\alpha}^{(r)}\|<2$. Now since 
\begin{align*}
\sup_{s< k}\|\bL(\bS^{(s)}_\alpha )\|&\le\sup_{s< k} \|\bL(\bS^{(s)}_\alpha -\alpha\bP^*+\bH_n)\|+ \|\bL(\alpha\bP^*-\bH_n)\|\\
&\le (\alpha n\|\bP\|_\infty+1/n)\cdot \max_{i\in[n],\ell\in[2]}|y_{i\ell}^{(0)}|^2+\alpha\|\bL(\bP^*-\bP)\|+\|\bL(\alpha\bP^*-\bH_n)\|,
\end{align*}
it suffices to have { $h\|\bL(\alpha\bP^*-\bH_n)\|\le c<2$  for some constant $c$ (the last inequality is ensured by $h\|\bL(\alpha\bP^*-\bH_n)\|\le h\|\bL(\alpha\bP^*)\|\le 1+h/n$ in (T1.E))}.
Hence, combining the previous arguments, we have
\begin{align*}
\|\epsilon_i\|_2&\lesssim\bigg\|  h'\sum_{j:j\sim i}S_{ij}^{(k)}(y_j^{(k)}-y_i^{(k)})\bigg\|_2+R\bigg\|h'\sum_{j\in H_s}S_{ij}^{(k)}(y_j^{(k)}-y_i^{(k)})-f_{is}^{(k)}\bigg\|_2\\
&\lesssim \bigg[\sqrt{n}h'(\|\bP^*\|_\infty+n^{-2}) R_n+\frac{h'\eta^{(k)}}{n}\bigg]\max_{\ell\in[2]}\|\yy_{\ell}^{(0)}\|_2.
\end{align*}
On the other hand, define $D_{rs}=z_r-z_s$. Then
\begin{align*}
&\quad\bigg\|f_{is}^{(k)}-\frac{h'n_s}{n(n-1)}D_{rs}\bigg\|_2=\bigg\|\frac{h'n_s}{n(n-1)}\bigg(y_i^{(k)}-\frac{1}{n_s}\sum_{j\in H_s}y_j^{(k)}\bigg) -\frac{h'n_s}{n(n-1)}D_{rs}\bigg\|_2\\
&=\frac{h'}{n(n-1)}\bigg\|\sum_{j\in H_s}(y_i^{(k)}-y_j^{(k)}) -n_s(z_r-z_s)\bigg\|_2=\frac{h'}{n(n-1)}\bigg\|[\bt^\top (\yy_{1}^{(k)}-O\bz_{1}),  \bt^\top (\yy_{2}^{(k)}-O\bz_{2})] \bigg\|\\
&\lesssim \frac{h'n_s}{n^2}\max_{\ell\in[2]}\|\yy_{\ell}^{(k)}-O\bz_{\ell}\|_\infty\lesssim \frac{h'n_s}{n^2}\max_{\ell\in[2]}\|\yy_{\ell}^{(k)}-O\bz_{\ell}\|_2\lesssim \frac{h'n_s}{n^2} R_n\max_{\ell\in[2]}\|\yy_{\ell}^{(0)}\|_2,
\end{align*}
For sufficiently large $n$, { assuming $\|\yy_{1}^{(0)}\|_2\asymp \|\yy_{2}^{(0)}\|_2$},
\beq \label{D}
\frac{\|D_{rs}\|_2}{\max_{\ell\in[2]}\|\yy_{\ell}^{(0)}\|_2}\gtrsim \min_{\ell\in[2]}\frac{\min_{r\ne s}|z_{\ell r}-z_{\ell r}|}{\|\yy_{\ell}^{(0)}\|_2}\ge \frac{c}{{n}},
\eeq
Consequently, if { $nR_n=o(1)$,}
\begin{align*}
\|f_{is}^{(k)}\|_2&\ge \frac{h'n_s}{n^2}\|D_{rs}\|_2-\frac{h'n_sR_n}{n^2}\max_{\ell\in[2]}\|\yy_{\ell}^{(0)}\|_2\gtrsim[\frac{h'n_s}{n^3}-\frac{h'n_sR_n}{n^2}]\max_{\ell\in[2]}\|\yy_{\ell}^{(0)}\|_2\gtrsim \frac{h'n_s}{n^3}\max_{\ell\in[2]}\|\yy_{\ell}^{(0)}\|_2.
\end{align*}
If in addition { $\frac{(\|\bP^*\|_\infty+n^{-2}) R_n n^{7/2}}{n_s}\to 0$ and $n^2\eta^{(k)}/n_s\to0$ (this is implied by Proposition \ref{cont.prop} and $\sqrt{n}\max_{i\in[n],\ell\in[2]}|y_{i\ell}^{(0)}|\to0$)}, we have
\[
\|\epsilon_i\|_2/\|f_{is}^{(k)}\|_2\lesssim \frac{[\sqrt{n}h'(\|\bP^*\|_\infty+n^{-2}) R_n+h'\eta^{(k)}/n]n^3}{h'n_s}=\frac{(\|\bP^*\|_\infty+n^{-2}) R_n n^{7/2}}{n_s}+n^2\eta^{(k)}/n_s=o(1).
\]
This completes the proof for $k=K_0$.

\paragraph{Case II. $k>K_0$.} In order to show that the above results still hold for $k>k_0$, we show that (i) as $n\to\infty$
\beq \label{Y.in}
%\sup_{k_0\le k\le K}\max_{(i,j):i\sim j}\max_{\ell\in[2]}|y_{\ell i}^{(k)}-y_{\ell j}^{(k)}|/\|\yy_\ell^{(k)}\|_2=V_n=o(n^{-1}),
\sup_{k_0\le k\le K}\max_{(i,j):i\sim j}\|y_{i}^{(k)}-y_{j}^{(k)}\|_2\lesssim V_n,%=o(n^{-1}),
\eeq
and (ii) as $n\to \infty$
\beq \label{Y.out}
%\inf_{k_0\le k\le K}\min_{(i,j):i\nsim j}\min_{\ell\in[2]}|y_{\ell i}^{(k)}-y_{\ell j}^{(k)}|/\|\yy_\ell^{(k)}\|_2=B_n\gtrsim n^{-1}.
\inf_{k_0\le k\le K}\min_{(i,j):i\nsim j}\|y_{i}^{(k)}-y_{j}^{(k)}\|_2\gtrsim B_n.%\gtrsim n^{-1}.
\eeq
For any $r,s\in\{1,...,R\}$ and $r\ne s$, we choose $c(r)$ and $c(s)$ such that $c(r)\in H_r$ and $c(s)\in H_s$, and define 
\[
D_{rs}^{(k)}=y_{c(r)}^{(k)}-y_{c(s)}^{(k)}.
\]
In particular, by (\ref{Y.in}), as $n\to\infty$, the choices of specific $c(r)$ and $c(s)$ are unimportant. Now
By (\ref{Y.out}), for each $k_0\le k\le K$, we have
\begin{align*}
&\quad\|f_{is}^{(k)}\|_2=\bigg\|\frac{h'n_s}{n(n-1)}\bigg(y_i^{(k)}-\frac{1}{n_s}\sum_{j\in H_s}y_j^{(k)}\bigg) \bigg\|_2=\frac{h'}{n(n-1)}\bigg\|\sum_{j\in H_s}(y_i^{(k)}-y_j^{(k)})\bigg\|_2\\
&\ge \frac{h'n_s}{n(n-1)}(\|y_i^{(k)}-y_{j^*}^{(k)}\|_2-\max_{j_1,j_2\in H_s}\|y_{j_1}^{(k)}-y_{j_2}^{(k)}\|_2)\gtrsim \frac{h'n_s}{n^2}(B_n-V_n).%\max_{\ell\in[2]}\|\yy_{\ell}^{(k)}\|_2.
\end{align*}
%	\begin{align*}
%	&\quad\bigg\|f_{is}^{(k)}-\frac{hn_s}{n(n-1)}D^{(k)}_{rs}\bigg\|_2=\bigg\|\frac{hn_s}{n(n-1)}\bigg(y_i^{(k)}-\frac{1}{n_s}\sum_{j\in H_s}y_j^{(k)}\bigg) -\frac{hn_s}{n(n-1)}D_{rs}\bigg\|_2\\
%	&=\frac{h}{n(n-1)}\bigg\|\sum_{j\in H_s}(y_i^{(k)}-y_j^{(k)}) -n_s(y^{(k)}_{c(r)}-y^{(k)}_{c(s)})\bigg\|_2=\frac{h}{n(n-1)}\bigg\|[\bt^\top (\yy_{1}^{(k)}-\yy^{(k)}_{1^*}),  \bt^\top (\yy_{2}^{(k)}-\yy^{(k)}_{2^*})] \bigg\|\\
%	&\lesssim \frac{hn_s}{n^2}\max_{\ell\in[2]}\|\yy_{\ell}^{(k)}-\yy_{\ell^*}^{(k)}\|_2\lesssim \frac{hn_s}{n^2}n^{1/2}V_n\max_{\ell\in[2]}\|\yy_{\ell}^{(k)}\|_2,
%	\end{align*}
%	where $\yy_{\ell^*}^{(k)}$ has its $i$-th component with $i\in H_r$ as $y_{\ell,c(r)}^{(k)}$,
%	and
%	\[
%	\frac{\|D_{rs}^{(k)}\|_2}{\max_{\ell\in[2]}\|\yy_{\ell}^{(k)}\|_2}\gtrsim \inf_{k_0\le k\le K}\min_{\ell\in[2]}\frac{\min_{s\ne c}|y_{\ell,c(r)}-y_{\ell, c(s)}|}{\|\yy_{\ell}^{(k)}\|_2}= B_n.
%	\]
Hence,
\beq
\|f_{is}\|_2\gtrsim \frac{h'n_sB_n}{n^2},%\max_{\ell\in[2]}\|\yy_{\ell}^{(k)}\|_2,
\eeq
%	\begin{align*}
%	\|f_{is}\|_2&\ge \bigg[\frac{hn_sB_n}{n(n-1)}-\frac{hn_sV_n}{n^{3/2}}\bigg]\max_{\ell\in[2]}\|\yy_{\ell}^{(k)}\|_2\\
%	&\gtrsim \frac{hn_sB_n}{n^2}\max_{\ell\in[2]}\|\yy_{\ell}^{(k)}\|_2,
%	\end{align*}
whenever { $B_n\gg V_n$}. On the other hand, for the error term, we  have
\[	
\|\epsilon_i\|_2\lesssim\bigg\|  h'\sum_{j:j\sim i}S_{ij}^{(k)}(y_j^{(k)}-y_i^{(k)})\bigg\|_2+R\bigg\|h'\sum_{j\in H_s}S_{ij}^{(k)}(y_j^{(k)}-y_i^{(k)})-f_{is}^{(k)}\bigg\|_2,
\]
where
\[
\bigg\|  h'\sum_{j:j\sim i}S_{ij}^{(k)}(y_j^{(k)}-y_i^{(k)})\bigg\|_2\lesssim h'(n^{-2}+\|\bP^*\|_\infty) n_r \max_{i\sim j}\|y^{(k)}_{i}-y^{(k)}_{j}\|_2\lesssim h'(n^{-2}+\|\bP^*\|_\infty) n_rV_n, %\max_{\ell\in[2]}\|\yy_{\ell}^{(k)}\|_2,
\]
and 
\begin{align*}
&\quad\bigg\|h'\sum_{j\in H_s}S_{ij}^{(k)}(y_j^{(k)}-y_i^{(k)})-f_{is}^{(k)}\bigg\|_2\le h'\bigg\|\sum_{j\in H_s}S_{ij}^{(k)}(y_j^{(k)}-y_i^{(k)})-\frac{1}{n(n-1)}\sum_{j\in H_s}(y_i^{(k)}-y_j^{(k)})\bigg\|_2\\
&\lesssim \frac{h'\eta^{(k)}}{n^2}\bigg\|\sum_{j\in H_s}(y_i^{(k)}-y_j^{(k)})\bigg\|_2\lesssim \frac{h'\eta^{(k)}n_sB_n}{n^2},
\end{align*}
where we used the key inequality
\beq\label{diam.bnd}
\max_{i,j\in[n]}\|y_i^{(k)}-y_j^{(k)}\|_2\lesssim B_n.
\eeq
Hence, it follows that
\beq
\|\epsilon_i\|_2\lesssim (\|\bP^*\|_\infty+n^{-2}) n_rh'V_n +\frac{h'\eta^{(k)}{n_s}B_n}{n^2}
\eeq
and 
\beq
\frac{\|\epsilon_i\|_2}{\|f_{is}^{(k)}\|_2}\lesssim \frac{(\|\bP^*\|_\infty+n^{-2}) V_nn^3+\eta^{(k)}{n_s}B_n}{n_s B_n}=o(1)
\eeq
whenever { $(\|\bP^*\|_\infty +n^{-2})V_nn^2/B_n\to 0$ and $\eta^{(k)}\to 0$.} In fact, we will show in the next part that $V_n=R_n\max_{\ell\in[2]}\|\yy_\ell^{(0)}\|_2$ and $B_n=n^{-1}\max_{\ell\in[2]}\|\yy_\ell^{(0)}\|_2$, so these conditions become { $\|\bP^*\|_\infty R_nn^3\to 0$ and $\eta^{(k)}=o(1)$, and both are true under (I3) and (T1.E).}
\paragraph{Proof of (\ref{Y.in}), (\ref{Y.out}) and (\ref{diam.bnd}).} To show these two inequalities, we need to obtain a general iteration formula over the embedding stage. Note that for any $i,j\in[n]$,
\begin{align*}
\|y_i^{(k+1)}-y_j^{(k+1)}\|_2&\le \|y^{(k)}_i-y^{(k)}_j\|_2+\bigg\| h'\sum_{m:m\sim i}S_{im}^{(k)}(y_m^{(k)}-y_i^{(k)})\bigg\|_2+\bigg\| h'\sum_{m:m\sim j}S_{jm}^{(k)}(y_m^{(k)}-y_j^{(k)})\bigg\|_2\\
&\quad+\bigg\| h'\sum_{m:m\nsim i}S_{im}^{(k)}(y_m^{(k)}-y_i^{(k)})- h'\sum_{m:m\nsim j}S_{jm}^{(k)}(y_m^{(k)}-y_j^{(k)})\bigg\|_2,
\end{align*}
where
\[
\bigg\| h'\sum_{m:m\sim i}S_{im}^{(k)}(y_m^{(k)}-y_i^{(k)})\bigg\|_2\lesssim h'(\|\bP^*\|_\infty +n^{-2})n_r \max_{i\sim j}\|y_i^{(k)}-y_j^{(k)}\|_2,
\]
and, by Proposition \ref{S.lem},
\begin{align*}
&\quad\bigg\| h'\sum_{m:m\nsim i}S_{im}^{(k)}(y_m^{(k)}-y_i^{(k)})- h'\sum_{m:m\nsim j}S_{jm}^{(k)}(y_m^{(k)}-y_j^{(k)})\bigg\|_2\\
&\lesssim h'R \bigg\|\sum_{m\in H_s}[S_{im}^{(k)}(y_m^{(k)}-y_i^{(k)})- S_{jm}^{(k)}(y_m^{(k)}-y_j^{(k)})]\bigg\|_2\\
&\lesssim \frac{h'Rn_s}{n^2} \max_{i\sim j}\|y_j^{(k)}-y_i^{(k)}\|_2.
\end{align*}
Hence, we have the key iteration formula
\beq \label{upper.it}
\max_{i\sim j}\|y_i^{(k+1)}-y_j^{(k+1)}\|_2\le \bigg[1+C\bigg(h'(\|\bP^*\|_\infty +n^{-2})n_r +\frac{h'Rn_s}{n^2} \bigg)\bigg]	\max_{i\sim j}\|y_i^{(k)}-y_j^{(k)}\|_2,
\eeq
and
\beq \label{upper.diam}
\max_{i,j\in[n]}\|y_i^{(k+1)}-y_j^{(k+1)}\|_2\le \bigg[1+C\bigg(h'(\|\bP^*\|_\infty +n^{-2}) n_r +\frac{h'Rn_s}{n^2} \bigg)\bigg]	\max_{i,j\in[n]}\|y_i^{(k)}-y_j^{(k)}\|_2,
\eeq
Apply (\ref{upper.it}) and (\ref{upper.diam}) iteratively, we have
\beq
\max_{i\sim j}\|y_i^{(k_0+k)}-y_j^{(k_0+k)}\|_2\le \bigg[1+C\bigg(h'(\|\bP^*\|_\infty +n^{-2}) n_r +\frac{h'Rn_s}{n^2} \bigg)\bigg]^k	\max_{i\sim j}\|y_i^{(k_0)}-y_j^{(k_0)}\|_2,
\eeq
and
\beq
\max_{i,j\in[n]}\|y_i^{(k_0+k)}-y_j^{(k_0+k)}\|_2\le \bigg[1+C\bigg(h'(\|\bP^*\|_\infty +n^{-2})n_r +\frac{h'Rn_s}{n^2} \bigg)\bigg]^k	\max_{i,j\in[n]}\|y_i^{(k_0)}-y_j^{(k_0)}\|_2,
\eeq
Therefore, as long as { $kh'((\|\bP^*\|_\infty +n^{-2})n +\frac{R}{n})=O(1)$,} we have
\beq
\max_{i\sim j}\|y_i^{(k_0+k)}-y_j^{(k_0+k)}\|_2\lesssim \max_{i\sim j}\|y_i^{(k_0)}-y_j^{(k_0)}\|_2,
\eeq
and
\beq
\max_{i,j\in[n]}\|y_i^{(k_0+k)}-y_j^{(k_0+k)}\|_2\lesssim \max_{i,j\in[n]}\|y_i^{(k_0)}-y_j^{(k_0)}\|_2,
\eeq
Similarly, for $i\nsim j$, we also have
\begin{align*}
\|y_i^{(k+1)}-y_j^{(k+1)}\|_2&\ge \|y^{(k)}_i-y^{(k)}_j\|_2-\bigg\| h'\sum_{m:m\sim i}S_{im}^{(k)}(y_m^{(k)}-y_i^{(k)})\bigg\|_2-\bigg\| h'\sum_{m:m\sim j}S_{jm}^{(k)}(y_m^{(k)}-y_j^{(k)})\bigg\|_2\\
&\quad-\bigg\| h'\sum_{m:m\nsim i}S_{im}^{(k)}(y_m^{(k)}-y_i^{(k)})- h'\sum_{m:m\nsim j}S_{jm}^{(k)}(y_m^{(k)}-y_j^{(k)})\bigg\|_2.
\end{align*}
If {$\max_{i\sim j}\|y_j^{(k)}-y_i^{(k)}\|_2\lesssim \min_{i\nsim j}\|y_j^{(k)}-y_i^{(k)}\|_2,$} we have
\beq
\|y_i^{(k+1)}-y_j^{(k+1)}\|_2\ge\|y^{(k)}_i-y^{(k)}_j\|_2-C\bigg(h'(\|\bP^*\|_\infty +n^{-2}) n_r +\frac{h'Rn_s}{n^2} \bigg)\min_{i\nsim j}\|y_j^{(k)}-y_i^{(k)}\|_2,
\eeq
or
\beq \label{lower.it}
\min_{i\nsim j}\|y_i^{(k+1)}-y_j^{(k+1)}\|_2\ge \bigg[1-C\bigg(h'(\|\bP^*\|_\infty +n^{-2}) n_r +\frac{h'Rn_s}{n^2} \bigg)\bigg]\min_{i\nsim j}\|y_j^{(k)}-y_i^{(k)}\|_2.
\eeq
As long as { $h'(\|\bP^*\|_\infty +n^{-2}) n+\frac{h'R}{n}\le c$} for some small constant $c>0$, we have
\[
\min_{i\nsim j}\|y_i^{(k+1)}-y_j^{(k+1)}\|_2\gtrsim \min_{i\nsim j}\|y_j^{(k)}-y_i^{(k)}\|_2\gtrsim \max_{i\sim j}\|y_j^{(k)}-y_i^{(k)}\|_2\gtrsim \max_{i\sim j}\|y_j^{(k+1)}-y_i^{(k+1)}\|_2.
\]
Thus, if $\max_{i\sim j}\|y_j^{(K_0)}-y_i^{(K_0)}\|_2\lesssim \min_{i\nsim j}\|y_j^{(K_0)}-y_i^{(K_0)}\|_2,$ we can also apply (\ref{lower.it})  iteratively, to have
\beq
\min_{i\nsim j}\|y_i^{(K_0+k)}-y_j^{(K_0+k)}\|_2\ge \bigg[1-C\bigg(h'(\|\bP^*\|_\infty +n^{-2})n_r +\frac{h'Rn_s}{n^2} \bigg)\bigg]^k\min_{i\nsim j}\|y_j^{(K_0)}-y_i^{(K_0)}\|_2.
\eeq
Under the same condition that $kh'((\|\bP^*\|_\infty +n^{-2}) n +\frac{R}{n})=O(1)$, we have
\beq
\min_{i\nsim j}\|y_i^{(K_0+k)}-y_j^{(K_0+k)}\|_2\gtrsim \min_{i\nsim j}\|y_j^{(K_0)}-y_i^{(K_0)}\|_2.
\eeq
By the above arguments, we only need to show that
\beq \label{k0.1}
\max_{i\sim j}\|y_i^{(k)}-y_j^{(k)}\|_2\ll n^{-1}\max_{\ell\in[2]}\|\yy_\ell^{(0)}\|_2,
\eeq
\beq\label{k0.2}
\min_{i\nsim j}\|y_i^{(k)}-y_j^{(k)}\|_2\gtrsim n^{-1}\max_{\ell\in[2]}\|\yy_\ell^{(0)}\|_2
\eeq
hold for $k=K_0$. Then it suffices to set $V_n=R_n\max_{\ell\in[2]}\|\yy_\ell^{(0)}\|_2$ and $B_n=n^{-1}\max_{\ell\in[2]}\|\yy_\ell^{(0)}\|_2$ and $B_n\gg V_n$ holds naturally. 

To see (\ref{k0.1}),  note that for $k=K_0$,
\begin{align*}
\max_{i\sim j}\|y_i^{(k)}-y_j^{(k)}\|_2\lesssim\max_{i\sim j}|y_{\ell i}^{(k)}-y_{\ell j}^{(k)}|\lesssim \max_{\ell\in[2]}\|\yy_\ell^{(k)}-O\bz_\ell\|_\infty\lesssim R_n\max_{\ell\in[2]}\|\yy_\ell^{(0)}\|_2,
\end{align*}
which implies (\ref{k0.1}) and $V_n=R_n\max_{\ell\in[2]}\|\yy_\ell^{(0)}\|_2$ by previous assumption  $nR_n=o(1)$.
%		The last term can be bounded by the following lemma.
%		
%		\bel[$\ell_\infty$ perturbation bound]
%		Let
%		\eel

To see (\ref{k0.2}), note that for $i\nsim j$ such that $i\in H_r$ and $j\in H_s$,
\begin{align*}
\|y_i^{(k)}-y_j^{(k)}\|_2/\max_{\ell\in[2]}\|\yy_\ell^{(0)}\|_2&\ge \|D_{rs}\|_2/\max_{\ell\in[2]}\|\yy_\ell^{(0)}\|_2-\|y^{(k)}_i-y^{(k)}_j-D_{rs}\|_2/\max_{\ell\in[2]}\|\yy_\ell^{(0)}\|_2\ge cn^{-1},
\end{align*}
and that $B_n=n^{-1}\max_{\ell\in[2]}\|\yy_\ell^{(0)}\|_2$, where the last inequality follows from (\ref{D}) and 
\[
\|y^{(k)}_i-y^{(k)}_j-D_{rs}\|_2/\max_{\ell\in[2]}\|\yy_\ell^{(0)}\|_2\lesssim \max_{\ell\in[2]}\|\yy_\ell^{(k)}-O\bz_\ell\|_\infty/\max_{\ell\in[2]}\|\yy_\ell^{(0)}\|_2=o(n^{-1})
\]
Finally, to see (\ref{diam.bnd}), based on a similar argument we have
\[
\max_{i,j\in[n]}\|y_i^{(k)}-y_j^{(k)}\|_2/\max_{\ell\in[2]}\|\yy_\ell^{(0)}\|_2\le \|D_{rs}\|_2/\max_{\ell\in[2]}\|\yy_\ell^{(0)}\|_2+\|y^{(k)}_i-y^{(k)}_j-D_{rs}\|_2/\max_{\ell\in[2]}\|\yy_\ell^{(0)}\|_2\le cn^{-1},
\]
where $i\in H_s$ and $j\in H_r$.

\subsection{Proof of Theorem \ref{ini.thm}}

By construction, we have $\|\yy_1^{(0)}\|_2= \|\yy_2^{(0)}\|_2=\sigma_n$, and
\beq
\max_{i\in[n],\ell\in[2]}|y_{i\ell}^{(0)}|\lesssim \frac{\sigma_n\Phi(1-\delta)}{\sqrt{n}}=o(n^{-1/2})
\eeq
with probability at least $1-\delta$. Finally, note that by Theorem \ref{tsne.cor2},
\beq
z_{\ell r}=\btheta_r^\top\yy_\ell^{(0)}/\sqrt{n_r}=\frac{1}{n_r}\sum_{i\in H_r} y_{i\ell}^{(0)}.
\eeq
Then
\[
n|z_{\ell i}-z_{\ell j}|/\|\yy_\ell^{(0)}\|_2=\sigma_n^{-1}n\bigg|\frac{1}{n_i}\sum_{k\in H_i} y_{k\ell}^{(0)}-\frac{1}{n_j}\sum_{k\in H_j} y_{k\ell}^{(0)}\bigg|=n\bigg|\frac{1}{n_i}\sum_{k\in H_i} g_{k\ell}^{(0)}-\frac{1}{n_j}\sum_{k\in H_j} g_{k\ell}^{(0)}\bigg|/\|\bg_{\ell}\|_2.
\]
Note that $\frac{\sqrt{n}}{n_i}\sum_{k\in H_i} g_{k\ell}^{(0)}$ and $\frac{\sqrt{n}}{n_j}\sum_{k\in H_j} g_{k\ell}^{(0)}$ are independent centered random variables with variances $n^2/n_i^2$ and $n^2/n_j^2$ respectively. There exist some constants $(C,\delta)$ such that
\[
C^{-1}\le \bigg|\frac{\sqrt{n}}{n_i}\sum_{k\in H_i} g_{k\ell}^{(0)}-\frac{\sqrt{n}}{n_j}\sum_{k\in H_j} g_{k\ell}^{(0)}\bigg|\le C,
\]
with probability at least $1-\delta$. Now since $1/2\le \|\bg_{\ell}\|_2^2/n\le 2$ with probability at least $1-n^{-c}$. Then by combining the above two results, for sufficiently large $n$, we have
\beq
\frac{1}{C\sqrt{2}}\le n|z_{\ell i}-z_{\ell j}|/\|\yy_\ell^{(0)}\|_2\le C \sqrt{2}
\eeq
with probability at least $1-2\delta$. This proves the theorem.

\subsection{Proof of Theorem \ref{expand.thm}}

Define $i_0^{(k)}=\argmax_i y_{i1}^{(k)}$ and $i_1^{(k)}=\argmin_i y_{i1}^{(k)}$. 
For simplicity, we drop the superscript $(k)$ in $i_0$ and $i_1$ 
when there is no risk of confusion. In general, it suffices to show that
\beq
y_{i_0\ell}^{(k+1)}>y_{i_0\ell}^{(k)},\quad y_{i_1\ell}^{(k+1)}<y_{i_1\ell}^{(k)}
\eeq
for $\ell=1,2$, at each iteration. Without loss of generality, we only show that $	y_{i_01}^{(k+1)}>y_{i_01}^{(k)}$ as the proofs of the other results are the same.
Note that
\begin{align}
y_{i_01}^{(k+1)}&=y_{i_01}^{(k)}+h'\sum_{j:j\nsim i_0}S_{i_0j}^{(k)}(y_{j1}^{(k)}-y_{i_01}^{(k)})+h'\sum_{j:j\sim i_0}S_{i_0j}^{(k)}(y_{j1}^{(k)}-y_{i_01}^{(k)})\nonumber \\
&\equiv y_{i_01}^{(k)}+F_{i_01}^{(k)}+E_{i_01}^{(k)}.\label{decomp2}
\end{align}
By definition of $i_0$ and Proposition \ref{S.lem}, we have $y_{j1}^{(k)}-y_{i_01}^{(k)}<0$ and $S_{i_0j}^{(k)}<0$ for all $j\nsim i$, so that $F_{i_01}^{(t)}>0$. Moreover, since
\[
F_{i_01}^{(k)}\gtrsim \frac{h'}{n}\min_{j\nsim i}|y_{j1}^{(k)}-y_{i1}^{(k)}|,%\gtrsim \frac{h'}{n^3}\|\yy_1^{(0)}\|_2,
\]
and
\begin{align*}
|E_{i_01}^{(t)}|&\lesssim \max_{\substack{j\ne i_0\\ j\sim i_0}}|y_{j1}^{(k)}-y_{i_01}^{(k)}|\cdot h'\sum_{\substack{j\ne i_0\\ j\sim i_0}}S_{i_0j}^{(k)}\lesssim   h'n\|\bP^*\|_{\infty}\max_{i\sim j}|y_{j1}^{(k)}-y_{i1}^{(k)}|,
\end{align*}
by the same argument that leads to (\ref{Y.in}) and (\ref{Y.out}) in the proof of Theorem \ref{net.force.thm}, we have $\max_{i\sim j}|y_{j1}^{(k)}-y_{i1}^{(k)}|\ll \min_{j\nsim i}|y_{j1}^{(k)}-y_{i1}^{(k)}|$. Then, in equation (\ref{decomp2}), we have $F_{i_01}^{(t)}\gg|E_{i_01}^{(t)}|$ under the condition that  $\|\bP^*\|_\infty\lesssim n^{-2}$.

\section{Analysis of Two Examples}

\subsection{Proofs of the Gaussian Mixture Model}

For given $\{z_i\}_{1\le i\le n}$, we define the equivalence relationship over $[n]$ such that $i\sim j$ whenever $z_i=z_j$. 
Thus, for given $\{X_i\}_{1\le i\le n}$, we can define the symmetric matrix $\bP^*=(p_{ij}^*)\in\R^{n\times n}$ such that %$p^*_{ij}=F_{ij}(\E\bK)$ 
$p^*_{ij}=p_{ij}$ if $i\sim j$,  and $p^*_{ij}=0$ otherwise. The following proposition concerns properties of the similarity matrix $\bP$ under the Gaussian mixture model.

\bep \label{GMM.prop}
Under conditions of Corollary \ref{gmm.cor}, we have
\beq \label{P.bnd.gmm}
P\bigg(\|\bP\|_\infty\lesssim \frac{1}{n^2} \bigg)\ge 1-\frac{1}{n^c},
\eeq
and the following events 
\beq \label{P.bnd2.gmm}
\mathcal{B}_1=\bigg\{\|\bL(\bP-\bP^*)\|\lesssim \frac{1}{n}e^{-c\rho^2/p}, \|\bP-\bP^*\|_\infty\lesssim  \frac{1}{n^2}e^{-c\rho^2/p}\bigg\},
\eeq
\beq \label{P*}
\mathcal{B}_2=\bigg\{\min_{i\sim j}p^*_{ij}\gtrsim \frac{1}{n^2}, \lambda_{R+1}(\bL(\bP^*))\asymp \|\bL(\bP^*)\|\asymp\frac{1}{n}\bigg\},
\eeq
hold with probability at least $1-n^{-c}$.
\eep

\begin{proof}
	Firstly, we define
	the Gaussian kernel matrix associated with the data points $\{X_i\}_{1\le i\le n}$ as
	\beq \label{G.kernel}
	\bK = (K(X_i,X_j))_{1\le i,j\le n},\quad K(X_i,X_j)=\exp\bigg(-\frac{\|X_i-X_j\|_2^2}{2\tau_i^2}\bigg).
	\eeq
	Let $K_{ij}$ be the $(i,j)$-th entry of $\bK$.
	By the definition of $p_{ij}$, for each pair $\{i,j\}\subset \{1,...,n\}$ such that $i\ne j$, we define the map $F_{ij}: \R_{+}^{n\times n}\to (0,1)$ where
	\[
	p_{ij}=F_{ij}(\bK)=\frac{K_{ij}}{2n\sum_{\ell\ne i}K_{i\ell}}+\frac{K_{ji}}{2n\sum_{\ell\ne j}K_{j\ell}}.
	\]
	To show (\ref{P.bnd.gmm}), it suffices to show
	$
	\sum_{\ell\ne i}K_{i\ell}\gtrsim n,
	$
	with high probability,
	as by definition $K_{ij}\le 1$. This is done in the following lemma.
	
	\bel \label{lem.21}
	Under conditions of Corollary \ref{gmm.cor}, for any $i\sim j$,
	\beq \label{K.lower}
	P\big(\min\{K_{ij}, \E K_{ij}\}\gtrsim c'\big) \ge 1-n^{-c},
	\eeq
	and, for any given $i\in\{1,...,n\}$,
	\beq \label{num.sim}
	P(  |\{ s\in\{1,...,n\}: s\sim i\} |\ge c'n) \ge 1-e^{-cn}.
	\eeq
	\eel

	Next, to show (\ref{P.bnd2.gmm}), we only need to obtain an upper bound for $\max_{i\nsim j}K_{ij}$, or a lower bound for $\max_{i\nsim j}\|X_i-X_j\|^2_2$.
	We write $X_i=\mu_{z_i}+\Sigma^{1/2}W_i$ where $W_i\sim N(0,I_p)$ so that
	\[
	\|X_i-X_j\|_2^2=\|\mu_{z_i}-\mu_{z_j}\|_2^2+(W_i-W_j)^\top\Sigma(W_i-W_j)+2(\mu_{z_i}-\mu_{z_j})^\top\Sigma^{1/2}(W_i-W_j).
	\]
	On the one hand, by the Hanson-Wright inequality \citep{rudelson2013hanson}, we have, for $t\gtrsim p$,
	\[
	P(|(W_i-W_j)^\top\Sigma(W_i-W_j)-2\text{tr}(\Sigma)|>t )\le 2e^{-ct}.
	\]
	On the other hand, standard concentration inequality for sub-Gaussian random variables indicates
	\[
	P(|2(\mu_{z_i}-\mu_{z_j})^\top\Sigma^{1/2}(W_i-W_j)|>C\|\mu_{z_i}-\mu_{z_j}\|_2\sqrt{t} )\le 2e^{-ct}.
	\]
	By choosing $t=C\max\{p,\log n\}$ in the above inequalities, we have
	\[
	P(\|X_i-X_j\|_2^2\ge [\|\mu_{z_i}-\mu_{z_j}\|_2+C\max\{\sqrt{p},\sqrt{\log n}\}]^2)\le n^{-c}.
	\]
	Now since { $\rho^2 \gtrsim \max\{p,\log n\}$}, we have
$
	P(\|X_i-X_j\|_2^2\ge \|\mu_{z_i}-\mu_{z_j}\|_2^2)\le n^{-c}.
$
	In other words, under the same event, we have
	\[
	P\big(\max_{i\nsim j} K_{ij}\lesssim e^{-c\rho^2/\max\{p,\log n\}}\big)\ge 1-n^{-c}.
	\]
	Finally, to show (\ref{P*}), we define the matrix  $\bP_0\in\R^{n\times n}$ such that $p_{0,ij}=p^*_{ij}$ for $i\nsim j$, and $p_{0,ij}=F_{ij}(\E [\bK|z_1,...,z_n])$ otherwise. Conditional on $\{z_i\}$,     $\bP_0$ is a block-wise constant matrix. Then $\lambda_{R+1}(\bL(\bP_0))=\|\bL(\bP_0)\|\asymp n^{-1}$. To see this, we need the following lemma.

	\bel \label{lem.22}
	Under conditions of Corollary \ref{gmm.cor}, for any  pair $\{i,j\}\subset \{1,...,p\}$ such that $i\sim j$, we have
	\beq \label{entry.rate}
	p_{0,ij}\asymp \frac{1}{n^2},
	\eeq
	with probability at least $1-e^{-cn}$.
	\eel

	The following lemma concerns the relation between $\bP^*$ and $\bP_0$.
	
	\bel \label{lem.23}
	Under conditions of Corollary \ref{gmm.cor}, with probability at least $1-n^{-c}-n^2e^{-cn}$ for some constant $c>0$, we have
	\beq \label{p-p*}
	\|\bP^*-\bP_0\|_\infty\lesssim \frac{1}{n^2}\|\bK-\E [\bK|z_1,...,z_n]\|_{\infty}.
	\eeq
	Moreover, with probability at least $1-n^2(n^{-c}+e^{-cn}+2e^{-t^2})$, we have
	\beq \label{p-p*2}
	\|\bP^*-\bP_0\|_\infty\lesssim \frac{t}{n^2\max\{\sqrt{p},\sqrt{\log n}\}}.
	\eeq
	Consequently,  we have
	\beq\label{conc.P}
	P\bigg(\|\bP^*-\bP_0\|_\infty\lesssim \frac{\sqrt{\log n}}{n^2\max\{\sqrt{p},\sqrt{\log n}\}}\bigg) \ge 1-n^{-c}
	\eeq
	\eel

	Finally, in order to show (\ref{P*}), it suffices to use Weyl's inequality
$
	|\lambda_{R+1}(\bL(\bP_0))- \lambda_{R+1}(\bL(\bP^*))|\le \|\bL(\bP_0)-\bL(\bP^*)\|.
$
	This proves the proposition.
\end{proof}

The verification of (T1.D) and (T2.D) is straightforward.
To check (T1.E) (T2.E) and (T3.E), we note that
$
\alpha hK_0\|\bL(\bP^*-\bP)\|\lesssim K_0e^{-c\rho^2/\max\{p,\log n\}}.
$
Hence, for (T1.E) to hold, we need $K_0ne^{-c\rho^2/\max\{p,\log n\}}=o(1)$. Moreover, the condition $\textup{diam}(\{y_i^{(K_0+K_1)}\}_{1\le i\le n})=o(1)$ follows from Theorem \ref{expand.thm} and the condition $h'K_1=O(n)$.

\subsection{Proof of the Noisy Nested Sphere Model}

Similarly, we define the symmetric matrix $\bP^*=(p_{ij}^*)\in\R^{n\times n}$ such that $p^*_{ij}=p_{ij}$ if $i\sim j$,  and $p^*_{ij}=0$ otherwise. 
The following proposition concerns properties of the similarity matrix $\bP$ under the noisy nested sphere model.

\bep \label{NNS.prop}
Under the conditions of Corollary \ref{nns.cor}, we have
\beq \label{P.bnd}
P\bigg(\|\bP\|_\infty\lesssim \frac{1}{\gamma n^2}  \bigg\}\bigg)\ge 1-\frac{1}{n^{c'}},
\eeq
\beq \label{P.bnd.2.1}
P\bigg(\|\bP-\bP^*\|_\infty\lesssim  \frac{1}{\gamma n^2} \exp\bigg\{-\min_{r\in[R-1]}\frac{c(\rho_{r+1}-\rho_{r})^2}{\gamma\rho_{r+1}^2}\bigg\} \bigg)\ge 1-\frac{1}{n^{c'}},
\eeq
\beq \label{P.bnd.2}
P\bigg(\|\bL(\bP-\bP^*)\|\lesssim  \frac{1}{\gamma n} \exp\bigg\{-\min_{r\in[R-1]}\frac{c(\rho_{r+1}-\rho_{r})^2}{\gamma\rho_{r+1}^2}\bigg\} \bigg)\ge 1-\frac{1}{n^{c'}}.
\eeq
\eep

\begin{proof}
	As in the proof of Proposition \ref{GMM.prop}, we note that
	$
	p_{ij}=F_{ij}(\bK)=\frac{K_{ij}}{2n\sum_{\ell\ne i}K_{i\ell}}+\frac{K_{ji}}{2n\sum_{\ell\ne j}K_{j\ell}}.
	$
	Then we need to find lower bound for $\sum_{\ell\ne j}K_{j\ell}$ and  $\sum_{\ell\ne i}K_{i\ell}$,
	as $K_{ij}\le 1$. 
	
	\bel \label{K.lower2.lem}
	Under the conditions of Corollary \ref{nns.cor}, conditional on the event in (\ref{num.sim}), for any $i\in[1:n]$ 
	\beq \label{K.lower2}
	P\bigg({\sum_{j:j\sim i} \min\{K_{ij},\E K_{ij}\}}\ge C\gamma n\bigg) \ge 1-n^{-c}.
	\eeq
	\eel

	Now, for $i\nsim j$, we have
	\begin{align*}
	\|X_i-X_j\|_2&=\bigg\|\mu_i-\mu_j+\frac{\mu_i}{\|\mu_i\|_2}\xi_i-\frac{\mu_j}{\|\mu_j\|_2}\xi_j\bigg\|_2\\
	&=\|\theta_i(\rho_{z_i}+\xi_i)-\theta_j(\rho_{z_j}+\xi_j)\|_2\\
	&\ge |(\rho_{z_i}+\xi_i)-(\rho_{z_j}+\xi_j)|\\
	&\ge |\rho_{z_i}-\rho_{z_j}|-|\xi_i-\xi_j|\\
	&\ge |\rho_{z_i}-\rho_{z_j}|-\sigma\sqrt{\log n}
	\end{align*}
	Now  	as long as { $\min_{i\nsim j}|\rho_{z_i}-\rho_{z_j}|\ge C\sigma\sqrt{\log n}$ for some sufficiently large $C$}, we have
	\[
	\|X_i-X_j\|_2\ge C|\rho_{z_i}-\rho_{z_j}|.
	\]
	Hence, with probability at least $1-n^{-c}$, we have
	\begin{align}\label{K.up}
	\max_{i\nsim j}({K}_{ij}+K_{ji})&\lesssim \exp\bigg(-\frac{C(\rho_{z_i}-\rho_{z_j})^2}{\tau_i^2}\bigg)+\exp\bigg(-\frac{C(\rho_{z_i}-\rho_{z_j})^2}{\tau_j^2}\bigg)\\
	&\lesssim \exp\bigg(-\frac{C(\rho_{z_i}-\rho_{z_j})^2}{\max\{\tau_i^2,\tau_j^2\}}\bigg)
	\end{align}
	so that under the same event,	for $i\nsim j$, we have
	\[
	p_{ij}\lesssim \frac{K_{ij}+K_{ji}}{\gamma n^2}\lesssim \frac{1}{\gamma n^2} \exp\bigg(-\frac{C(\rho_{z_i}-\rho_{z_j})^2}{\max\{\tau_i^2,\tau_j^2\}}\bigg),
	\]
	or (\ref{P.bnd.2.1}). Finally, note that for $C'\le C/c$,
	\[
	\min_{i\nsim j}\frac{C(\rho_{z_i}-\rho_{z_j})^2}{\max\{\tau_i^2,\tau_j^2\}}=	\min_{i\nsim j}\frac{C(\rho_{z_i}-\rho_{z_j})^2}{C'\gamma\max\{\rho_{z_i}^2,\rho_{z_j}^2\}}\ge \min_{r\in[R-1]}\frac{c(\rho_{r+1}-\rho_{r})^2}{\gamma\rho_{r+1}^2}.
	\]
	This along with the fact $\|\bL(\bP^*-\bP)\|\lesssim n\|\bP^*-\bP\|_\infty$ implies (\ref{P.bnd.2}). 
\end{proof}

Now we check (T1.D) and (T2.D). Specifically, we need $K_0\to\infty$, $K_0h=o(n)$, $nh\alpha\|\bP\|_\infty=O(1)$, $K_0h\alpha\|\bL(\bP-\bP^*)\|=o(1)$,  $1/2+h/n\le h\alpha\lambda_{R+1}(\bL(\bP^*))\le h\alpha \|\bL(\bP^*)\|\le 1+h/n$ and $K_0h(n\alpha\|\bP\|_\infty+1/n)\max_{i\in[n],\ell\in[2]}|y_{i\ell}^{(0)}|^2=o(1).$ To have these conditions hold with probability at least $1-n^{-c}$, by Proposition \ref{NNS.prop}, we need
\[
K_0h=o(n),\quad\frac{h\alpha}{\gamma n}=O(1),\quad \frac{K_0h\alpha}{\gamma n}\exp\bigg\{-\min_{r\in[R-1]} \frac{c(\rho_{r+1}-\rho_{r})^2}{\gamma \rho_{r+1}^2}  \bigg\}=o(1),
\]
\beq \label{eigen.cond}
0.5\le h\alpha\lambda_{R+1}(\bL(\bP))\le \frac{h\alpha}{\gamma n}\le 1.5,
\eeq
and
\[
K_0h(\alpha /\gamma+1)\sigma_n^2\log n=o(n^2),\qquad K_0\to\infty.
\]
This proves the first statement. To show (T1.E) (T2.E) and (T3.E) hold, we only need to check the following conditions
\[
n(1-\kappa)^{K_0}+hK_0[(\alpha n\|\bP^*\|_\infty+1/n)\sigma^2_n\log n+\alpha n\|\bL(\bP^*-\bP)\|](1+n^2\|\bP^*\|_\infty)=o(1),
\]
and
\[
K_1h'(n\|\bP^*\|_\infty+1/n)=O(1),\quad  n^2\|\bP-\bP^*\|_\infty=o(1).
\]
Again, by Proposition \ref{NNS.prop}, the above conditions hold with probability at least $1-n^{-c}$ if $K_0\gg\log n$,
\[
n(1-\kappa)^{K_0}+hK_0\bigg[(\alpha/\gamma +1)\sigma^2_n\log n/n+\frac{\alpha }{\gamma }\exp\bigg\{-\min_{r\in[R-1]} \frac{c(\rho_{r+1}-\rho_{r})^2}{\gamma \rho_{r+1}^2}  \bigg\}\bigg](1+1/\gamma)=o(1),
\]
and
\[
K_1h'(1/\gamma +1)=O(n),\quad  \frac{1}{\gamma}\exp\bigg\{-\min_{r\in[R-1]} \frac{c(\rho_{r+1}-\rho_{r})^2}{\gamma \rho_{r+1}^2}  \bigg\}=o(1).
\]
This completes the proof of the corollary.

\section{Proof of Auxiliary Lemmas}

\subsection{Proof of Lemma \ref{lem.19}}

Note that
$
\bL(\alpha\bP-\bH_n)=\bL(\alpha\bP)-\bL(\bH_n)=\bL(\alpha\bP)+\frac{1}{n(n-1)}{\bf 11}^\top-\frac{1}{n-1}{\bf I}_n,
$
and the Laplacian $\bL(\alpha\bP)$ is positive semi-definite (as a result of the Ger{\v{s}}gorin circle theorem \citep{varga2010gervsgorin} and that $\bL(\alpha\bP)$ is a symmetric  diagonally dominant matrix with real non-negative diagonal entries) and has the smallest eigenvalue $\lambda_1=0$ with an eigenvector $n^{-1/2}{\bf 1}$. Then, if $\lambda_1\le ...\le \lambda_n$ are  the eigenvalues of $\bL(\alpha\bP)$, the eigenvalues of $\bL(\alpha\bP-\bH_n)$ are $(\lambda_1, \lambda_2-(n-1)^{-1},...,\lambda_n-(n-1)^{-1})$. Consequently, the smallest eigenvalue of $h\bL(\alpha\bP-\bH_n)$ is
$
\min\{0, h(\lambda_2-(n-1)^{-1})\}\in \big[-\frac{h}{n-1},0\big].
$
On the other hand, we also have $\|h\bL(\alpha\bP-\bH_n)\|\le \|h\bL(\alpha\bP)\|<2$. Then, it follows that
$
1\le \| {\bf I}-h\bL(\alpha\bP-\bH_n)\|\le 1+\frac{h}{n-1}.
$
%This completes the proof.

\subsection{Proof of Lemma \ref{lem.21}}

	\underline{Proof of (\ref{K.lower}).} On the one hand, if we let $\Sigma=U\Lambda U^\top$ be the eigen-decomposition of $\Sigma$, where $\Lambda=\text{diag}(\lambda_1,...,\lambda_p)$, then in  light of (\ref{Ek}) below,
\beq \label{A.eq}
A=\frac{1}{4}U(\frac{2}{\tau^2}{\bf I}+\Lambda ^{-1})U^\top=\frac{1}{4}U \text{diag}\big({2/\tau^2+\lambda_1^{-1}},...,{2/\tau^2+\lambda_p^{-1}}\big) U^\top,
\eeq
and therefore
\[
|2A|\cdot|2\Sigma|= \prod_{i=1}^p\bigg(\frac{1}{\tau^2}+\frac{1}{2\lambda_i}\bigg)(2\lambda_i) = \prod_{i=1}^p\bigg(\frac{2\lambda_i}{\tau^2}+1\bigg)\le \bigg(\frac{2\lambda_1}{\tau^2}+1\bigg)^p\lesssim \bigg(\frac{C}{p}+1\bigg)^p\le C'.
\]
In other words, we have shown
\beq \label{EK.lower}
\E K_{ij} \gtrsim c',\quad \text{if $i\sim j$}.
\eeq
On the other hand, we show that
\beq \label{width}
P\bigg( \frac{\|X_i-X_j\|_2^2}{2\tau^2} \le C\bigg)\le 1-n^{-c},\quad \text{if $i\sim j$.}
\eeq
To see (\ref{width}), note that $X_i-X_j\sim N(0,2\Sigma)$, by the Hanson-Wright inequality \citep{rudelson2013hanson}, for $t\gtrsim \|\Sigma\|_F^2/\|\Sigma\|_2\asymp p$, we have
$
P\big(\|X_i-X_j\|_2^2 \ge C(\text{tr}(\Sigma)+t)\big)\le e^{-ct/\|\Sigma\|_2}.
$
Whenever $\text{tr}(\Sigma)\lesssim p$ and $\|\Sigma\|_2\le C$, by setting $t\asymp \max\{p,\log n\}$, we have
$
P(\|X_i-X_j\|_2^2 \ge C(p+\log n))\le n^{-c}.
$
This implies (\ref{width})  if we choose $\tau^2\asymp \max\{p,\log n\}$. Combining (\ref{EK.lower}) and (\ref{width}), we have (\ref{K.lower}).

\underline{Proof of (\ref{num.sim}).} It is equivalent to show that, for any $x_i\stackrel{i.i.d.}{\sim} \text{Bernoulli}(\eta)$ where $i=1,...,n$ and $p\in(0,1)$, with probability at least $1-e^{-cn}$, we have
$
\sum_{i=1}^n x_i \ge c'n,
$
for some constants $c,c'>0$. By Hoeffding's inequality, for any $t\ge 0$,
$
P\big(  \frac{1}{n}\sum_{i=1}^n x_i- \eta\ge t\big)\le e^{-2nt^2},
$
which implies
$
P\bigg(  \sum_{i=1}^n x_i\ge (1-\eta)n\bigg)\le e^{-2n},
$
if we set $t=1$. This proves (\ref{num.sim}).

\subsection{Proof of Lemma \ref{lem.22}}

	Let $\mu_i=\E X_i$ and $\mu_j=\E X_j$. Since $Z_{ij}=X_i-X_j\sim N(\mu_i-\mu_j, 2\Sigma)$, we have
\begin{align}
\E K_{ij}&=\E \exp(-Z_{ij}^\top Z_{ij}/2\tau^2)\nonumber\\
&=\frac{1}{(2\pi)^{p/2}|2\Sigma|^{1/2}}\int \exp(-Z_{ij}^\top Z_{ij}/2\tau^2)\exp(-(Z_{ij}-\mu_i+\mu_j)^\top \Sigma^{-1}(z_{ij}-\mu_i+\mu_j)/4)dZ_{ij} \nonumber  \\
&=\frac{1}{(2\pi)^{p/2}|2\Sigma|^{1/2}}\int \exp(-Z_{ij}^\top Z_{ij}/2\tau^2-(Z_{ij}-\mu_i+\mu_j)^\top \Sigma^{-1}(Z_{ij}-\mu_i+\mu_j)/4)dZ_{ij}\nonumber\\
&=\frac{1}{(2\pi)^{p/2}|2\Sigma|^{1/2}}\int \exp( -Z_{ij}^\top (\frac{2}{\tau^2}{\bf I}+\Sigma^{-1})Z_{ij}/4+Z_{ij}^\top \Sigma^{-1}\Delta_{ij}/2-\Delta_{ij}^\top \Sigma^{-1}\Delta_{ij}/4)dZ_{ij}\nonumber\\
&=\frac{1}{(2\pi)^{p/2}|2\Sigma|^{1/2}}\exp(b^\top A^{-1}b/4-\Delta_{ij}^\top\Sigma^{-1}\Delta_{ij}/4)\int \exp(-(Z_{ij}-A^{-1}b)^\top A (Z_{ij}-A^{-1}b))dZ_{ij}\nonumber\\
&=\frac{1}{|2A|^{1/2}|2\Sigma|^{1/2}}\exp(b^\top A^{-1}b/4-\Delta_{ij}^\top\Sigma^{-1}\Delta_{ij}/4) \label{Ek}
\end{align}
where $A=(\frac{2}{\tau^2}{\bf I}+\Sigma^{-1})/4$ and $b=\Sigma^{-1}\Delta_{ij}/2$. The results  follows by noting that $\Delta_{ij}=0$.

On the one hand, by (\ref{A.eq}), we have
\[
|2A|\cdot|2\Sigma|= \prod_{i=1}^p\bigg(\frac{1}{\tau^2}+\frac{1}{2\lambda_i}\bigg)(2\lambda_i) = \prod_{i=1}^p\bigg(\frac{2\lambda_i}{\tau^2}+1\bigg)\le \bigg(\frac{2\lambda_1}{\tau^2}+1\bigg)^p\lesssim \bigg(\frac{C}{p}+1\bigg)^p\le C'.
\]
On the other hand,
\[
|2A|\cdot|2\Sigma|= \prod_{i=1}^p\bigg(\frac{1}{\tau^2}+\frac{1}{2\lambda_i}\bigg)(2\lambda_i) = \prod_{i=1}^p\bigg(\frac{2\lambda_i}{\tau^2}+1\bigg)\ge 1.
\]
This implies $\E K_{ij}\asymp 1$.

To obtain bounds for $p_{0,ij}$, it suffices to see that 
$
P\big(\sum_{\ell\ne i}\E K_{\ell i}\gtrsim n\big)\ge 1-e^{-cn},
$
which follows from (\ref{num.sim}).

\subsection{Proof of Lemma \ref{lem.23}}

We start with the proof of (\ref{p-p*}).
For any $\{i,j\}\subset \{1,...,n\}$ such that $i\ne j$, we have
\beq \label{dF}
\frac{\partial F_{ij}(\bK)}{\partial K_{k\ell}}=\left\{ \begin{array}{ll}
	\frac{\sum_{s\notin\{i,j\} }K_{is}}{2n(\sum_{s\ne i}K_{is})^2}+\frac{\sum_{s\notin\{i,j\} }K_{js}}{2n(\sum_{s\ne j}K_{js})^2}, & \textrm{if $\{k,\ell\}=\{i,j\}$}\\
	-\frac{K_{k\ell}}{2n(\sum_{s\ne i}K_{is})^2}, & \textrm{if $i\in\{k,\ell\}, j\notin\{k,\ell\}$}\\
	-\frac{K_{k\ell}}{2n(\sum_{s\ne j}K_{js})^2}, & \textrm{if $j\in\{k,\ell\}, i\notin\{k,\ell\}$}
	\\
	0, & \textrm{otherwise}
\end{array} \right. .
\eeq
For any $\bK_1,\bK_2\in \R^{n\times n}_{+}$ and any $\{i,j\}\subset \{1,...,p\}$ such that $i\ne j$, we have
\begin{align*}
F_{ij}(\bK_1)-F_{ij}(\bK_2)&=  \sum_{1\le k\ne \ell\le n}\frac{\partial F_{ij}(K^*)}{\partial K_{k\ell}} ([\bK_1]_{k\ell}-[\bK_2]_{k\ell})\le \|\bK_1-\bK_2\|_\infty\sum_{1\le k\ne \ell\le n} \bigg| \frac{\partial F_{ij}(\bK^*)}{\partial K_{k\ell}} \bigg|,
\end{align*}
where $\bK^*=t\bK_1+(1-t)\bK_2$ for some $t\in(0,1)$.

In the following, we show that for $\bK_1=\bK$ and $\bK_2=\E [\bK|z_1,...,z_n]$, for given $i\ne j$, it holds that
\beq \label{p-p*0}
\bigg| \frac{\partial F_{ij}(\bK^*)}{\partial K_{k\ell}} \bigg| \lesssim \frac{1}{n^2},\quad \text{for $\{k,\ell\}=\{i,j\}$},
\eeq
and
\beq \label{p-p*1}
\bigg| \frac{\partial F_{ij}(\bK^*)}{\partial K_{k\ell}} \bigg| \lesssim \frac{1}{n^3},\quad \text{for $j\in\{k,\ell\}, i\notin\{k,\ell\}$ or $i\in\{k,\ell\}, j\notin\{k,\ell\}$},
\eeq
with probability at least $1-e^{-cp}-e^{-cn}$ for some constant $c>0$,
and conclude that
\beq
\sum_{1\le k\ne \ell\le n} \bigg| \frac{\partial F_{ij}(\bK^*)}{\partial K_{k\ell}} \bigg|\lesssim \frac{1}{n^2},
\eeq
under the same event. This along with a union bound argument leads to (\ref{p-p*}). To show (\ref{p-p*0}), we note that
\[
\bigg| \frac{\partial F_{ij}(\bK)}{\partial K_{ij}} \bigg|=\frac{\sum_{s\notin\{i,j\} }K_{is}}{2n(\sum_{s\ne i}K_{is})^2}+\frac{\sum_{s\notin\{i,j\} }K_{js}}{2n(\sum_{s\ne j}K_{js})^2}\le \frac{1}{2n\sum_{s\ne i}K_{is}}+\frac{1}{2n\sum_{s\ne j}K_{js}}.
\]
Note that by (\ref{K.lower}), we have
\[
\frac{1}{\sum_{s\ne i}K^*_{is}}+\frac{1}{\sum_{s\ne j}K^*_{js}} \lesssim \frac{1}{n},
\]
with probability at least $1-e^{-cp}-e^{-cn}$ for some constant $c>0$.  
Therefore (\ref{p-p*0}) holds with high probability. On the other hand, to show (\ref{p-p*1}), it suffices to see that, for $j\in\{k,\ell\}, i\notin\{k,\ell\}$, 
\[
\bigg| \frac{\partial F_{ij}(\bK^*)}{\partial K_{k\ell}} \bigg| \lesssim \frac{1}{n(\sum_{s\ne i}K^*_{is})^2}\lesssim \frac{1}{n^3}
\]
with probability at least $1-e^{-cp}-e^{-cn}$.  This completes the proof of (\ref{p-p*}).

Next we prove (\ref{p-p*2}). The proof follows directly from
\beq
P( \|\bK-\E [\bK|z_1,...,z_n]\|_\infty \ge {t}/\max\{\sqrt{p},\sqrt{\log n}\})\le 2n^2 e^{-t^2}.
\eeq
Note that, the kernel function $K(X_i,X_j): \R^{2p}\to \R$ is $L$-Liptschitz with respect to the $\ell_2$ norm on $\R^{2p}$, with $L= \frac{\sqrt{2}}{e\tau}\asymp \frac{1}{\max\{\sqrt{p},\sqrt{\log n}\}}$. By the concentration inequality for Liptschitz continuous functions (see, e.g., Theorem 4 of \cite{amini2021concentration}), it holds that
\[
P( |K(X_i,X_j)-\E K(X_i,X_j)| \ge Ct/\max\{\sqrt{p},\sqrt{\log n}\})\le 2e^{-ct^2}.
\]
By applying the union bound, we have
\begin{align*}
P( \|\bK-\E [\bK|z_1,...,z_n]\|_\infty \ge C{t}/{\sqrt{p}})&=P\big(\max_{1\le i\ne j\le n}|K(X_i,X_j)-\E K(X_i,X_j)| \ge C{t}/\max\{\sqrt{p},\sqrt{\log n}\}\big)\\
&\le 2n^2e^{-ct^2}.
\end{align*}
This  completes the proof.

\subsection{Proof of Lemma \ref{K.lower2.lem}}

	Suppose $\{z_i\}$ and $\{\mu_i\}$ are given.
Let $S_i(\gamma)=\{j\in[n]: j\sim i, \frac{\mu_i^\top\mu_j}{\|\mu_i\|_2\|\mu_j\|_2}\ge 1-\gamma\}$ for some  $\gamma\in(0,1)$. We show that for some properly chosen $\{\tau_i\}$,
\beq \label{K.low}
\min\{K_{ij}, \E K_{ij}\}\ge C,\quad \text{ for any $j\in S_i(\gamma)$, }
\eeq
for some constant $C>0$ with probability at least $1-n^{-c}$. Apparently, this leads to
$
\sum_{j:j\sim i} \min\{K_{ij}, \E K_{ij}\}\ge \sum_{j\in S_i(\gamma)} \min\{K_{ij}, \E K_{ij}\}\ge C|S_i(\gamma)|,
$
and we only need to show that $|S_i(\gamma)|\ge \gamma n$ with the claimed probability. 
We first show (\ref{K.low}).
Since for any $j\in S_i(\gamma)$, if we denote $\theta_i=\mu_i/\|\mu_i\|_2$, we have
\begin{align*}
\|X_i-X_j\|_2^2&\lesssim \|\theta_i-\theta_j\|_2^2\rho_{z_i}^2+|\xi_i+\xi_j|^2\\
&\lesssim (1-\theta_i^\top\theta_j)\rho^2_{z_i}+\sigma^2\log n\\
&\lesssim \gamma\rho^2_{z_i}+\sigma^2\log n
\end{align*}
with probability at least $1-n^{-c}$.  Here we used the tail bound
$
P(|\xi_i+\xi_j|\le C\sigma\sqrt{\log n})\ge 1-n^{-c},
$
In particular, if { $\gamma\min_{r\in[R]}\rho_{r}^2\gg\sigma^2\log n$,} the above argument leads to
\beq
P\bigg(\min\{K_{ij}, \E K_{ij}\}\gtrsim \exp\bigg\{-C\frac{\gamma\rho^2_{z_i}}{\tau_i^2}\bigg\}\bigg)\ge 1-n^{-c}.
\eeq
Therefore, if we choose $\tau_i$ such that { $\tau_i^2\gtrsim \gamma\rho^2_{z_i}$}, we have (\ref{K.low}).

Secondly, we obtain lower bound for $|S_i(\gamma)|$. Note that $\{\theta_i\}$ are uniformly drawn on $\mathbb{S}^{p-1}$. It follows from the spherical area formula that  $|S_i(\gamma)|$ is a binomial random variable with distribution $\text{Bin}(n, \gamma/2)$, so that
$
P\big( \bigg||S_i|-\frac{n\gamma}{2}\bigg|\lesssim t\sqrt{\gamma(1-\gamma)n} \big)\ge 1-e^{-t^2}.
$
By choosing $t=C\sqrt{\log n}$, we have
$
P( |S_i|\ge n\gamma/2- C\sqrt{\gamma n\log n} )\ge 1-n^{-c}.
$
If { $\gamma\gtrsim {\log n/n}$}, we have
$
P( |S_i|\gtrsim n\gamma)\ge 1-n^{-c}.
$
This proves the lemma.

	\section{Supplementary Figures} \label{supp.fig.sec}
	
\begin{figure}[h!]
		\centering
		\includegraphics[angle=0,width=11cm]{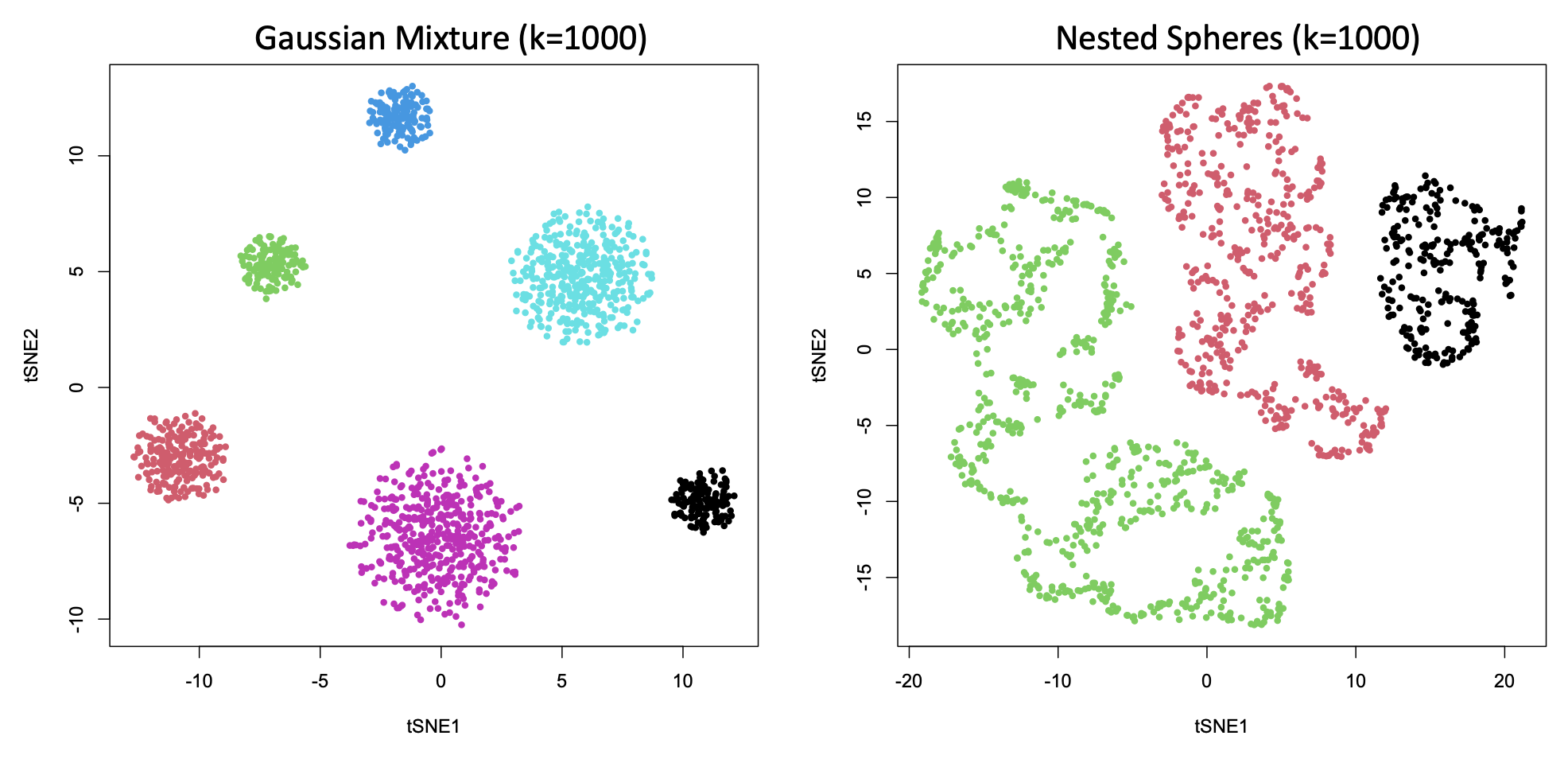}
		\caption{The final t-SNE embeddings of the model-generated samples as described in Section \ref{examples.sec}, using the tuning parameters with $\delta=1/2$.} 
		\label{simu-supp.fig}
	\end{figure}

\begin{figure}[h!]
		\centering
		\includegraphics[angle=0,width=11cm]{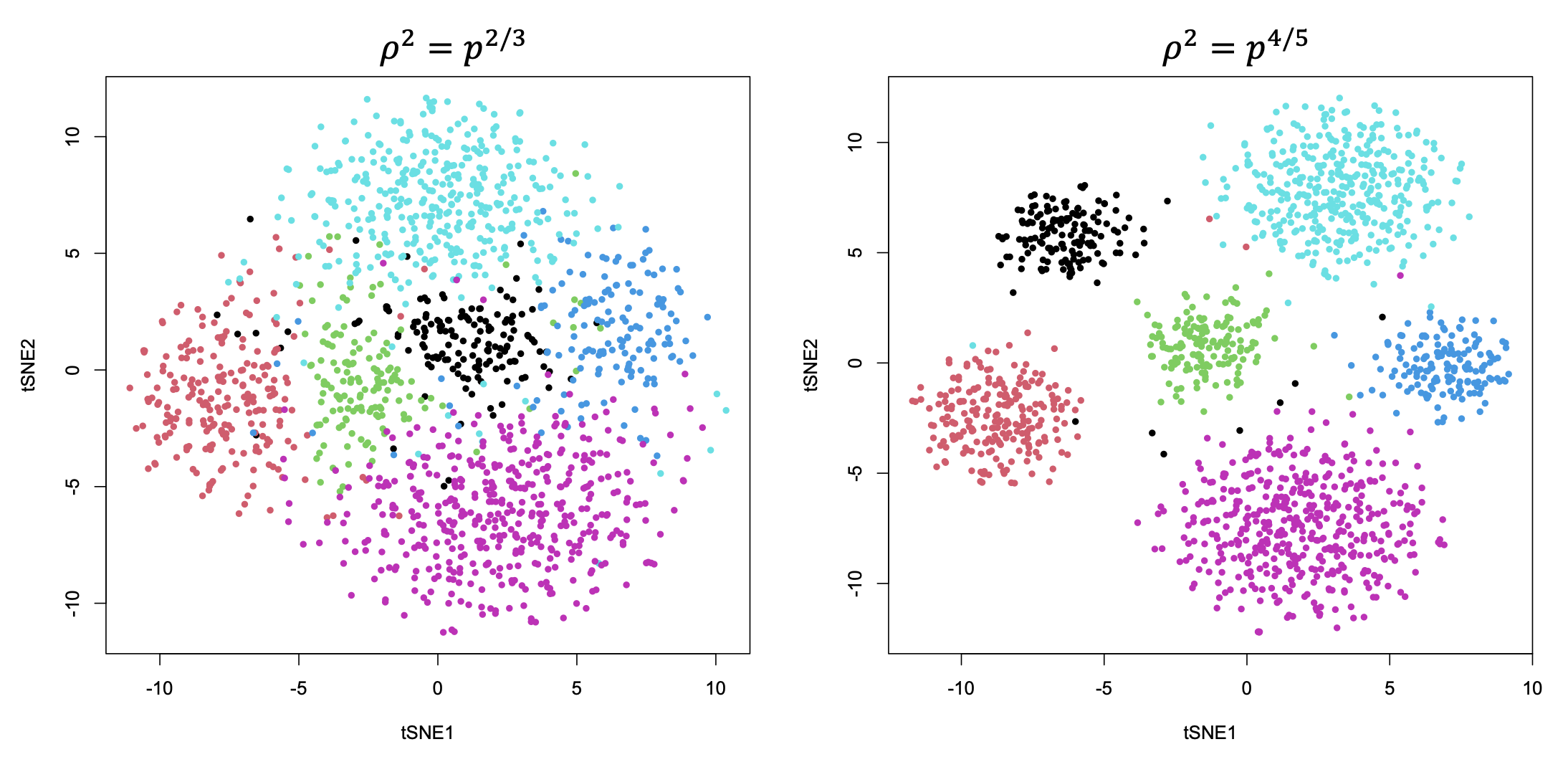}
		\caption{The final t-SNE embeddings of the samples generated from the Gaussian mixture model with separation $\rho^2=p^{2/3}$ (left) and $\rho^2=p^{4/5}$ (right), using the tuning parameters with $\delta=1/3$.} 
		\label{simu-supp-rho.fig}
	\end{figure}

	This section includes additional figures from the numerical studies presented in Sections \ref{examples.sec} and \ref{real.sec}.  
	Specifically,  Figure \ref{simu-supp.fig} contains the final t-SNE embeddings of the model-generated samples as described in Section \ref{examples.sec}, but with different tuning parameters where $\delta=1/2$.  Figure \ref{simu-supp-rho.fig} shows that when the separation condition $\rho^2\gg p$ is slightly violated, t-SNE is still able to visualize clusters from the Gaussian mixture model, which demonstrates the robustness of t-SNE with respect to the separation condition.

	\begin{figure}[h!]
		\centering
		\includegraphics[angle=0,width=5cm]{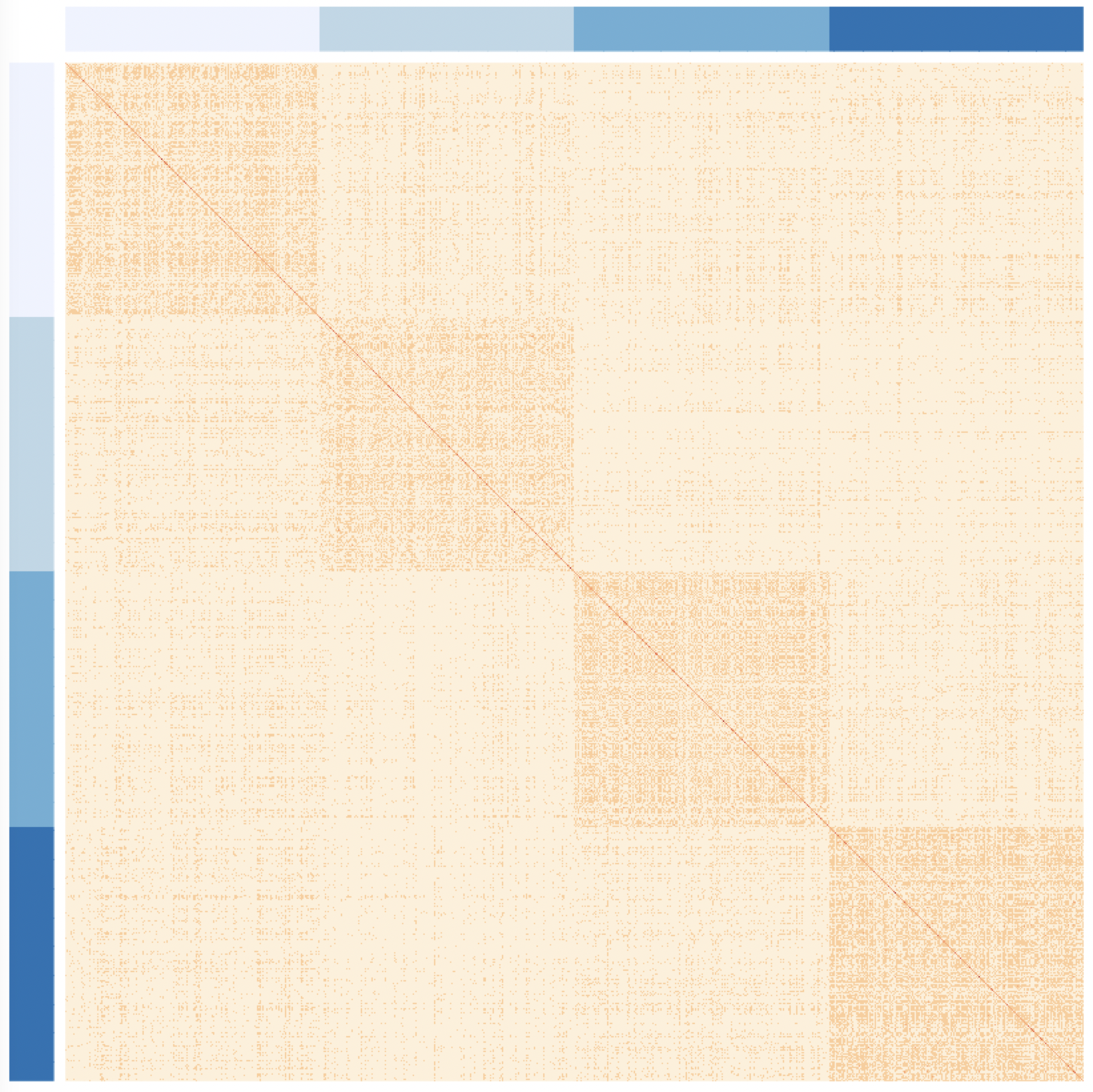}
		\caption{A heatmap of the similarity matrix $\bP$ for the $n=1600$ MNIST samples corresponding to digits ``2," ``4," ``6," and ``8," analyzed in Section \ref{real.sec}. The color bars represent the cluster labels of the columns and rows.} 
		\label{block-supp.fig}
	\end{figure} 
	
	\begin{figure}[h!]
		\centering
		\includegraphics[angle=0,width=15cm]{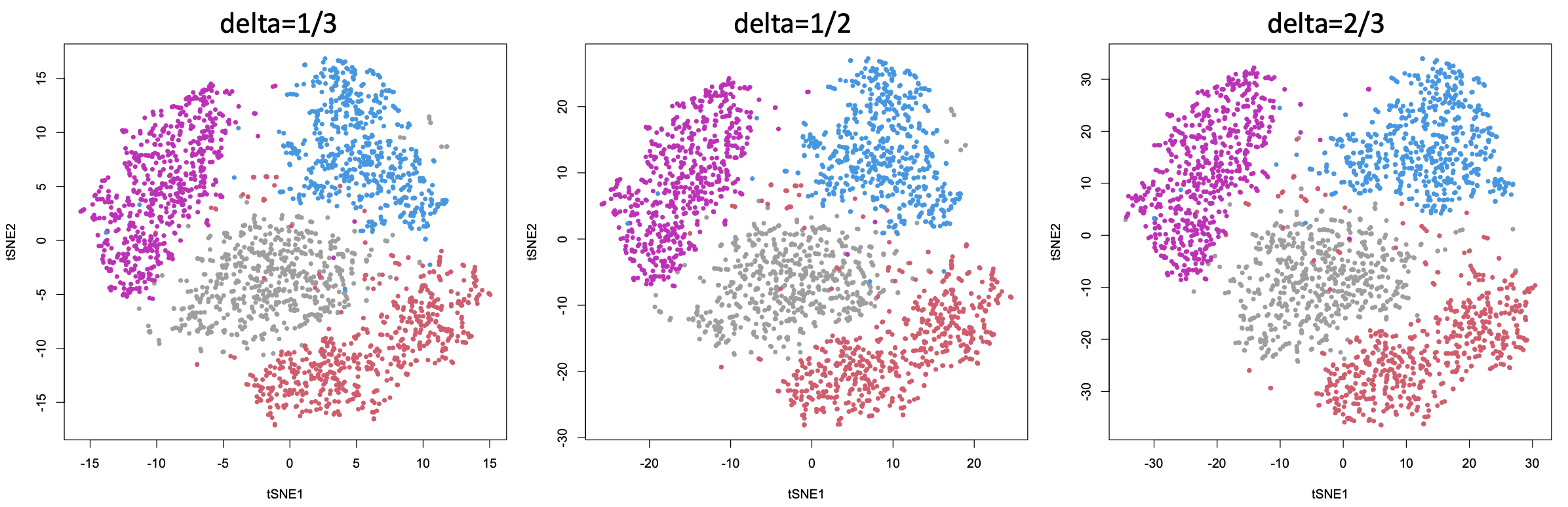}
		\caption{t-SNE visualizations of $n=2400$ MNIST samples with an identical random initialization but different values of $\delta$ for the tuning parameters in (\ref{tuning}).} 
		\label{delta.fig}
	\end{figure}

	Figure \ref{block-supp.fig} is a heatmap of the similarity matrix $\bP$ for the $n=1600$ MNIST samples corresponding to digits ``2," ``4," ``6," and ``8," analyzed in Section \ref{real.sec}. It justifies our assumption on the approximate block structure on $\bP$. Figure \ref{delta.fig} contains t-SNE visualizations of $n=2400$ MNIST samples with an identical random initialization but different values of $\delta$ for the tuning parameters in (\ref{tuning}). The similarity in the cluster patterns indicates robustness and flexibility of our theory-guided choices for the tuning parameters.

\end{document}